\documentclass{article}

\usepackage{microtype}
\usepackage{graphicx}
\usepackage{booktabs}
\usepackage{hyperref}

\usepackage[utf8]{inputenc}
\usepackage[T1]{fontenc}
\usepackage{adjustbox}
\usepackage[noend]{algorithmic}
\usepackage{algorithmic}
\usepackage{amsfonts}
\usepackage[font=scriptsize]{subfig}
\usepackage{amsmath}
\usepackage{amssymb}
\usepackage{balance}
\usepackage{bbm}
\usepackage{bm}
\usepackage[font=small]{caption}
\usepackage{cite}
\usepackage{clipboard}
\usepackage{contour}
\usepackage{enumitem}
\usepackage{float}
\usepackage[bottom,flushmargin]{footmisc}
\usepackage{makecell}
\usepackage{mathbbol}
\usepackage{mathtools, nccmath}
\usepackage{mathrsfs}
\usepackage{microtype}
\usepackage{multirow}
\usepackage{nicefrac}
\usepackage{pifont}
\usepackage{relsize}
\usepackage{setspace}
\usepackage{stackengine}
\usepackage{tcolorbox}
\usepackage{thmtools}
\usepackage{thm-restate}
\usepackage{tikz}
\usepackage{transparent}
\usepackage{url}
\usepackage{wrapfig}
\usepackage{xcolor}

\DeclareSymbolFontAlphabet{\mathbb}{AMSb}
\DeclareSymbolFontAlphabet{\mathbbl}{bbold}

\usepackage[noend]{algorithmic}
\renewcommand\algorithmiccomment[1]{
\hfill$\triangleright$ {#1}\hspace{-0.6em}
}

\usepackage[accepted]{styles/icml2020}

\definecolor{crimson}{rgb}{0.7,0.01,0.02}
\definecolor{fern}{rgb}{0.05,0.5,0.15}
\definecolor{prussian}{rgb}{0,0.08,0.45}
\definecolor{faintgray}{gray}{0.9}
\definecolor{faintblue}{rgb}{0,0.08,0.65}

\newcommand{\pix}{\kern 0.1em}
\newcommand{\pmm}{\kern 0.25em$\pm$\kern 0.15em}
\newcommand{\pms}{\kern 0.10em$\pm$\kern 0.05em}
\newcommand{\cm}{\ding{51}}
\newcommand{\xm}{{\color{faintgray}\ding{55}}}

\newcommand{\QED}{\hfill\raisebox{-0.5pt}{\scalebox{0.88}{$\square$}}}
\newcommand{\dayum}[1]{{#1\parfillskip=0pt\par}}

\newcommand{\spaceqq}{\kern 0.2em=}
\newcommand{\spaceapp}{\kern 0.2em\approx}
\newcommand{\spacegeqq}{\kern 0.2em\geq}
\newcommand{\spaceleqq}{\kern 0.2em\leq}
\newcommand{\eqqcolon}{\kern 0.2em=:}
\newcommand{\shrink}[1]{\scalebox{0.9}{#1}}

\declaretheorem[name=Theorem]{retheorem}

\declaretheorem[name=Lemma,numberlike=retheorem]{relemma}

\declaretheorem[name=Definition]{redefinition}

\declaretheorem[name=Objective]{reobjective}
\declaretheorem[name=Example]{reexample}

\newcommand{\tttext}[1]{\shrink{\texttt{#1}}}

\makeatletter
\newcommand{\raisemath}[1]{\mathpalette{\raisem@th{#1}}}
\newcommand{\raisem@th}[3]{\raisebox{#1}{$#2#3$}}
\makeatother

\usepackage[normalem]{ulem}

\contourlength{0.6pt}
\newcommand{\muline}[1]{%
\uline{\phantom{#1}}%
\llap{\contour{white}{#1}}%
}

\usepackage{etoolbox}

\usepackage{tikz}

\newrobustcmd*{\mytriangle}[1]{\tikz{\filldraw[draw=#1,fill=#1] (0,0) --
(0.15cm,0) -- (0.08cm,0.15cm);}}

\let\OLDthebibliography\thebibliography

\renewcommand\thebibliography[1]{
\OLDthebibliography{#1}
\setlength{\itemsep}{5pt plus 2pt minus 3pt}
}

\newif\ifdropfigures
\dropfiguresfalse
\ifdropfigures
\renewcommand{\includegraphics}[2][]{%
}
\fi

\usepackage{etoc}

\makeatletter
\def\@part[#1]#2{
\addcontentsline{toc}{part}{#1}
{\large\bfseries #2}
\nobreak
\vspace{-2em}
\@afterheading}
\makeatother

\newcommand{\mytitle}{Curiosity in Hindsight:\\\mbox{Intrinsic Exploration in Stochastic Environments}}

\icmltitlerunning{Curiosity in Hindsight}

\begin{document}


\twocolumn[
\icmltitle{\mytitle}

\icmlsetsymbol{equal}{*}
\begin{icmlauthorlist}
\icmlauthor{Daniel Jarrett}{deep}
\icmlauthor{Corentin Tallec}{deep}
\icmlauthor{Florent Altch\'{e}}{deep}
\icmlauthor{Thomas Mesnard}{deep}
\icmlauthor{R\'{e}mi Munos}{deep}
\icmlauthor{Michal Valko}{deep}
\end{icmlauthorlist}

\vspace{-0.75em}

\icmlaffiliation{deep}{DeepMind}
\icmlcorrespondingauthor{Dan Jarrett}{jarrettd@google.com}
\icmlkeywords{Machine Learning, ICML}

\vskip 0.3in
]

\printAffiliationsAndNotice{}

\allowdisplaybreaks


\begin{abstract}
\dayum{
Consider the problem of exploration in sparse-reward or reward-free environments, such as in Montezuma's Revenge.
In the \textit{curiosity-driven} paradigm,
the agent is rewarded for how much each realized outcome differs from their predicted outcome.
But using predictive error as intrinsic motivation is fragile in \textit{stochastic environments}, as the agent may become trapped by high-entropy areas of the state-action space, such as a ``noisy TV''.
In this work, we study a natural solution derived from structural causal models of the world:
Our key idea is to learn representations of the future that capture precisely the \textit{unpredictable} aspects of each outcome---which we use as additional input for predictions, such that intrinsic rewards only reflect the \textit{predictable} aspects of world dynamics.
First, we propose incorporating such hindsight representations into models to disentangle ``noise'' from ``novelty'', yielding \textit{Curiosity in Hindsight}: a simple and scalable generalization of curiosity that is robust to stochasticity.
Second, we instantiate this framework for the recently introduced \tttext{BYOL-Explore} algorithm as our prime example, resulting in the noise-robust \tttext{BYOL-Hindsight}.
Third, we illustrate its behavior under a variety of different stochasticities in a grid world, and find improvements over \tttext{BYOL-Explore} in hard-exploration Atari games with sticky actions. Notably, we show state-of-the-art results in exploring Montezuma's Revenge with sticky actions, while preserving performance in the non-sticky setting.
}
\end{abstract}

\vspace{-2em}
\section{Introduction}\label{sec:1}

\dayum{
Learning to understand the world without supervision is a hallmark of intelligent behavior \cite{kriegeskorte2018cognitive}, and \textit{exploration} is a key pillar of research in reinforcement learning agents \cite{yang2021exploration}. How might an agent learn meaningful behaviors when external rewards are sparse or absent?
A predominant approach is given by the \textit{curiosity-driven} paradigm \cite{schmidhuber1991possibility}, in which an agent's ability to predict the future is used as a proxy for their ``understanding'' of the world.
Maintaining a learned model of the environment, at each step the agent receives an intrinsic reward proportional to how much the realized outcome differs from their predicted outcome---which naturally directs them towards new areas that have not been seen.
}

\dayum{
There are two major hurdles.
The first concerns \textit{dimensionality}:
While outcomes can be predicted directly at the level of observations \cite{thrun1995exploration,barto2004intrinsically,oh2015action,finn2016unsupervised,gregor2019shaping}, pixel-based losses have generally not worked well in higher dimensions \cite{burda2019large}.
Popular solutions thus operate on lower-dimensional \textit{latent representations}, such as frame-predictive features \cite{stadie2016incentivizing}, inverse dynamics features \cite{pathak2017curiosity}, random features
\cite{burda2019exploration}, or features that maximize information across time \cite{kim2019emi}.
Most recently, bootstrapped features are employed in \tttext{BYOL-Explore} \cite{guo2022byol}---achieving superhuman performance on hard-exploration games in Atari with a much simpler design than comparable agents.
}

\dayum{
The second concerns \textit{stochasticity}, which is our focus here:
Curiosity-driven agents are often susceptible to bad behavior in environments with stochastic transitions, since they are often hopelessly attracted to high-entropy elements in the state-action space \cite{burda2019large}.
A classic example is the problem of a ``noisy TV'', which generates a stream of intrinsic rewards around which predictive error-based agents become stuck indefinitely \cite{mavor2022stay}.
More generally, this problem manifests with respect to any aspect of environment dynamics that is inherently unpredictable, including noise specific to certain states, as well as noise actively induced by the agent.
}

\dayum{
\textbf{Novelty vs. Noise}~
In the presence of stochasticity, predictive error \textit{per se} is no longer a good measure for an agent's lack of ``understanding'' of the world. Intuitively, we wish to measure their understanding by how much \textit{epistemic} knowledge they have acquired (viz. necessary truths about how the world works in general), which is entirely orthogonal to how much \textit{aleatoric} variation each outcome can display (viz. contingent facts about how the world happens to be).
Precisely, we want to distinguish between aspects of world dynamics that are inherently predictable---for which (reducible) errors stem from ``novelty''---and aspects that are inherently unpredictable---for which (irreducible) errors stem from ``noise''. Crucially, while the former should contribute to intrinsic rewards for exploration, the latter should not.
}

\dayum{
\textbf{Contributions}~
We operationalize this distinction by deriving a solution based on structural causal models of the world: Our key idea is to learn representations of the future that capture precisely the unpredictable aspects of each outcome---no more, no less---which we use as additional input for predictions, such that intrinsic rewards vanish in~the~limit.
First, we propose incorporating such hindsight representations into the agent's model to disentangle ``noise'' from ``novelty'', yielding \textit{Curiosity in Hindsight}: a simple and scalable generalization of curiosity-driven exploration that is robust to stochasticity (Section \ref{sec:3}).
Second, we instantiate this framework for the recently introduced \tttext{BYOL-Explore} algorithm as our prime example, giving rise to the noise-robust \tttext{BYOL-Hindsight} (Section \ref{sec:4}).
Third, we illustrate its behavior under a variety of different stochasticities in a grid world, and find improvements over \tttext{BYOL-Explore} in hard-exploration Atari games with sticky actions (a standard protocol for introducing stochasticity in training/evaluation). Notably, we show state-of-the-art results in exploring Montezuma's Revenge with sticky actions, while preserving its original performance in the non-sticky setting (Section \ref{sec:5}).
}
\vspace{-0.5em}
\section{Motivation}\label{sec:2}

\vspace{-0.15em}
\subsection{Problem Formalism}

\dayum{
Consider the standard MDP setup.
We employ uppercase for random variables and lowercase for specific values: Let $X$ denote the \textit{state} variable, taking on values $x\in\mathcal{X}$, and $A$ the \textit{action} variable, taking on values $a\in\mathcal{A}$. While we keep notation simple, $X$ may play the role of ``contexts'', ``features'', ``embeddings'', or ``beliefs'' depending on environment observability and the design of the agent. Let $\tau$\pix$\in$\pix$\Delta(\mathcal{X})^{\mathcal{X}\times\mathcal{A}}$ denote the world's dynamics such that $X_{t+1}$\pix$\sim$\pix$\tau(\cdot|x_{t},a_{t})$, and $\pi$\pix$\in$\pix$\Delta(\mathcal{A})^{\mathcal{X}}$ the agent's policy such that $A_{t}$\pix$\sim$\pix$\pi(\cdot|x_{t})$. Lastly, let $\rho_{\pi}$ denote the distribution of states induced by $\pi$.
}

\setcounter{footnote}{1}

\begin{redefinition}[restate=defcuriosity,name=Curiosity-driven Exploration]\upshape\label{def:curiosity}
In this work we focus on \textit{predictive error-based} curiosity---subsuming most popular approaches to curiosity. Intrinsic rewards are:
\begin{align}
\mathcal{R}_{\eta}(x_{t},a_{t})
&
\coloneqq
-
\mathbb{E}_{X_{t+1}\sim\tau(\cdot|x_{t},a_{t})}
\log\tau_{\eta}(X_{t+1}|x_{t},a_{t})
\label{eqn:original}
\end{align}
\dayum{%
where $\tau_{\eta}$ is the agent's world model\footnote{Note that this is only used for computing rewards, and need not be related to the underlying RL algorithm, which can be model-free.}
parameterized by $\eta$, and is trained using the trajectories collected by rolling out a policy that seeks to maximize this same prediction error:
}
\vspace{-1em}
\begin{align}
\overset{\text{(policy)}}{
\underset{\pi}{\text{maximize}}
}
~~
\overset{\text{(model)}}{
\underset{\eta}{\text{min}}
}
~~
\mathbb{E}_{\substack{X_{t}\sim\rho_{\pi}\\A_{t}\sim\pi(\cdot|X_{t})}}
\mathcal{R}_{\eta}(X_{t},A_{t})
\end{align}
\end{redefinition}
\vspace{-1.25em}

\begin{reexample}[restate=exabyolexplore,name=Bootstrapping Representations]\upshape\label{exa:byolexplore}
As our key example,
recall \tttext{BYOL-Explore} \cite{guo2022byol}, a most recent and suc-cessful incarnation of this paradigm.
The \textit{prediction loss} for a given transition $(x_{t},a_{t},x_{t+1})$ is defined as the following:
\end{reexample}

\begin{align}
\mathcal{L}_{\eta}^{\tiny\tttext{BYOL}}(x_{t},a_{t},x_{t+1})
\coloneqq
\big\|
x_{t+1}-\hat{x}_{t+1}
\big\|_{2}^{2}
~~~
\end{align}

\vspace{-0.75em}
and the (state-action) \textit{prediction bonus} for the agent's policy:
\vspace{-2em}

\begin{align}
\mathcal{R}_{\eta}^{\tiny\tttext{BYOL}}(x_{t},a_{t})
\coloneqq
\mathbb{E}_{X_{t+1}\sim\tau(\cdot|x_{t},a_{t})}
\mathcal{L}_{\eta}^{\tiny\tttext{BYOL}}(x_{t},a_{t},X_{t+1})
\end{align}

\vspace{-0.75em}
where the novelty of the method lies in the manner in which specific quantities are defined and learned:
\textbf{(i.)} Input states are RNN ``belief'' representations $x_{t}$\pix$\coloneqq$\pix$b_{t}$ of previous actions \smash{$\{a_{t'}\}_{t'<t}$} and observation encodings \smash{$\{\omega(o_{t'})\}_{t'\leq t}$},
where $\omega$ is an encoding function;
\textbf{(ii.)} Target states are $\ell_{2}$-normalized encodings $x_{t+1}\coloneqq\tttext{sg}(\omega_{\text{target}}(o_{t+1})/\|\omega_{\text{target}}(o_{t+1})\|_{2})$ of future observations,
with $\omega_{\text{target}}$ being an exponential moving average of $\omega$, and $\tttext{sg}$ denotes the stop-gradient operator; and
\textbf{(iii.)} Predictions are $\ell_{2}$-normalized transformations of current beliefs and actions: $\hat{x}_{t+1}\coloneqq h_{\eta}(b_{t},a_{t})/\|h_{\eta}(b_{t},a_{t})\|_{2}$,
where $h_{\eta}$ is a prediction function.
Multi-step open-loop predictions are a straightforward extension. See Appendix \ref{app:c} for a more detailed review of the \tttext{BYOL-Explore} algorithm.

\dayum{
\textbf{Stochastic Traps}~
In stochastic environments, Equation \ref{eqn:original}
does not converge to zero even with infinite experience: It
converges to the entropy $\mathbb{H}[X_{t+1}|x_{t},a_{t}]$,
so the agent may become stuck on repeatedly experiencing (intrinsically rewarding) transitions where entropy is high.
Instead, what we desire is a reward that converges to zero in the limit.
The notion of ``optimistic'' exploration offers a hint of what might be possible---Consider constructing a reward that satisfies:
}\vspace{-0.5em}
\begin{align}
\mathcal{R}_{\eta}(x_{t},a_{t})
&
\geq
D_{\text{KL}}
\big(
\tau(X_{t+1}|x_{t},a_{t})\|\tau_{\eta}(X_{t+1}|x_{t},a_{t})
\big)
\label{eqn:optimism}
\end{align}
\dayum{%
which upper bounds the distance between the world and the agent's model. On the one hand, while Definition~\ref{def:curiosity} verifies this, the bound fails to tighten even in the limit. On the other hand, it is hard to measure this distance directly, since the entropy term is by construction unknown. As it turns out, we shall later see that our proposed technique effectively gives a reward that verifies the inequality---and is tight in the limit.
}

\begin{table*}[t]\small
\newcolumntype{A}{>{          \arraybackslash}m{3.5 cm}}
\newcolumntype{B}{>{\centering\arraybackslash}m{1.7 cm}}
\newcolumntype{C}{>{\centering\arraybackslash}m{1.2 cm}}
\newcolumntype{D}{>{\centering\arraybackslash}m{1.2 cm}}
\newcolumntype{E}{>{\centering\arraybackslash}m{1.5 cm}}
\newcolumntype{F}{>{\centering\arraybackslash}m{2.5 cm}}
\newcolumntype{G}{>{\centering\arraybackslash}m{2.8 cm}}
\newcolumntype{H}{>{\centering\arraybackslash}m{1.7 cm}}
\newcolumntype{I}{>{\centering\arraybackslash}m{1.4 cm}}
\newcolumntype{J}{>{\centering\arraybackslash}m{1.5 cm}}
\setlength\tabcolsep{8pt}
\renewcommand{\arraystretch}{1.07}
\caption{\dayum{\textit{Relationship with Curiosity-driven Exploration}. Curiosity in Hindsight is a drop-in modification on top of any prediction error-based method, and is characterized by being robust to different noises, being dynamics aware, and being general to any representation space.}}
\vspace{-1em}
\label{tab:related}
\begin{center}
\begin{adjustbox}{max width=1.005\textwidth}
\begin{tabular}{A|BHF|CIJ|G}
\toprule
  \vspace{-0.5em}\makecell[l]{\textbf{Curiosity-driven}\\\textbf{Exploration Method}}
& \textit{Prediction Inputs}
& \textit{Prediction Target}
& \vspace{-0.5em}\makecell{\textit{Measure of}\\\textit{Learning}}
& \vspace{-0.5em}\makecell{Random\\Noise}
& {$X$-/$A$-Dep. Noise}
& {Dynamics Awareness}
& \vspace{-0.5em}\makecell{\textbf{Representation}\\\textbf{Space}}
\\
\midrule
  AE \cite{stadie2016incentivizing}
& $X_{t},A_{t}$ & $X_{t+1}$
& $\mathcal{L}_{\eta}^{\text{predict}}$
& \xm & \xm & \cm & reconstructive
\\
  ICM \cite{pathak2017curiosity}
& $X_{t},A_{t}$ & $X_{t+1}$
& $\mathcal{L}_{\eta}^{\text{predict}}$
& \cm & \xm & \cm & action predictive
\\
  EMI \cite{kim2019emi}
& $X_{t},A_{t}$ & $X_{t+1}$
& $\mathcal{L}_{\eta}^{\text{predict}}$
& \xm & \xm & \cm & MI-maximizing
\\
  RND \cite{burda2019exploration}
& $X_{t+1}$ & $f_{\text{random}}(X_{t+1})$
& $\mathcal{L}_{\eta}^{\text{predict}}$
& \cm & \cm & \xm & random projection
\\
  Dora \cite{choshen2018dora}
& $X_{t},A_{t}$ & const. zero
& $\mathcal{L}_{\eta}^{\text{predict}}$
& \cm & \cm & \xm & pixel space
\\
  AMA \cite{mavor2022stay}
& $X_{t},A_{t}$ & $X_{t+1}$
& $\mathcal{L}_{\eta}^{\text{predict}}-\text{Tr}(\hat{\Sigma}_{t+1})$
& \cm & \cm & \cm & pixel space
\\
  \texttt{BYOL-Explore} \cite{guo2022byol}
& $X_{t},A_{t}$ & $X_{t+1}$
& $\mathcal{L}_{\eta}^{\text{predict}}$
& \xm & \xm & \cm & bootstrapped
\\
\midrule
\makecell[l]{\textbf{Curiosity in Hindsight}\\(e.g. \texttt{BYOL-Hindsight})}
& \vspace{0.5em}$X_{t},A_{t},Z_{t+1}$ & \vspace{0.5em}$X_{t+1}$
& \vspace{0.5em}\hspace{-10px}$\mathcal{L}_{\theta,\eta}^{\text{reconstruct}}+\mathcal{L}_{\theta,\nu}^{\text{invariance}}$
& \vspace{0.5em}\cm & \vspace{0.5em}\cm & \vspace{0.5em}\cm & \vspace{0.5em}\textbf{\textit{any representation}}
\\
\bottomrule
\end{tabular}
\end{adjustbox}
\end{center}
\vspace{-1.5em}
\end{table*}

\vspace{-0.5em}
\subsection{Related Work}

\dayum{
Our work inherits from the curiosity-driven paradigm \cite{schmidhuber1991possibility,thrun1995exploration,barto2004intrinsically,oh2015action,finn2016unsupervised,gregor2019shaping,orseau2013universal,stadie2016incentivizing,pathak2017curiosity,burda2019large,burda2019exploration,hong2020adversarial,kim2019emi,guo2022byol,mavor2022stay,choshen2018dora}, among which some methods have been designed with robustness to certain stochasticities in mind (Table \ref{tab:related}). However, our method is uniquely characterized by the following:
}
\vspace{-0.75em}
\begin{enumerate}[leftmargin=1.15em,labelsep=0.45em]
\itemsep1.5pt
\item \dayum{\textbf{Stochasticity Types}: First, it is capable of handling all types of stochasticities in generality. Specifically, this includes stochasticity that is \textit{entirely random} (e.g. a viewport polluted by noise sampled according to a distribution independent of states and actions), stochasticity that is \textit{state-dependent} (e.g. a visible object that performs a random walk within the environment), as well as \textit{action-dependent} (e.g. a layer of random pixels that only appears if sampled on demand by specific actions). For instance, previous works have found that inverse dynamics features can learn to filter out random noise \cite{pathak2017curiosity}, but may break down in the presence of action-dependent noise \cite{burda2019large,pathak2019self}.}
\item \dayum{\textbf{Dynamics Awareness}: Second, it does not require entirely discarding dynamics learning. By way of contrast, consider purely frequency-oriented exploration strategies, such as learning to predict a random projection of observations \cite{burda2019exploration}, or simply to predict the constant zero \cite{choshen2018dora}. As these are deterministic functions of their inputs, they are in principle resilient to stochasticity. But empirically they can still behave poorly in the presence of action-dependent stochasticities \cite{mavor2022stay}: If the noise is sufficiently \textit{diffuse}, the agent may never learn the function well,
so in the absence of any other learning signal---such as the dynamics of the world---they may still become stuck \cite{kim2020active}.}
\item \dayum{\textbf{Generality and Scalability}: As a drop-in modification, it is generally be applicable to any underlying choice of representation space.
In contrast, existing techniques ca-pable of handling stochasticity are often tied to specific feature spaces, such as to employ inverse dynamics features \cite{pathak2017curiosity}, random features \cite{burda2019exploration}, or pixel-space features \cite{mavor2022stay}---which may limit their flexibility of application.
Moreover, unlike ensemble-based or disagreement-based techniques that require training a large number of models \cite{pathak2019self,henaff2019explicit,shyam2019model},\footnote{\dayum{Ensembles can in principle approximate uncertainty \cite{pathak2019self,henaff2019explicit,osband2016deep,shyam2019model}; in practice, training and scaling is difficult with larger architectures, and models often converge prematurely to the same outputs
\cite{osband2018randomized}.
}\vspace{-1em}} we shall see that incorporating hindsight is simpler and more \textit{scalable} by only requiring the addition of an auxiliary component to the usual prediction loss.}
\end{enumerate}
\vspace{-0.75em}
\dayum{
Alternative paradigms for exploration have been proposed:
Novelty-based methods encourage exploration on the basis of visitation counts \cite{strehl2008analysis}, hashes \cite{tang2017exploration}, density estimates \cite{bellemare2016unifying,ostrovski2017count,zhao2019curiosity,domingues2021density}, and adversarial guidance \cite{fu2017ex2,flet2021adversarially}; further extensions account for episodic memory \cite{savinov2019episodic,badia2020never,badia2020agent57} and the long-term value of exploratory actions \cite{choshen2018dora,machado2018count,oh2018directed,machado2020count}. Knowledge-based methods encourage exploration on the basis of the agent's uncertainty about the world \cite{cohn1996active,pathak2019self}, with most work focusing on estimating the information gain from different actions \cite{itti2009bayesian,araya2010pomdp,sun2011planning,still2012information,houthooft2016vime,henaff2019explicit,shyam2019model,sekar2020planning,mendonca2021discovering}, or directly estimating learning progress \cite{schmidhuber1991curious,oudeyer2007intrinsic,azar2019world}. Finally, diversity-based methods seek to maximize the state entropy \cite{hazan2019provably,liu2021behavior,guo2021geometric,yarats2021reinforcement}, or to encourage learning diverse skills \cite{gregor2017variational,achiam2018variational,eysenbach2019diversity,lee2019efficient,campos2020explore,sharma2020dynamics,baumli2021relative,groth2021curiosity,kwon2021variational,liu2021aps,eysenbach2022information,laskin2022cic} and reaching different goals \cite{andrychowicz2017hindsight,florensa2018automatic,nair2018visual,fang2019curriculum,zhang2020automatic}.
}
\section{Curiosity in Hindsight}\label{sec:3}

\dayum{
Consider the game of betting on a hidden dice roll: Suppose we take the action $A_{t}$\pix$=$ ``bet on 6'', then observe the outcome $X_{t+1}$\pix$=$ ``lost the bet''. Two facts are clear: (1) \textit{a priori}, we could not have predicted this result at all; (2) \textit{a posteriori}, we may deduce the (latent) fact $Z_{t+1}$\pix$=$ ``the die must have rolled 1--5''.
These are not contradictory. In particular, the former does \textit{not} imply that we lack an understanding of how the game works, nor does it suggest that we should engage in further such bets to improve our understanding.
Indeed, knowing how the game works, in hindsight (i.e. given what we deduced about $Z_{t+1}$), the outcome is obvious to us (i.e. we can now deterministically identify $X_{t+1}$).
Conversely, suppose we actually \textit{didn't} know how the game works: Then we couldn't have correctly inferred $Z_{t+1}$, nor would its knowledge have enabled us to identify $X_{t+1}$ with any certainty. If so, engaging in additional bets may indeed allow us to learn and improve our understanding of how they work.
}

\dayum{
Intuitively, we can thus measure our understanding of each transition based on how much the outcome makes sense \textit{in hindsight}.
So, instead of asking ``How well can we \muline{predict} $X_{t+1}$ \textit{a priori}?'', we actually want to ask ``How well can we \muline{reconstruct} $X_{t+1}$ \textit{a posteriori}---i.e. given hindsight $Z_{t+1}$?''.
First, we formalize this intuition using the language of \textit{posterior inference} when a known model of the world is available (Section \ref{subsec:31}). Then we generalize this approach to generating learned \textit{hindsight representations} when the world model needs to be learned at the same time (Section \ref{subsec:32}). Finally, we derive \textit{Curiosity in Hindsight} on the basis of these ingredients, showing it approximates optimistic exploration (Inequality \ref{eqn:optimism}) while being robust to stochasticities (Section \ref{subsec:33}).
}

\begin{figure}[t]
\caption{\small \dayum{\textit{Structural Causal Model}. By the reparameterization lemma, there exists an equivalent graphical representation under which all stochasticities are exogenous (i.e. dotted edges removed).}}
\label{fig:modelbased}
\centering
\vspace{0.25em}
~\includegraphics[width=\linewidth, trim=0em 0em 0em 0em]{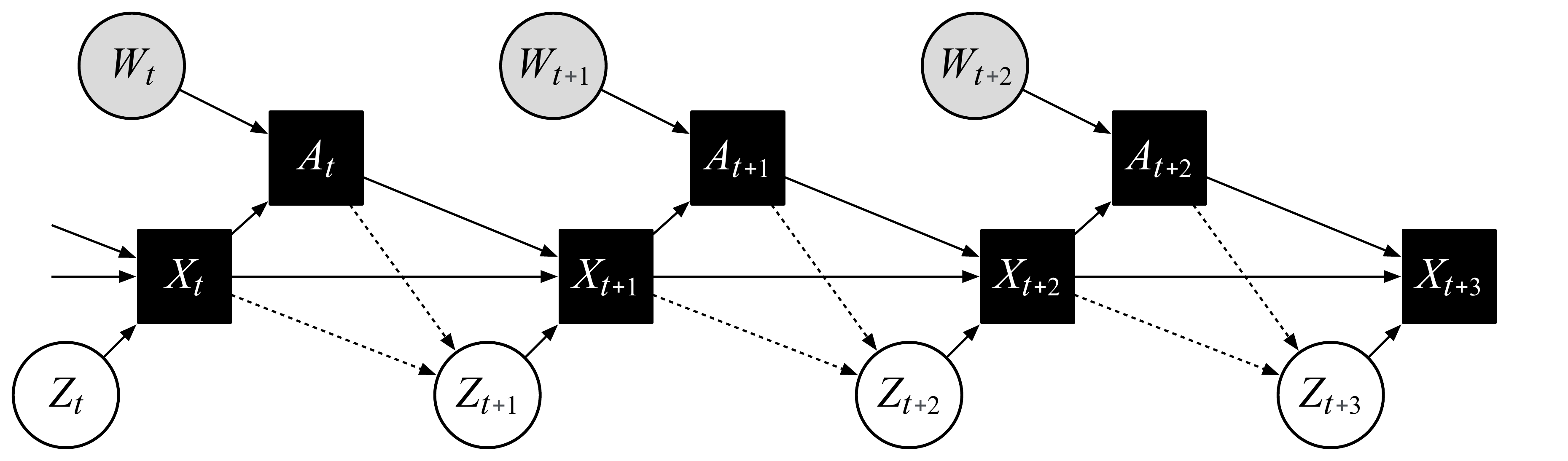}
\vspace{-2.75em}
\end{figure}

\vspace{-0.5em}
\subsection{Structural Causal Model}\label{subsec:31}

\dayum{
Let $Z$ denote a \textit{latent} variable, taking on values $z\in\mathcal{Z}$.
For each observed transition $(x_{t},a_{t},x_{t+1})$, we let $z_{t+1}$ encapsulate \textit{all} sources of unobserved stochasticity in the dynamics. By construction, $x_{t+1}=f(x_{t},a_{t},z_{t+1})$ for some deterministic function $f$, and 
a prior $p$ over $Z_{t+1}$ induces the environment dynamics---that is, the distribution $\tau(X_{t+1}|x_{t},a_{t})$.
Figure \ref{fig:modelbased} illustrates the structural causal model, with solid squares for deterministic nodes, shaded circles for observable stochastic nodes, and unshaded circles for unobservable stochastic nodes ($W$ captures any randomness in the policy).
In general, stochasticities can be entirely random (i.e. no edges into $Z_{t+1}$), state-dependent (i.e. edge $X_{t}\rightarrow Z_{t+1}$), or action-dependent (i.e. edge $A_{t}\rightarrow Z_{t+1}$).
However, by the \textit{reparameterization lemma} it is always possible to represent an environment such that all stochasticities are effectively exogenous \cite{buesing2018woulda,oberst2019counterfactual,lorberbom2021learning} (i.e. no directed edges into $Z_{t+1}$).
}

\dayum{
\textbf{From Prediction to Reconstruction}~
Consider the setting in which we know the model $f$. Suppose first that we somehow had access to each latent $z_{t+1}$. Then the outcome of a transition at state $x_{t}$ and action $a_{t}$ would be deterministically computable with no uncertainty ({\small i.e.} reconstruction error $=$ zero):
}\vspace{-0.75em}
\begin{equation}
x_{t+1}
\equiv
f(x_{t},a_{t},z_{t+1})
\end{equation}

\vspace{-0.75em}
\dayum{%
In reality, the latent variable $z_{t+1}$ is not observed. Thus it may seem like the best we can do is to compute the \textit{a~priori} expectation of the outcome ({\small i.e.} prediction error $=$ entropy):
}\vspace{-0.75em}
\begin{equation}
\mathbb{E}_{X_{t+1}\sim \tau(\cdot|x_{t},a_{t})} X_{t+1}
\equiv
\mathbb{E}_{Z_{t+1}\sim p} f(x_{t},a_{t},Z_{t+1})
\end{equation}

\vspace{-0.75em}
\dayum{%
However, while $z_{t+1}$ is not observable, based on the transition $(x_{t},a_{t},x_{t+1})$ we can infer \textit{a~posteriori} what its values could have been. Importantly, by the consistency property of counterfactuals we know $f(x_{t},a_{t},Z_{t+1})$\pix$=$\pix$x_{t+1}$ for any $Z_{t+1}$\pix$\sim$\pix$p(\cdot|x_{t},a_{t},x_{t+1})$ \cite{pearl2009causality}. That is to say, conditioned on hindsight, the reconstruction error of the true model is zero. This suggests when $f$ is unknown and learned by the agent, \textit{the reconstruction error may be an attractive candidate for an intrinsic reward}.
Of course, now the missing piece is how to sample $Z_{t+1}$ from the posterior---which we discuss next.
}

\begin{figure}[t]
\caption{\small \dayum{\textit{Hindsight Representations}. A learned generator $G_{t+1}\coloneqq p_{\theta}(\cdot|x_{t},a_{t},x_{t+1})$ generates hindsight vectors---denoted $Z_{t+1}^{*}$ in this figure to be distinguished from ``ground-truth'' latents $Z_{t+1}$.}}
\label{fig:modelfree}
\centering
\vspace{0.25em}
\includegraphics[width=\linewidth, trim=0em 0em 0em 0em]{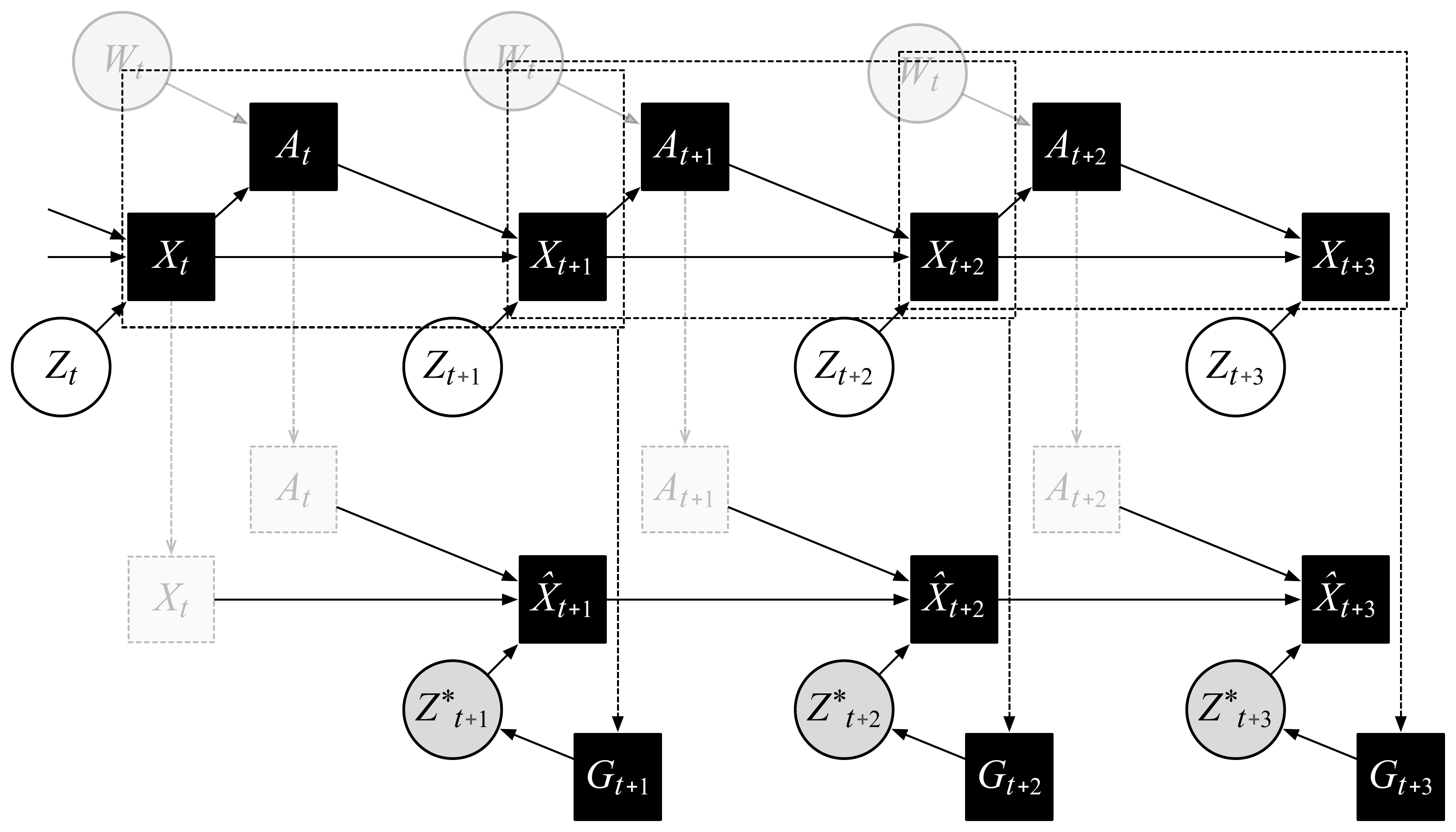}
\vspace{-2.55em}
\end{figure}

\vspace{-0.5em}
\subsection{Hindsight Representations}\label{subsec:32}

\dayum{
Realistically, the model $f(X_{t},A_{t},Z_{t+1})$ is unknown, so we learn to approximate it using a \textit{reconstructor} $f_{\eta}$ parameteriz-ed by $\eta$.\footnote{In general the model is not identifiable and there is no guarantee that
$f_{\eta}(x_{t},a_{t},Z_{t+1})$ be close to $x_{t+1}$ for arbitrary $Z_{t+1}$.
But we are not interested in making counterfactual queries: For reconstruction, we only wish to evaluate $f_{\eta}$ where
$Z_{t+1}$\pix$\sim$\pix$p_{\eta}(\cdot|x_{t},a_{t},x_{t+1})$.}
Exact posterior inference $p_{\eta}(Z_{t+1}|X_{t},A_{t},X_{t+1})$ is intractable, so we learn to approximate it using a \textit{generator} $p_{\theta}$ parameterized by $\theta$.
Two objectives are key. First, as noted above, representations $Z_{t+1}$ should be \textit{reconstructive} of outcomes $X_{t+1}$.
Here we can simply use the squared loss:
}

\begin{reobjective}[restate=objrec,name=Reconstruction]\upshape\label{obj:rec}
\dayum{
~Let the \textit{reconstruction loss} for a given transition $(x_{t},a_{t},z_{t+1},x_{t+1})$---including hinds-ight representation $z_{t+1}$ drawn from $p_{\theta}(\cdot|x_{t},a_{t},x_{t+1})$---be:
}\vspace{-0.9em}
\begin{align}
\hspace{-2px}
\mathcal{L}_{\eta}^{\text{rec.}}(x_{t},a_{t},z_{t+1},x_{t+1})
\coloneqq
\big\|
x_{t+1}
\hspace{-2px}
-
\hspace{-2px}
f_{\eta}(x_{t},a_{t},z_{t+1})
\big\|_{2}^{2}
\end{align}

\vspace{-0.75em}
and (state-action) \textit{reconstruction bonus} for the agent policy:
\vspace{-2.5em}

\begin{align}
\hspace{-2px}
\mathcal{R}_{\theta,\eta}^{\text{rec.}}&(x_{t},a_{t})
\coloneqq
\nonumber
\\[-3px]
&
~~~
\mathbb{E}_{\substack{X_{t+1}\sim\tau(\cdot|x_{t},a_{t})\\Z_{t+1}\sim p_{\theta}(\cdot|x_{t},a_{t},X_{t+1})}}
\mathcal{L}_{\eta}^{\text{rec.}}(x_{t},a_{t},Z_{t+1},X_{t+1})
\end{align}
\end{reobjective}
\vspace{-1em}

\dayum{
Driven to zero, this requires hindsight representations to encapsulate \textit{at least} all aspects of the world's dynamics that are unpredictable (so that we \textit{don't} reward the agent for irreducible error).
However, we also don't want $Z_{t+1}$ to simply leak information about the outcome that is actually predictable to begin with (so that we \textit{do} reward the agent for reducible error).\footnote{For example, consider the solution $Z_{t+1}\coloneqq X_{t+1}$, where reconstruction error is trivially zero, but is pathological for exploration.}
Thus our second objective requires it to be \textit{independent} of $X_{t},A_{t}$.
Denote the pointwise mutual information between state-action $x_{t},a_{t}$ and hindsight $z_{t}$ by
\smash{$
\text{PMI}_{\theta}(x_{t},a_{t};z_{t+1})
$$\coloneqq$$
\log(p_{\theta}(z_{t+1}|x_{t},a_{t})/p_{\theta}(z_{t+1}))
$}.
Then:
}

\begin{reobjective}[restate=objinv,name=Invariance]\upshape\label{obj:inv}
\dayum{
Let the \textit{invariance loss} for a giv-en transition $(x_{t},a_{t},z_{t+1},x_{t+1})$---again, where the hindsig-ht representation $z_{t+1}$ is drawn from $p_{\theta}(\cdot|x_{t},a_{t},x_{t+1})$---be:
}\vspace{-0.75em}
\begin{align}
\mathcal{L}_{\theta}^{\text{inv.}}(x_{t},a_{t},z_{t+1})
\coloneqq
\text{PMI}_{\theta}(x_{t},a_{t};z_{t+1})
~~~~
\end{align}

\vspace{-0.75em}
and (state-action) \textit{invariance bonus} for the agent's policy:
\vspace{-1.25em}

\begin{align}
~~~~~~~~~~
&\mathcal{R}_{\theta}^{\text{inv.}}(x_{t},a_{t})
\coloneqq
\nonumber
\\[-3px]
&
~~~~~~
\mathbb{E}_{\substack{X_{t+1}\sim\tau(\cdot|x_{t},a_{t})\\Z_{t+1}\sim p_{\theta}(\cdot|x_{t},a_{t},X_{t+1})}}
\mathcal{L}_{\theta}^{\text{inv.}}(x_{t},a_{t},Z_{t+1})
~~
\end{align}
\end{reobjective}
\vspace{-1em}

\dayum{
Driven to zero, this requires hindsight representations to encapsulate \textit{at most} all aspects of the world's dynamics that are unpredictable.
This suggests a combination---of reconstruction loss (of a dynamics model) plus invariance loss (of a hindsight model)---may make a good signal for exploration.
We now have all the ingredients required for Curiosity in Hi-ndsight, which it is instructive to contrast with Definition \ref{def:curiosity}:
}

\subsection{Optimistic Exploration}\label{subsec:33}

\begin{redefinition}[restate=defhindsight,name=Curiosity in Hindsight]\upshape\label{def:hindsight}
\dayum{
~Let the \textit{hindsight intrinsic reward function} be defined as the weighted combination of Objectives \ref{obj:rec} and \ref{obj:inv}, with a tradeoff coefficient $\lambda$:
}\vspace{-1em}
\begin{align}
\mathcal{R}_{\theta,\eta}(x_{t},a_{t})
\coloneqq
\pix
\frac{1}{\lambda}
\pix
\mathcal{R}_{\theta,\eta}^{\text{rec.}}(x_{t},a_{t})
+
\mathcal{R}_{\theta}^{\text{inv.}}(x_{t},a_{t})
\label{eqn:reward}
\end{align}
\vspace{-1.75em}

and the dynamics and hindsight models are jointly trained to minimize this quantity over the trajectories it collects, while rolling out a policy seeking to maximize this same quantity:\vspace{-0.4em}
\begin{align}
\overset{\text{(policy)}}{
\underset{\pi}{\text{maximize}}
}
~~
\overset{\text{(model)}}{
\underset{\theta,\eta}{\text{min}}
}
~~
\mathbb{E}_{\substack{X_{t}\sim\rho_{\pi}\\A_{t}\sim\pi(\cdot|X_{t})}}
\mathcal{R}_{\theta,\eta}(X_{t},A_{t})
\label{eqn:curiosity}
\end{align}
\end{redefinition}
\vspace{-1em}

\dayum{
Recall that in the presence of stochastic transitions, standard curiosity-driven exploration can be seen as a poor approximation to ``optimistic'' exploration (Inequality~\ref{eqn:optimism})---because the bound is never tight even in the limit, which renders it susceptible to stochastic traps. The following result shows that exploration by Curiosity in Hindsight can resolve this:
}

\begin{retheorem}[restate=thmoverall,name=Optimistic Exploration]\upshape\label{thm:overall}
\dayum{
~Let the coefficient $\lambda$ satisfy the inequality
\smash{$
\frac{1}{2}\log(\lambda\pi)
\leq
$}
\smash{$
\mathbb{H}_{\theta}
[X_{t+1}|x_{t},a_{t},Z_{t+1}]
$}
\smash{$
+
D_{\text{KL}}
\big(
p_{\theta}(Z_{t+1}|x_{t},a_{t})\|p_{\theta}(Z_{t+1})
\big)
$},
where $\pi$ denotes here the mathematical constant (not the agent's policy). Then:
}\vspace{-0.5em}
\begin{align}
\hspace{-5pt}
\scalebox{0.95}{$
\mathcal{R}_{\theta,\eta}(x_{t},a_{t})
\geq
D_{\text{KL}}
\big(
\tau(X_{t+1}|x_{t},a_{t})\|\tau_{\theta,\eta}(X_{t+1}|x_{t},a_{t})
\big)
$}\hspace{-2pt}
\end{align}

\vspace{-0.5em}
\dayum{
where
\smash{$\tau_{\theta,\eta}(X_{t+1}|x_{t},a_{t})$\pix$\coloneqq$\pix$\mathbb{E}_{Z_{t+1}\sim p_{\theta}}p_{\eta}(X_{t+1}|x_{t},a_{t},Z_{t+1})$}
denotes the learned world model.
Furthermore, assuming realizability, rewards vanish at optimal parameters \smash{$\theta^{*},\eta^{*}$}:
}\vspace{-0.75em}
\begin{align}
\mathcal{R}_{\theta^{*},\eta^{*}}(x_{t},a_{t})=0
\quad
\forall x_{t},a_{t}\in\text{supp}(\rho_{\pi})
\end{align}
\end{retheorem}
\vspace{-0.9em}
\textit{Proof}. Appendix \ref{app:a}. \QED

In other words, by choosing a small enough $\lambda$ term, the hindsight intrinsic reward (Equation \ref{eqn:reward}) is an upper bound on the KL-term we care about (Inequality \ref{eqn:optimism}).
Driving the reward to zero also drives the KL to zero, thus the reward-maximizing exploration policy (Equation \ref{eqn:curiosity}) is approximating precisely the sort of ``optimistic'' exploration that we desired to start.
\vspace{-0.65em}
\section{Practical Framework}\label{sec:4}

Two questions remain. Firstly, how should the invariance terms be computed? For this, we propose a contrastive learning framework to approximate them (Section \ref{subsec:41}). Secondly, what does a concrete implementation look like? For this, we instantiate this framework on top of \tttext{BYOL-Explore}, yielding its robust variant \tttext{BYOL-Hindsight} (Section \ref{subsec:42}).

\vspace{-0.5em}
\subsection{Contrastive Learning}\label{subsec:41}

To estimate the pointwise mutual information, we use an auxiliary \textit{critic} $g_{\nu}$ parameterized by $\nu$, trained as maximizer:

\vspace{-0.25em}
\begin{reobjective}[restate=objcon,name=Contrastive Learning]\upshape\label{obj:con}
\dayum{
~Let the \textit{contrastive loss} for a transition, with respect to a batch of $K$$-$$1$ negative hindsight samples \smash{$z_{t+1}^{1:K-1}\coloneqq z_{t+1}^{1},...,z_{t+1}^{K-1}$}, be defined as:
}
\end{reobjective}

\begin{align}
&\ell_{\theta,\nu}^{K,\text{con.}}(x_{t},a_{t},z_{t+1},z_{t+1}^{1:K-1})
\coloneqq
\\[-1px]
\nonumber
&
~~~
\log
\frac
{e^{g_{\nu}(x_{t},a_{t},z_{t+1})}}
{\frac{1}{K}\left(e^{g_{\nu}(x_{t},a_{t},z_{t+1})}+\sum_{i=1}^{K-1}e^{g_{\nu}(x_{t},a_{t},Z_{t+1}^{i})}\right)}
~~~~
\end{align}

\vspace{-1em}
\dayum{
such that the overall contrastive loss for the transition is its expectation over negative hindsight samples from rollouts:
}\vspace{-1em}
\begin{gather}
\begin{aligned}
&
~~~~
\mathcal{L}_{\theta,\nu}^{K,\text{con.}}(x_{t},a_{t},z_{t+1})
\coloneqq
\\[-1px]
&
~~~~~~~~~~
\mathbb{E}
\hspace{-15pt}
\scalebox{0.85}{$
_{\substack{
(X_{t}^{1},...,X_{t}^{K-1})\sim\prod_{i=1}^{K-1}\rho_{\pi}
\\
(A_{t}^{1},...,A_{t}^{K-1})\sim\prod_{i=1}^{K-1}\pi(\cdot|X_{t}^{i})
\\
(X_{t+1}^{1},...,X_{t+1}^{K-1})\sim\prod_{i=1}^{K-1}\tau(\cdot|X_{t}^{i},A_{t}^{i})
\\
(Z_{t+1}^{1},...,Z_{t+1}^{K-1})\sim\prod_{i=1}^{K-1}p_{\theta}(\cdot|X_{t}^{i},A_{t}^{i},X_{t+1}^{i})
}}
$}
\hspace{-20pt}
\ell_{\theta,\nu}^{K,\text{con.}}(x_{t},a_{t},z_{t+1},z_{t+1}^{1:K-1})
\end{aligned}
\raisetag{0.8\baselineskip}
\end{gather}

\vspace{-1em}
and (state-action) \textit{contrastive bonus} for the agent's policy:
\vspace{-1em}

\begin{align}
~
\mathcal{R}_{\theta,\nu}^{K,\text{con.}}&(x_{t},a_{t})
\coloneqq
\nonumber
\\[-3px]
&
\mathbb{E}_{\substack{X_{t+1}\sim\tau(\cdot|x_{t},a_{t})\\Z_{t+1}\sim p_{\theta}(\cdot|x_{t},a_{t},X_{t+1})}}
\mathcal{L}_{\theta,\nu}^{K,\text{con.}}(x_{t},a_{t},Z_{t+1})
\end{align}
\vspace{-1.25em}

\dayum{
How does Objective \ref{obj:con} approximate Objective \ref{obj:inv}? Precisely:
}
\vspace{-0.25em}

\begin{retheorem}[restate=thminv,name=Optimal Invariance]\upshape\label{thm:inv}
The contrastive bonus lower-bounds the (ideal) invariance bonus for any pair $x_{t},a_{t}$:
\vspace{-1.25em}
\begin{align}
\mathcal{R}_{\theta,\nu}^{K,\text{con.}}(x_{t},a_{t})
\leq
\mathcal{R}_{\theta}^{\text{inv.}}(x_{t},a_{t})
\end{align}

\vspace{-0.75em}
\dayum{%
Furthermore, assuming realizability, for optimal critic parameter $\nu_{K}^{*}
\coloneqq
\text{arg}\pix\text{max}_{\nu}~
\mathbb{E}
_{X_{t},A_{t}\sim\rho_{\pi}}
\mathcal{R}_{\theta,\nu}^{K,\text{con.}}(X_{t},A_{t})
$ the bound is asymptotically tight (in the batch size $K\rightarrow\infty$):}\vspace{-0.75em}
\begin{align}
\lim_{K\rightarrow\infty}
\mathcal{R}_{\theta,\nu_{K}^{*}}^{K,\text{con.}}&(x_{t},a_{t})
=
\mathcal{R}_{\theta}^{\text{inv.}}(x_{t},a_{t})
\end{align}
\end{retheorem}
\vspace{-0.5em}
\textit{Proof}. Appendix \ref{app:a}. \QED

\textbf{Practical Algorithm}~
\dayum{
This suggests a straightforward algorithm. In practice, batch size $K<\infty$ and critic $\nu$ is not fully optimized. The intrinsic reward (Definition \ref{def:hindsight}) now becomes:
}\vspace{-1.25em}

\begin{align}
\mathcal{R}_{\theta,\eta,\nu}^{K}(x_{t},a_{t})
&
\coloneqq
\pix
\frac{1}{\lambda}
\pix
\mathcal{R}_{\theta,\eta}^{\text{rec.}}(x_{t},a_{t})
+
\mathcal{R}_{\theta,\nu}^{K,\text{con.}}(x_{t},a_{t})
\end{align}

\vspace{-1em}
and optimization alternates between training the critic (maximizer) and dynamics and hindsight models (minimizers):
\vspace{-2em}

\begin{align}
\overset{\text{(policy)}}{
\underset{\pi}{\text{maximize}}
}
~~
\overset{\text{(model)}}{
\underset{\theta,\eta}{\text{min}}~
\underset{\nu}{\text{max}}
}
~~
\mathbb{E}_{\substack{X_{t}\sim\rho_{\pi}\\A_{t}\sim\pi(\cdot|X_{t})}}
\mathcal{R}_{\theta,\eta,\nu}^{K}(X_{t},A_{t})
\end{align}

\vspace{-1.1em}
\dayum{
Overall, our framework constitutes a simple drop-in modification on top of any curiosity-driven method: Instead of learning a \textit{predictive model} specifying $X_{t+1}$\pix$\sim$\pix$\tau_{\eta}(\cdot|X_{t},A_{t})$, we now learn a (hindsight-augmented) \textit{reconstructive model} specifying $X_{t+1}$\pix$=$\pix$f_{\eta}(X_{t},A_{t},Z_{t+1})$.
The main ingredients include the reconstructor $f_{\eta}(X_{t},A_{t},Z_{t+1})$, the generator $p_{\theta}(Z_{t+1}|X_{t},A_{t},X_{t+1})$, and the critic $g_{\nu}(X_{t},A_{t},Z_{t+1})$, and the main hyperparameters are the contrastive batch size $K$ and the coefficient $\lambda$ in the hindsight intrinsic reward.
}

\dayum{
Finally, note that for ease of exposition we have focused on modeling single-step transitions; however, it is straightforward to generalize this approach to the case of multi-step reconstruction horizons using open-loop rollouts (Figure \ref{fig:modelfree}).
}

\begin{figure}[t]
\centering
\vspace{-0.6em}
\caption{\small \dayum{\textit{From BYOL-Explore to BYOL-Hindsight}. The latter simply replaces the prediction loss with our reconstruction and contrastive losses (through the addition of hindsight), with no change to the underlying (bootstrapped) representation learning method.}
}
\vspace{0.15em}
\includegraphics[width=\linewidth, trim=6em 0em 3em 6em, clip]{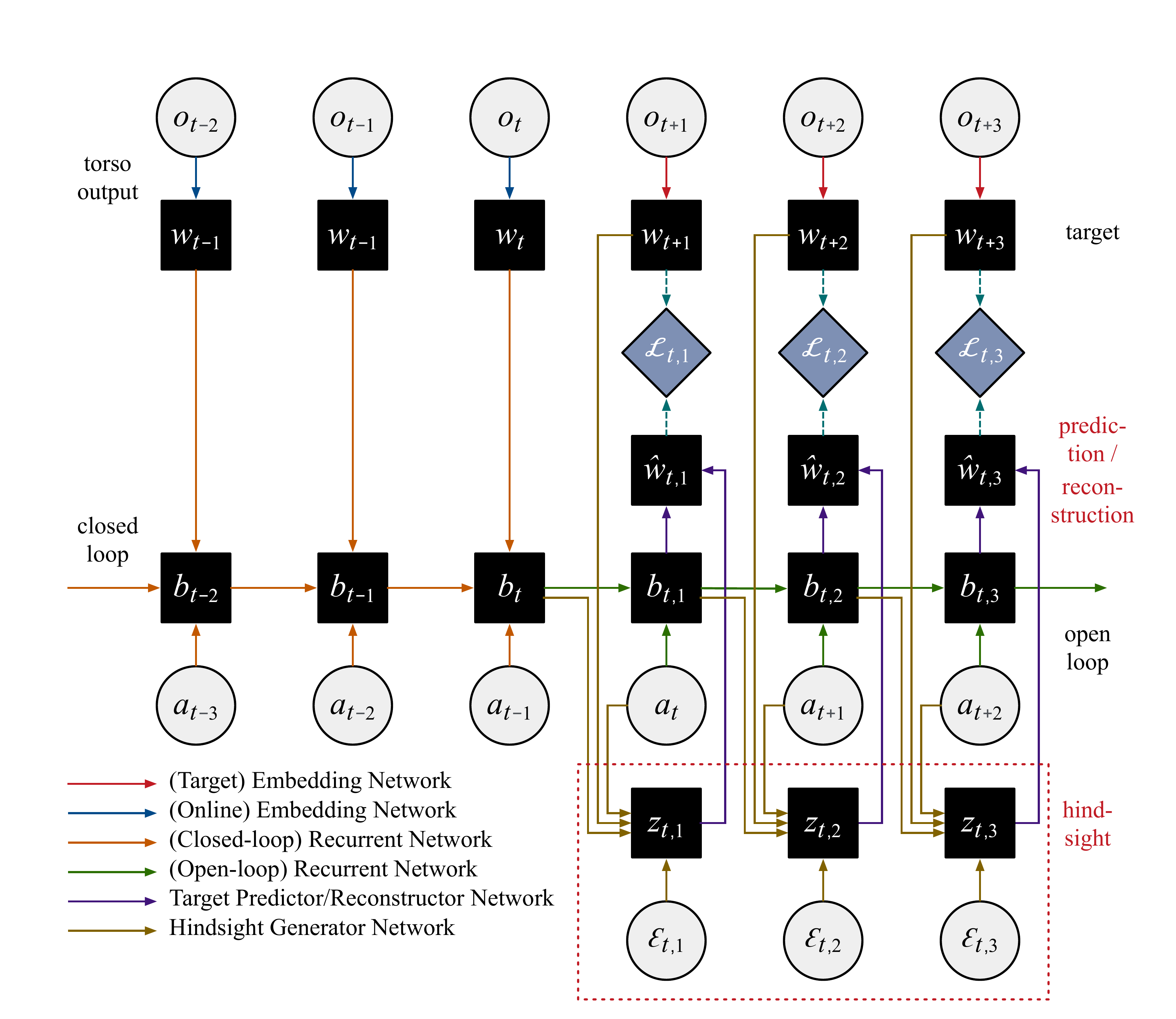}
\vspace{-3.75em}
\label{fig:byolcombine}
\end{figure}

\subsection{BYOL-Hindsight}\label{subsec:42}

\dayum{
The preceding discussion used MDP notation, but in practice input states are often representations of histories, and obser-vations are often represented in latent space.
Recall \tttext{BYOL-} \tttext{Explore} (Example \ref{exa:byolexplore}) is an incarnation of curiosity (Definition \ref{def:curiosity}) that learns such representations.
We now augment it with hindsight, giving rise to the novel \tttext{BYOL-Hindsight}:
}

\vspace{-0.5em}
\begin{reexample}[restate=exabyolhindsight,name=Bootstrapping with Hindsight]\upshape\label{exa:byolhindsight}
Let the \textit{hindsight loss} for a transition $(x_{t},a_{t},z_{t+1},x_{t+1})$ be defined as: \vspace{-2em}

\begin{align}
&
\mathcal{L}_{\theta,\eta,\nu}^{K,{\tiny~\substack{\tttext{BYOL-}\\[-1px]\tttext{Hind.}}}}(x_{t},a_{t},z_{t+1},x_{t+1})
\coloneqq
\\[-1px]
\nonumber
&
~~~~~~~
\pix
\frac{1}{\lambda}
\pix
\mathcal{L}_{\eta}^{\text{rec.}}(x_{t},a_{t},z_{t+1},x_{t+1})
+
\mathcal{L}_{\theta,\nu}^{K,\text{con.}}(x_{t},a_{t},z_{t+1})
\end{align}

\vspace{-1em}
and the (state-action) \textit{hindsight bonus} for the agent's policy:
\vspace{-2em}

\begin{align}
&
\mathcal{R}_{\theta,\eta,\nu}^{K,{\tiny~\substack{\tttext{BYOL-}\\[-1px]\tttext{Hind.}}}}(x_{t},a_{t})
\coloneqq
\\[-2px]
\nonumber
&
~~~~~
\mathbb{E}_{\substack{X_{t+1}\sim\tau(\cdot|x_{t},a_{t})\\Z_{t+1}\sim p_{\theta}(\cdot|x_{t},a_{t},X_{t+1})}}
\mathcal{L}_{\theta,\eta,\nu}^{K,{\tiny~\substack{\tttext{BYOL-}\\[-1px]\tttext{Hind.}}}}(x_{t},a_{t},Z_{t+1},X_{t+1})
\end{align}
\end{reexample}
\vspace{-0.75em}

where the key difference from \tttext{BYOL-Explore} lies in swapping out predictions for reconstructions. Specifically, input states $x_{t}$ and target states $x_{t+1}$ are defined just as before:
\textbf{(i.)} Input states are RNN ``belief'' representations; and
\textbf{(ii.)} Target states are $\ell_{2}$-normalized encodings of future observations.
However, instead of (target) predictions, we now have (target) reconstructions:
\textbf{(iii.)} Reconstructions are $\ell_{2}$-normalized transformations of beliefs, actions, and hindsi-ght:
$f_{\eta}(b_{t},a_{t},z_{t+1})$$\coloneqq$$h_{\eta}(b_{t},a_{t},z_{t+1})/\|h_{\eta}(b_{t},a_{t},z_{t+1})\|_{2}$,
where $h_{\eta}$ is a reconstruction function, and the contrastive loss encourages hindsight $Z_{t+1}$ to be independent of $B_{t},A_{t}$.

\dayum{
Figure \ref{fig:byolcombine} gives the concrete architecture for \tttext{BYOL-Explore}/ \tttext{BYOL-Hindsight}, with the full multi-step horizon setup:
First, an \textit{online} embedding network $\omega$ encodes observations $o_{t}$ into representations $w_{t}=\omega(o_{t})$.
A \textit{closed-loop} RNN then computes representations $b_{t}$ of histories up until each time step $t$.
This is used to initialize an \textit{open-loop} RNN that computes representations $b_{t,i}$ for horizon steps indexed as~$i$.
These representations are fed to a predictor/reconstructor network to output \smash{$\hat{w}_{t,i}$}.\footnote{The composition of open-loop RNN and predictor/reconstructor networks is what we have abstractly denoted $h_{\eta}$ in Examples \ref{exa:byolexplore}--\ref{exa:byolhindsight}.}
Finally, prediction/reconstruction targets are encoded with a \textit{target} embedding network that is an exponential moving average of the online network.
The prediction/reconstruction error $\mathcal{R}_{t,i}$ at each open-loop step is computed, and the intrinsic reward associated to each observed transition ($o_{s},a_{s},o_{s+1}$) is the sum of errors $\sum_{t+i=s+1}\mathcal{R}_{t,i}$.
In \tttext{BYOL-Hindsight}, the generator samples $Z_{t,i}$ by taking noise $\varepsilon_{t,i}$ as input, and an additional critic (not pictured) encourages $Z_{t,i}$ to be independent of $B_{t,i-1}$ and $A_{t+i-1}$.
See Algorithms \ref{alg:byole}--\ref{alg:byolh} in Appendix \ref{app:c} for details.
}
\vspace{-0.65em}
\section{Experiments}\label{sec:5}

\dayum{
Three questions deserve empirical study:
\textbf{(a.) Effectiveness}:
In stochastic environments, predictive error-based methods ---e.g. \tttext{BYOL-Explore}---may fail. Does \tttext{BYOL-Hindsight} address the problem?
\textbf{(b.) Robustness}:
Is the method robust to the different types of stochasticities---i.e. independent noise, state-dependent noise, and action-dependent noise?
\textbf{(c.) Non-specificity}: In environments with no stochasticity, hindsight should confer no benefit. Does \tttext{BYOL-Hindsight} manage to preserve the performance of \tttext{BYOL-Explore}?
}

\begin{figure}[b]
\vspace{-0.75em}
\begin{minipage}{.64\columnwidth}
{\transparent{0.85}
\includegraphics[width=0.97\linewidth, trim=0em 0em 0em 0em]{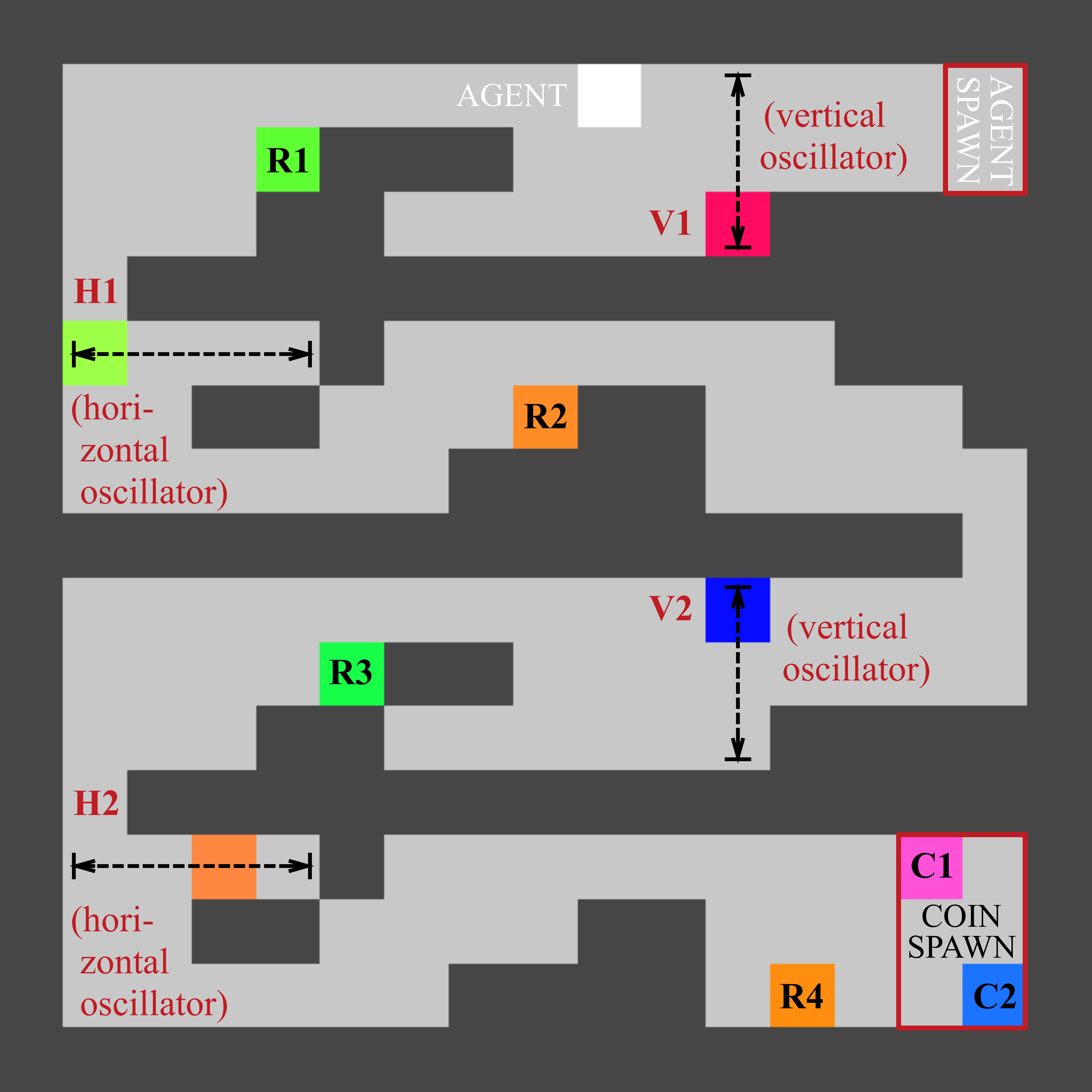}
}
\vspace{-0.8em}
\caption{\small \scalebox{0.98}{\textit{Pycolab Maze Environment Map}.}}
\label{fig:pycolabmap}
\end{minipage}
\pix\pix
\begin{minipage}{.31\columnwidth}
\vspace{-10.35px}
\begin{tabular}[b]{c}
\subfloat{
{\transparent{0.7}
\includegraphics[width=0.94\linewidth, trim=0em 0em 0em 0em]{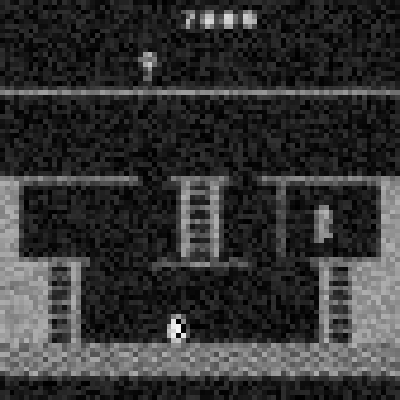}
}}
\\[-5px]
\subfloat{
{\transparent{0.7}
\includegraphics[width=0.94\linewidth, trim=0em 0em 0em 0em]{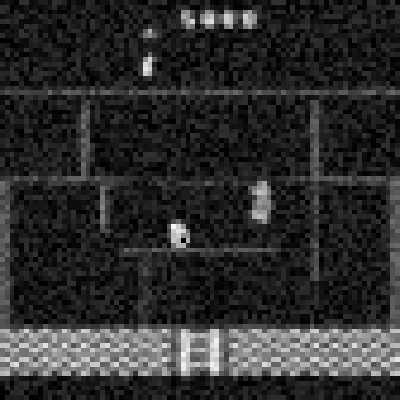}
}}
\end{tabular}
\vspace{-2.025em}
\caption{\small \scalebox{0.98}{\textit{Pixel Noise}.\hspace{-10px}}}
\label{fig:pixelframes}
\end{minipage}%
\vspace{-1.15em}
\end{figure}

\dayum{
\textbf{Implementation}~
In all experiments, we start from the same architecture/hyperparameters for \tttext{BYOL-Explore} as given in \cite{guo2022byol}, including target network EMA, open-loop horizon, intrinsic reward normalization/prioritization, representation sharing, and VMPO \cite{song2019v} as the underlying RL algorithm.
\tttext{BYOL-Hindsight} begins from the same setup. The generator, reconstructor, and critic networks are MLPs with three hidden layers of 512; the dimension of the generator noise $\epsilon$ and hindsight vector is 256; and $\lambda$\pix$=$\pix$1$. Finally, where shown for reference, RND and ICM are also implemented exactly as described in \cite{guo2022byol}.
See Appendix \ref{app:c} for additional detail.
}

\begin{figure*}[h!]
\vspace{-1em}
\centering
\makebox[1.0\textwidth][c]{
\hspace{-7px}
\subfloat[\textbf{Baseline: No Noise}]{
\includegraphics[height=0.195\linewidth, trim=0em 0em 0em 0em]
{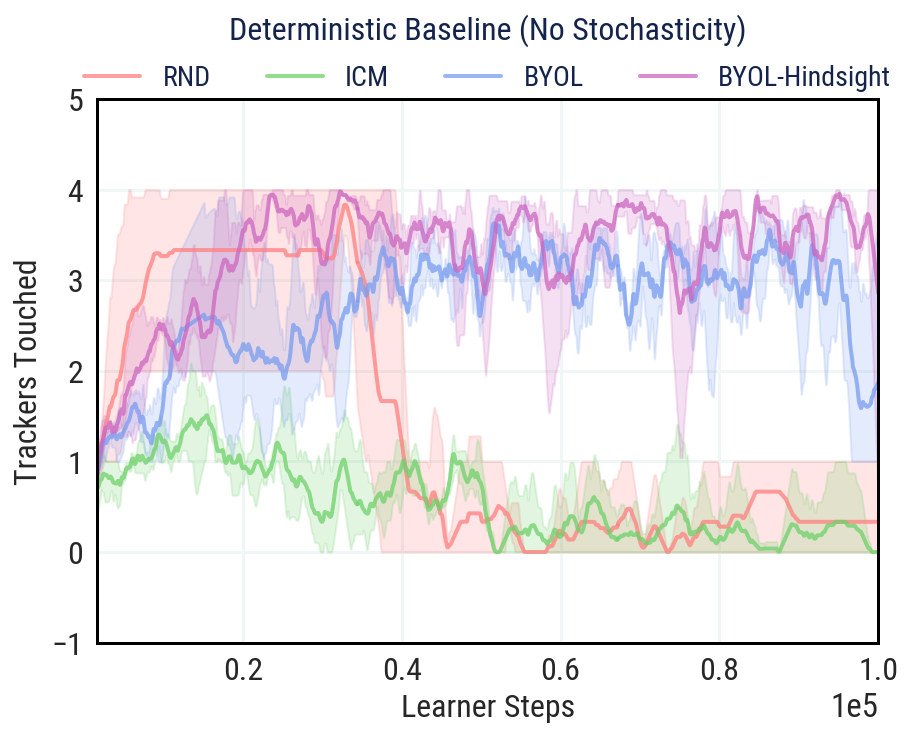}}
\hfill
\subfloat[\textbf{Brownian Oscillators}]{
\includegraphics[height=0.195\linewidth, trim=0em 0em 0em 0em]
{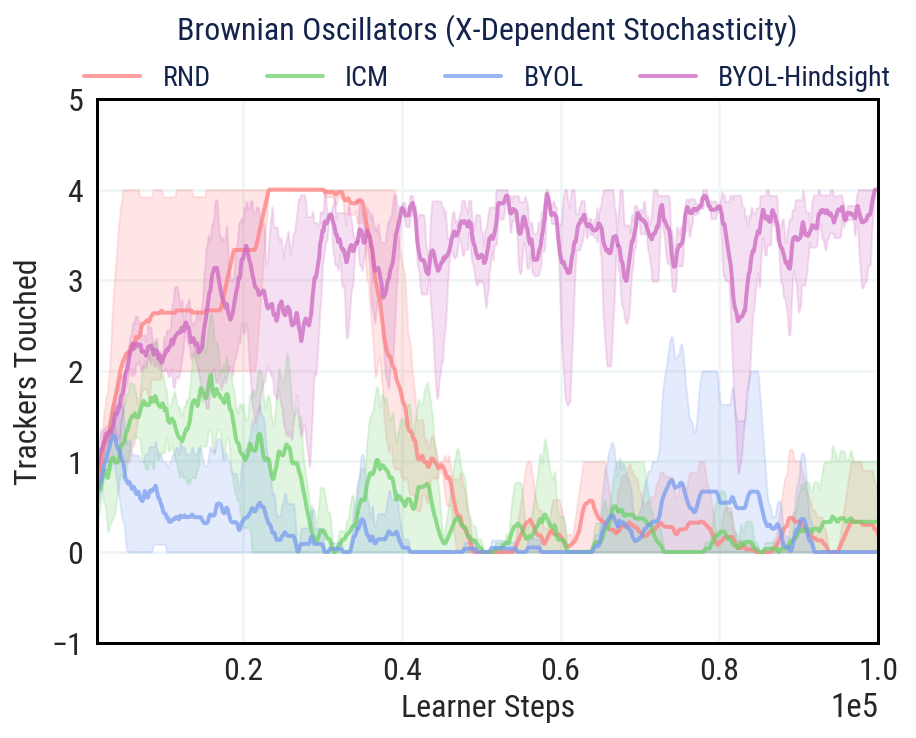}}
\hfill
\subfloat[\textbf{Random Pixel Noise}]{
\includegraphics[height=0.195\linewidth, trim=0em 0em 0em 0em]
{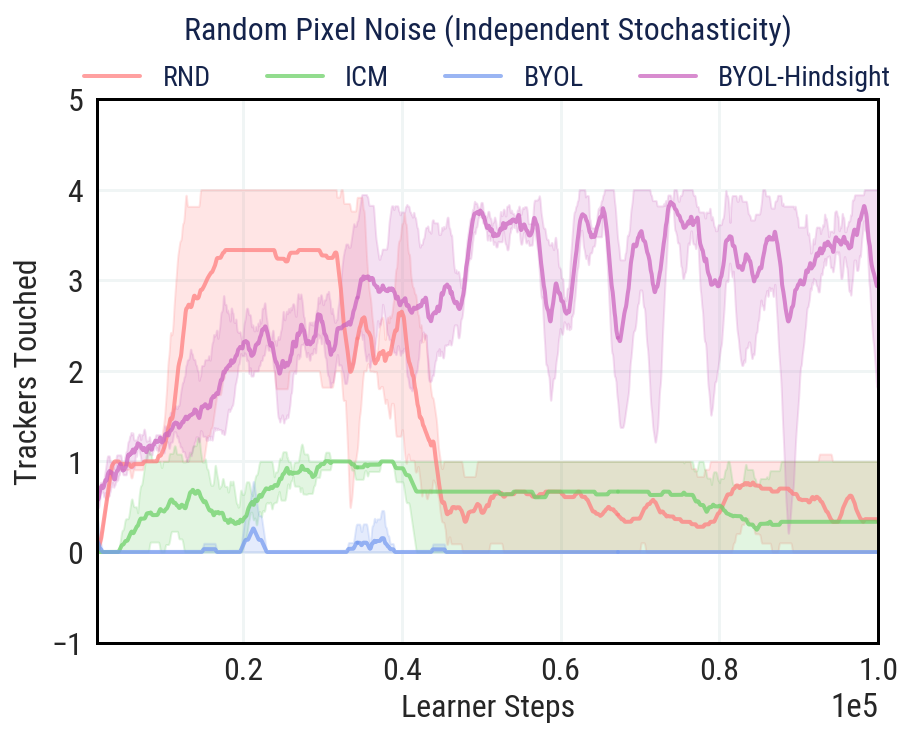}}
\hfill
\subfloat[\textbf{On-Demand Pixel Noise}]{
\includegraphics[height=0.195\linewidth, trim=0em 0em 0em 0em]
{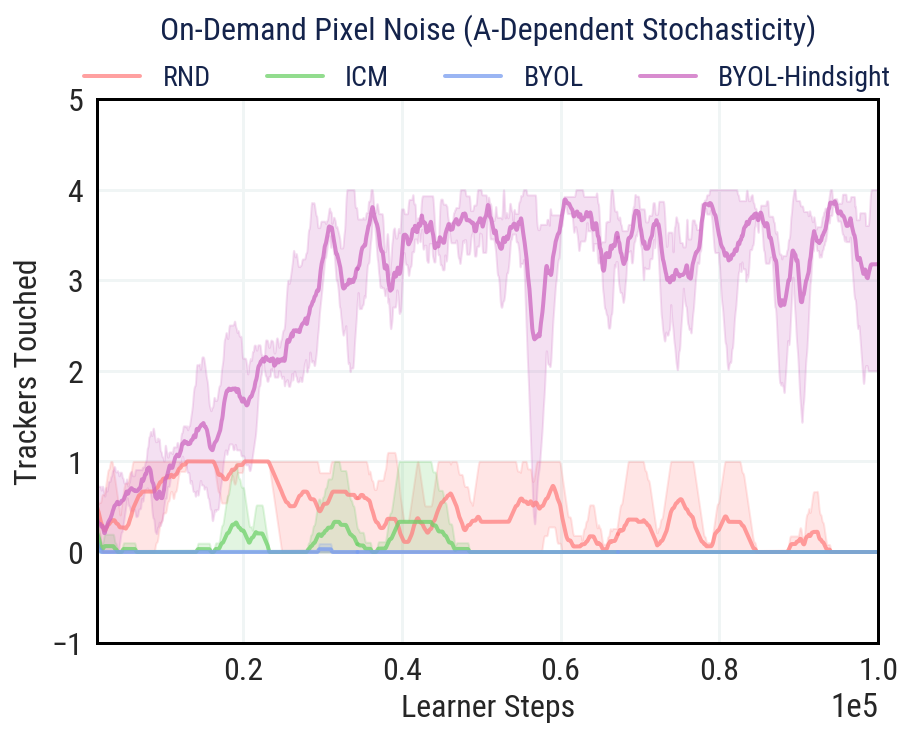}}
}
\vspace{-1em}
\caption{\small \dayum{\textit{Pycolab Maze, with Various Stochasticities}.
Performance measured by number of trackers touched in an episode (500-steps).
}}
\label{fig:pycolab}
\vspace{-1.5em}
\end{figure*}

\begin{figure*}[h!]
\centering
\makebox[1.0\textwidth][c]{
\hspace{-12px}
\subfloat[\textbf{Natural Traps (Intrinsic Only)}]{
\includegraphics[height=0.19\linewidth, trim=0em 0em 0em 0em]
{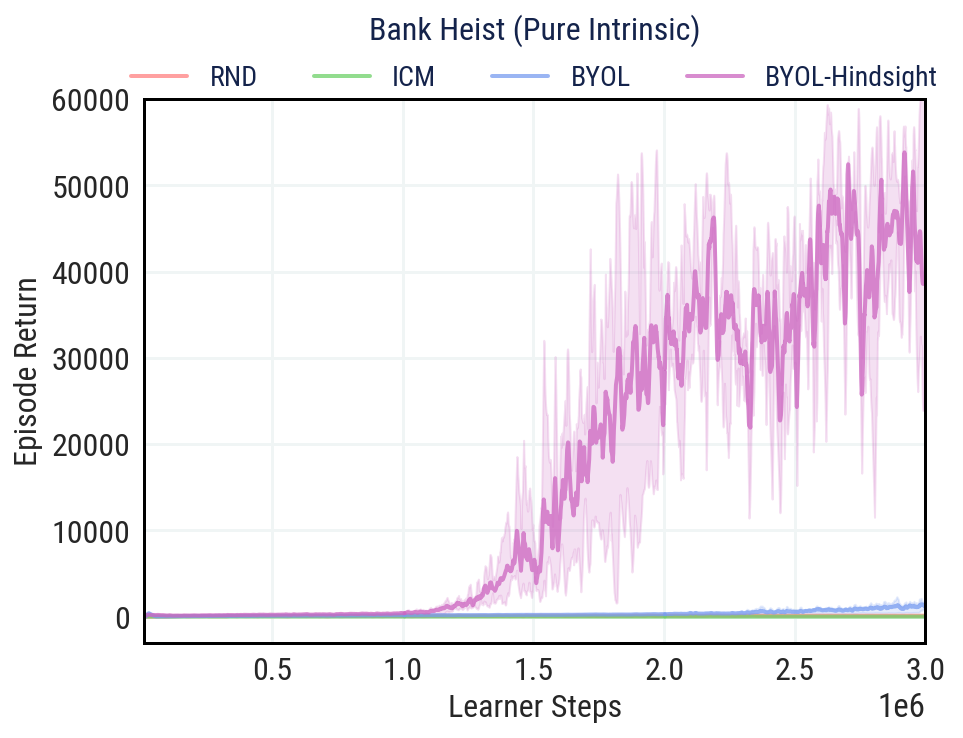}}
\hfill
\subfloat[\textbf{Natural Traps (Ext. + Intrinsic)}]{
\includegraphics[height=0.19\linewidth, trim=0em 0em 0em 0em]
{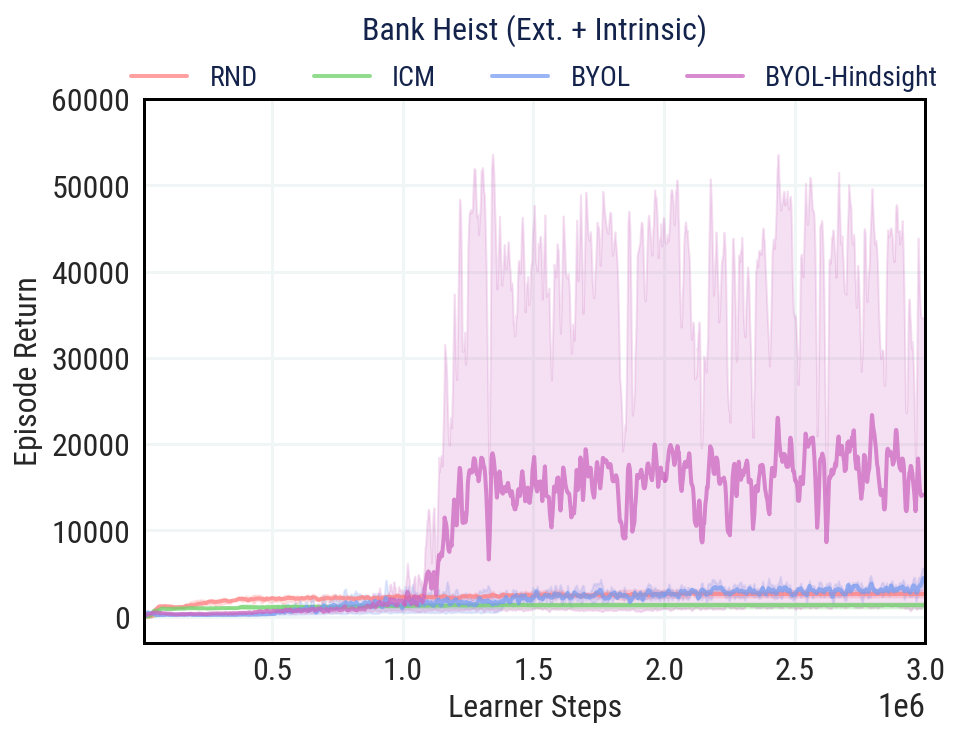}}
\hfill
\subfloat[\textbf{Sticky Actions (Intrinsic Only)}]{
\includegraphics[height=0.19\linewidth, trim=0em 0em 0em 0em]
{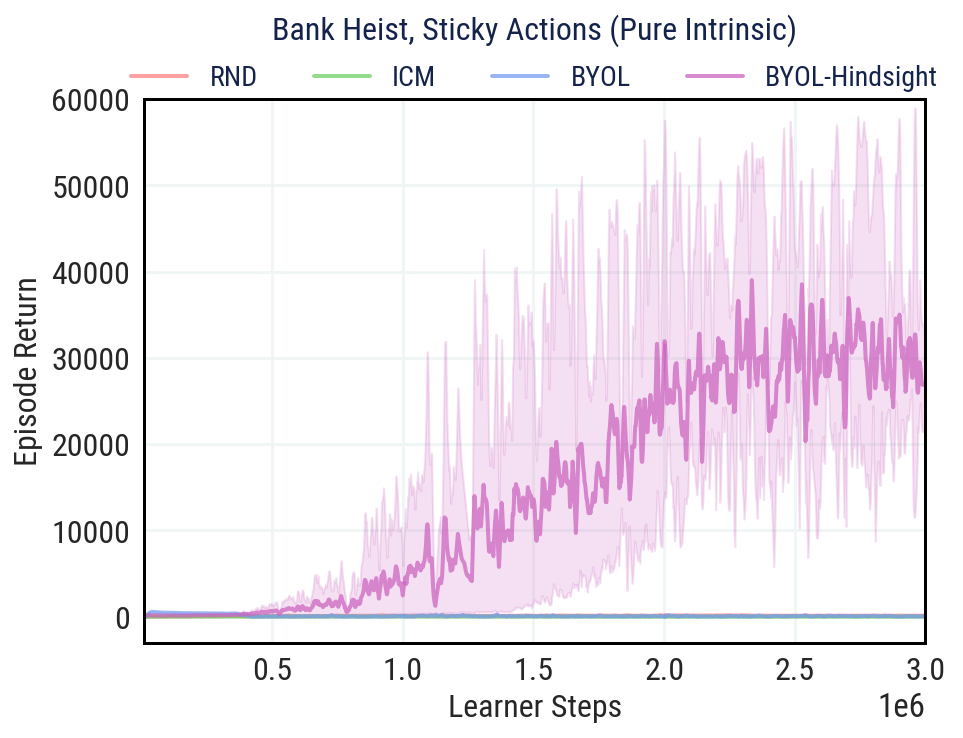}}
\hfill
\subfloat[\textbf{Sticky Actions (Ext. + Intrinsic)}]{
\includegraphics[height=0.19\linewidth, trim=0em 0em 0em 0em]
{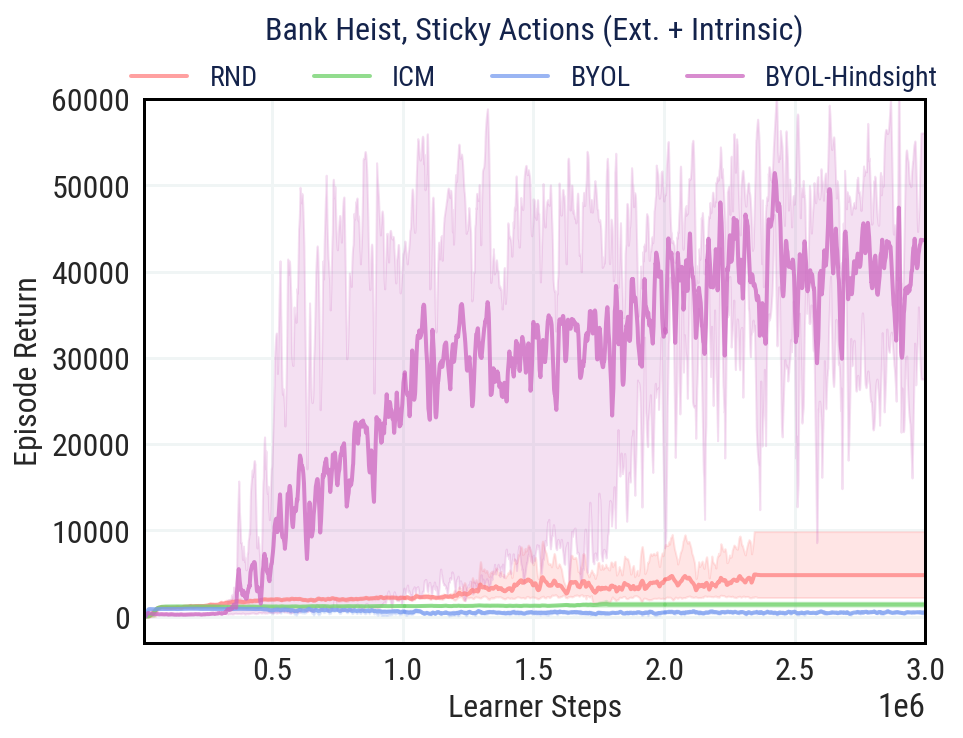}}
}
\vspace{-1em}
\caption{\small\dayum{\textit{Bank Heist, with Natural Traps and Sticky Actions}.
Performance measured by sum of extrinsic rewards obtained in an episode.
}}
\label{fig:bank:sticky:return}
\vspace{-1.35em}
\end{figure*}

\vspace{-0.5em}
\subsection{Pycolab Maze}\label{subsec:51}

\dayum{
First, to experiment with a variety of stochasticities in a controlled manner, we employ a Pycolab \cite{stepleton2017pycolab} maze (Figure~\ref{fig:pycolabmap}):
The agent spawns in the top right, and may explore past four (possibly stochastically oscillating) block elements (V1/2, H1/2), into the lower right where two coins are randomly spawned. The agent is purely intrinsically motivated, and progress is measured by trackers behind each of the block elements (R1--4). The agent only has access to a 5$\times$5 frame (i.e. square radius 2) of its immediate surroundings as input.
}

\textbf{Stochasticity}~
We use four settings: ``Baseline'' (no noise); ``Brownian Oscillators'' (a form of state-dependent noise, where oscillators perform random walks along their axes of movements); ``Random Pixel Noise'' (a form of independent noise, which adds an extra layer of randomly sampled pixels to each frame with independent probability 0.25); as well as  ``On-Demand Pixel Noise'' (a form of action-dependent noise, which does so whenever the no-op action is selected).

\dayum{
\textbf{Results}~
See Figure \ref{fig:pycolab} (100k learner steps, 3 seeds).
First, the ``Baseline'' setting tests non-specificity: Since there is no noise until the end, we expect curiosity-based exploration to do similarly with/without hindsight. For reference, we also show RND (in principle resilient to noise, as its targets are
deterministic). All algorithms reach all four trackers (with RND eventually losing interest due to vanished rewards, as the environment is small).
Second, in ``Brownian Oscillators'', \tttext{BYOL-Explore} fails to explore much beyond the first two trackers, as it is trapped by the unpredictable motion. In contrast, \tttext{BYOL-Hindsight} and RND both still explore the entire maze.
Third, in ``Random Pixel Noise'' the results are similar, except both \tttext{BYOL-Explore} and RND do worse as the noise is an entire layer of random pixels (i.e. extremely diffuse), which outcompetes all other dynamics of the world in magnitude. Interestingly, while \tttext{BYOL-Hindsight} requires slightly longer to adapt, it still performs just as well.
Lastly, the ``On-Demand Pixel Noise'' setting is most telling. \tttext{BYOL-Explore} is instantly trapped by the noise-inducing action, which it selects endlessly to generate intrinsic rewards. Even RND suffers greatly, which makes sense because the agent is no longer guaranteed a 0.75 probability of observing the world's unpolluted dynamics. In contrast, \tttext{BYOL-Hindsight} still performs as well as in the noise-free setting, underscoring robustness to different stochasticities.
}

\begin{figure*}[h!]
\vspace{-1em}
\centering
\makebox[1.0\textwidth][c]{
\hspace{-10px}
\subfloat[\textbf{Sticky Actions (Intrinsic Only)}]{
\includegraphics[height=0.195\linewidth, trim=0em 0em 0em 0em]
{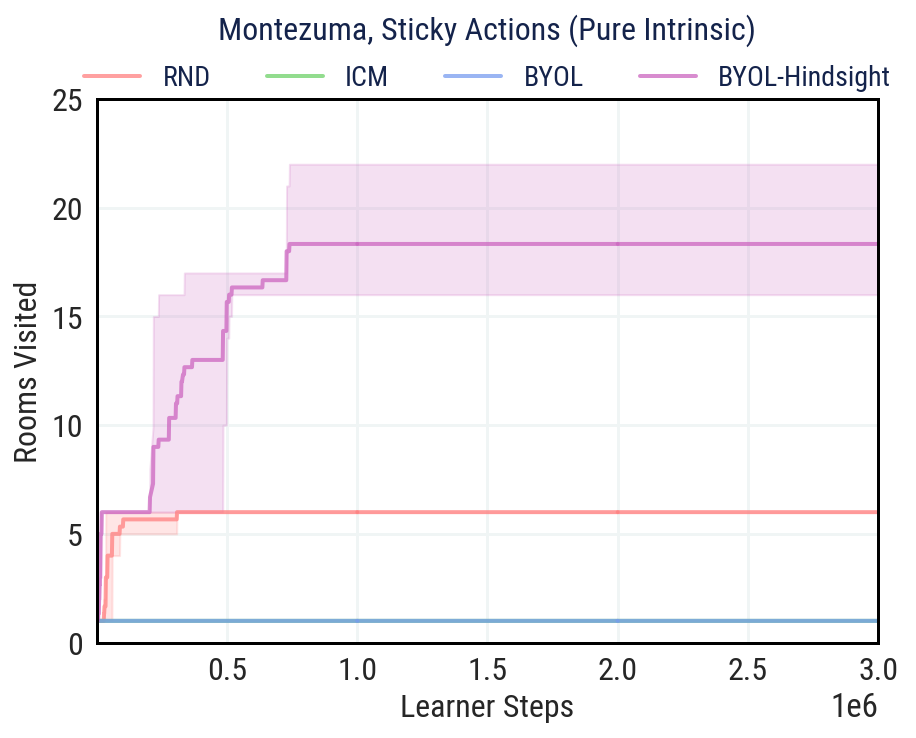}}
\hfill
\subfloat[\textbf{Sticky Actions (Ext. + Intrinsic)}]{
\includegraphics[height=0.195\linewidth, trim=0em 0em 0em 0em]
{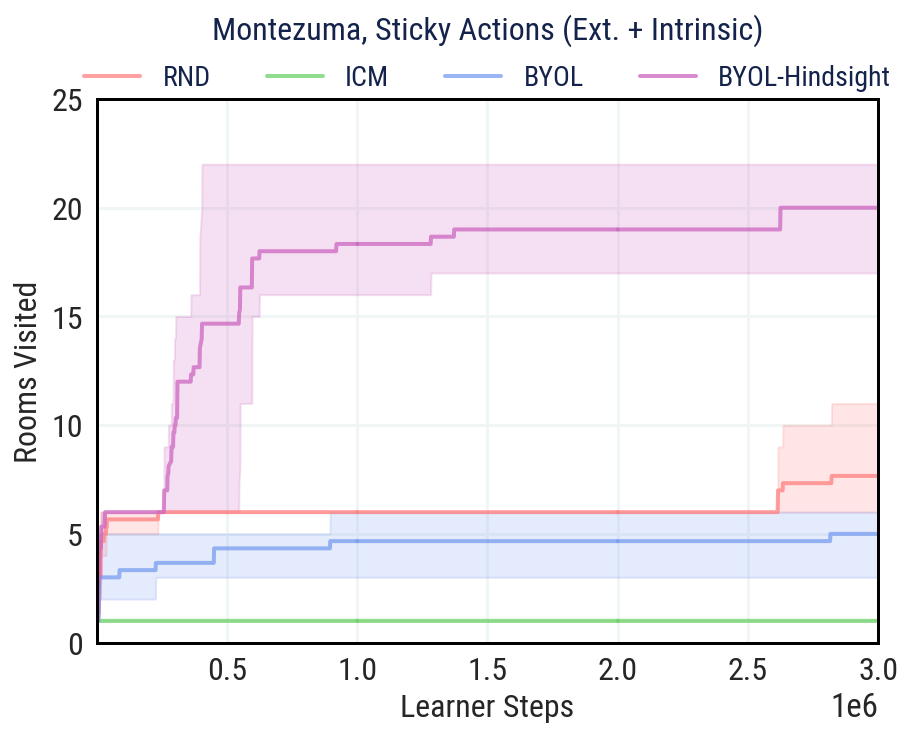}}
\hfill
\subfloat[\textbf{Non-Sticky Baseline (Intrinsic Only)}]{
\includegraphics[height=0.195\linewidth, trim=0em 0em 0em 0em]
{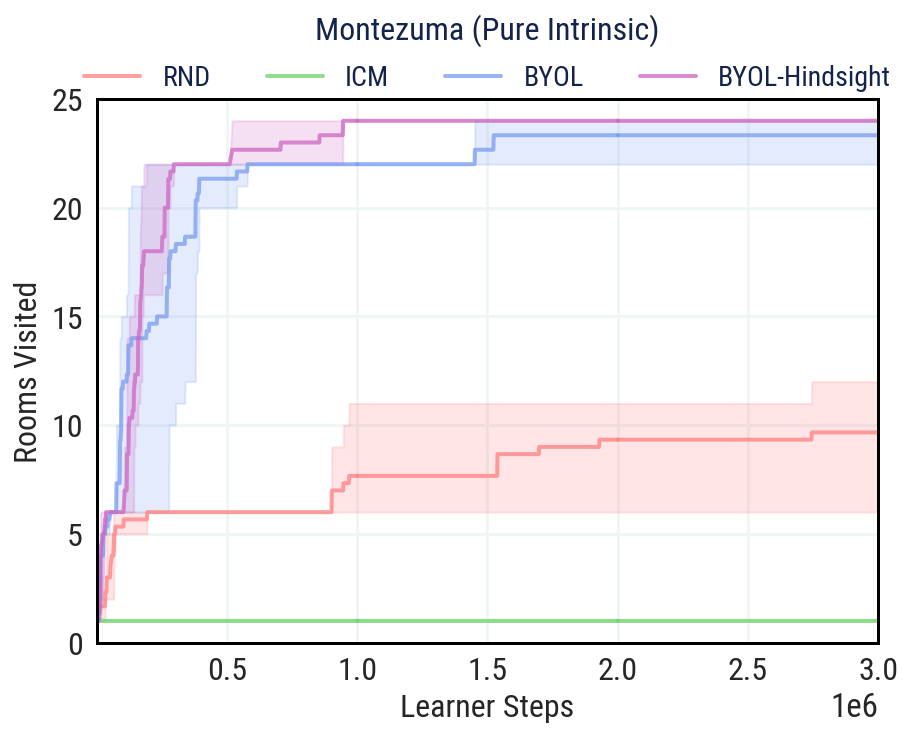}}
\hfill
\subfloat[\textbf{Non-Sticky Baseline (Ext. + Intrinsic)}]{
\includegraphics[height=0.195\linewidth, trim=0em 0em 0em 0em]
{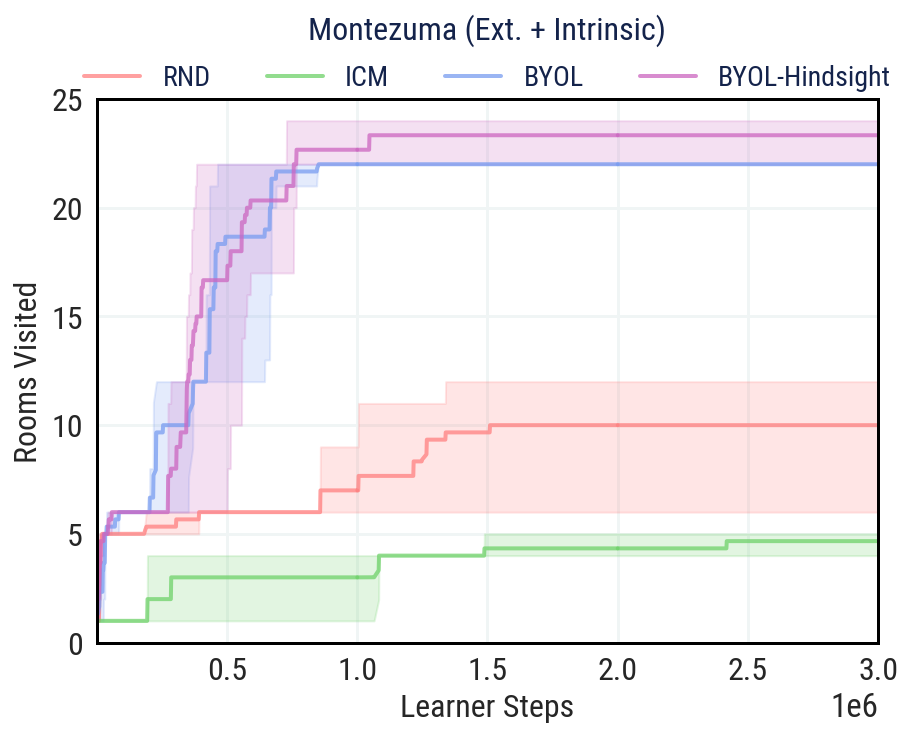}}
}
\vspace{-1em}
\caption{\small\dayum{\textit{Montezuma's Revenge, with Sticky Actions and Non-Sticky Baseline}.
Performance measured by rooms reached (episodic setting).
}}
\label{fig:montezuma:sticky:room}
\vspace{-1.1em}
\end{figure*}

\begin{figure}[h!]
\vspace{-0.9em}
\centering
\hspace{-2px}
\subfloat[\tiny\textbf{Alien (I)}]{
\includegraphics[height=0.146\linewidth, trim=0em 0em 0em 0em]
{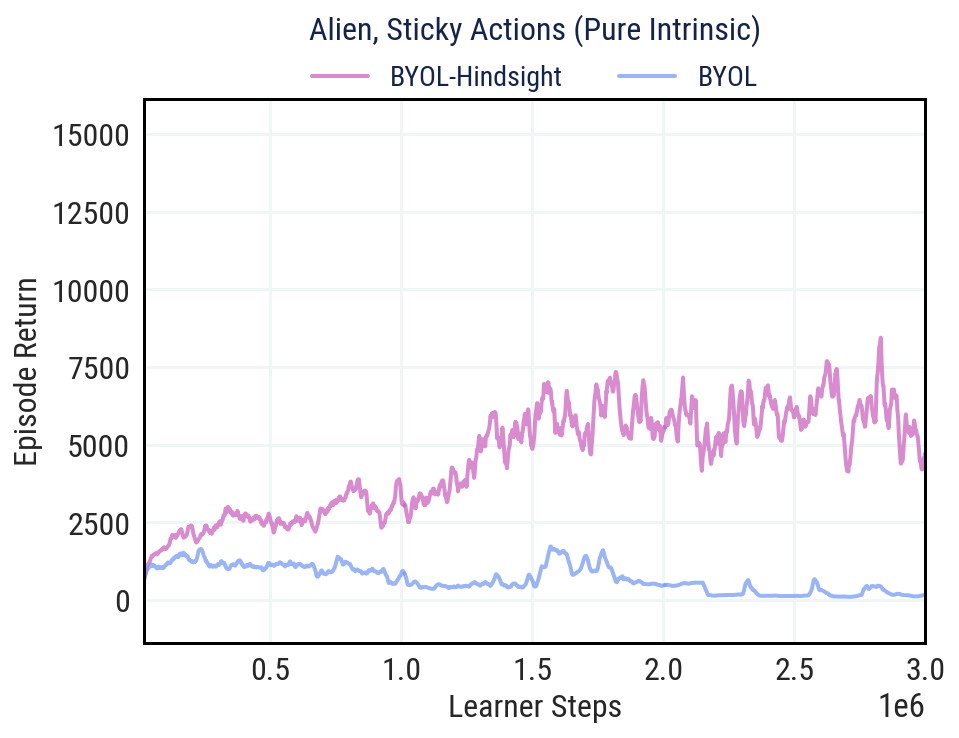}}
\hfill
\subfloat[\tiny\textbf{Bank (I)}]{
\includegraphics[height=0.146\linewidth, trim=0em 0em 0em 0em]
{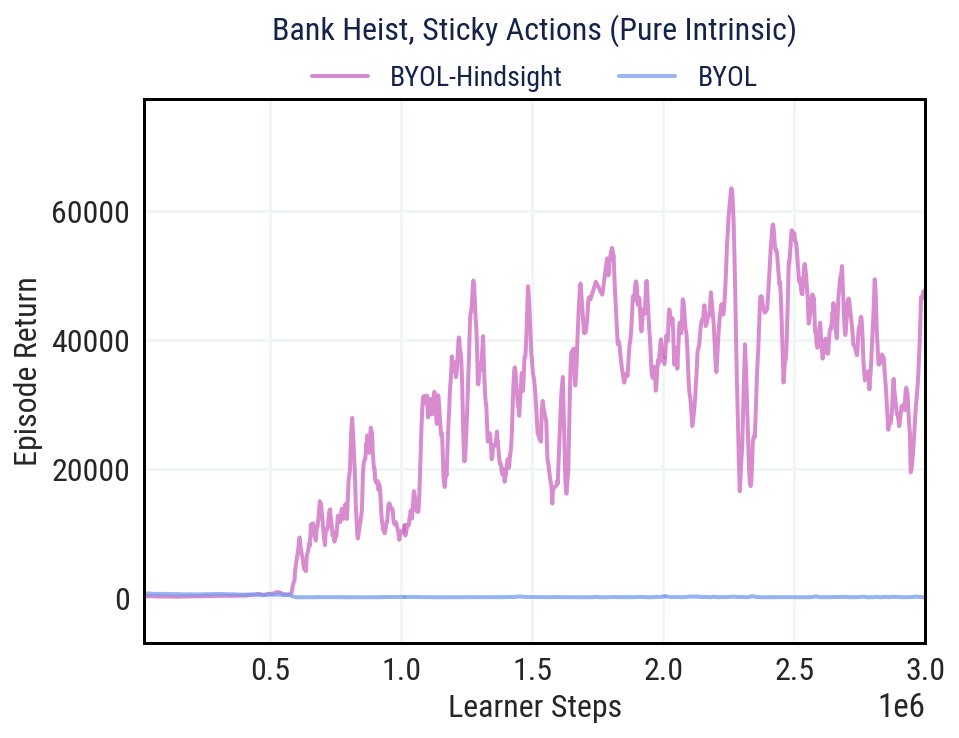}}
\hfill
\subfloat[\tiny\textbf{Freew. (I)}]{
\includegraphics[height=0.146\linewidth, trim=0em 0em 0em 0em]
{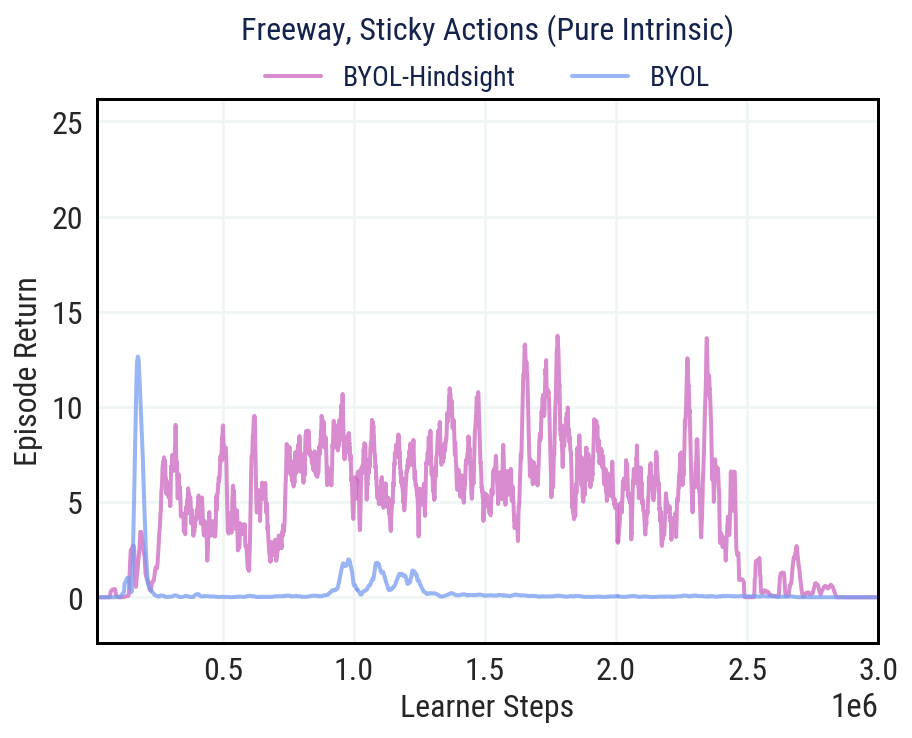}}
\hfill
\subfloat[\tiny\textbf{Gravitar (I)}]{
\includegraphics[height=0.146\linewidth, trim=0em 0em 0em 0em]
{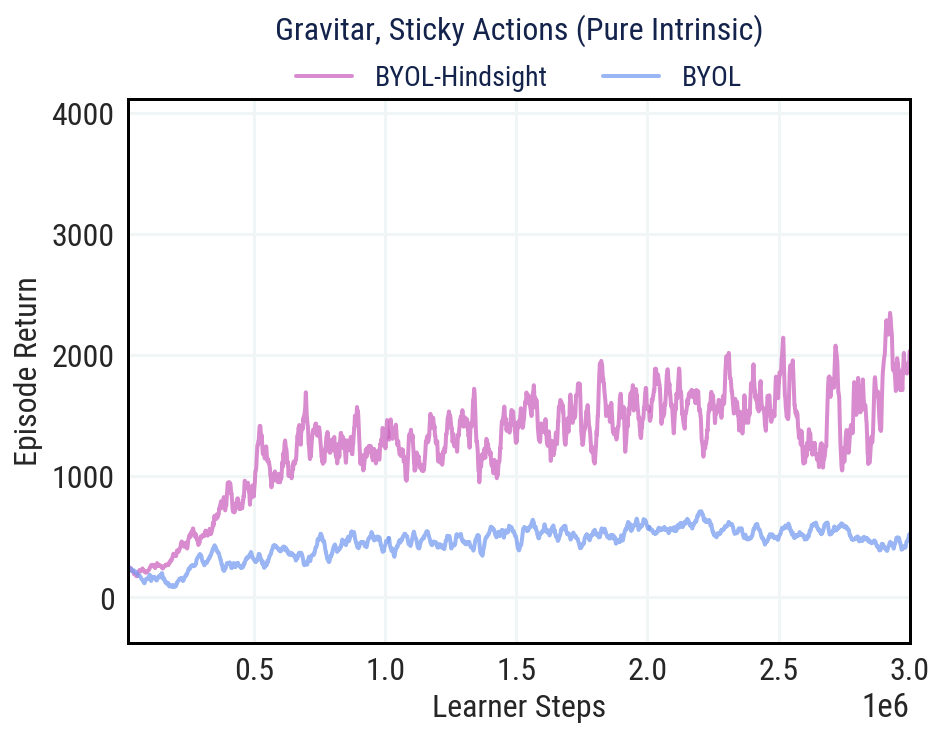}}
\hfill
\subfloat[\tiny\textbf{Hero (I)}]{
\includegraphics[height=0.146\linewidth, trim=0em 0em 0em 0em]
{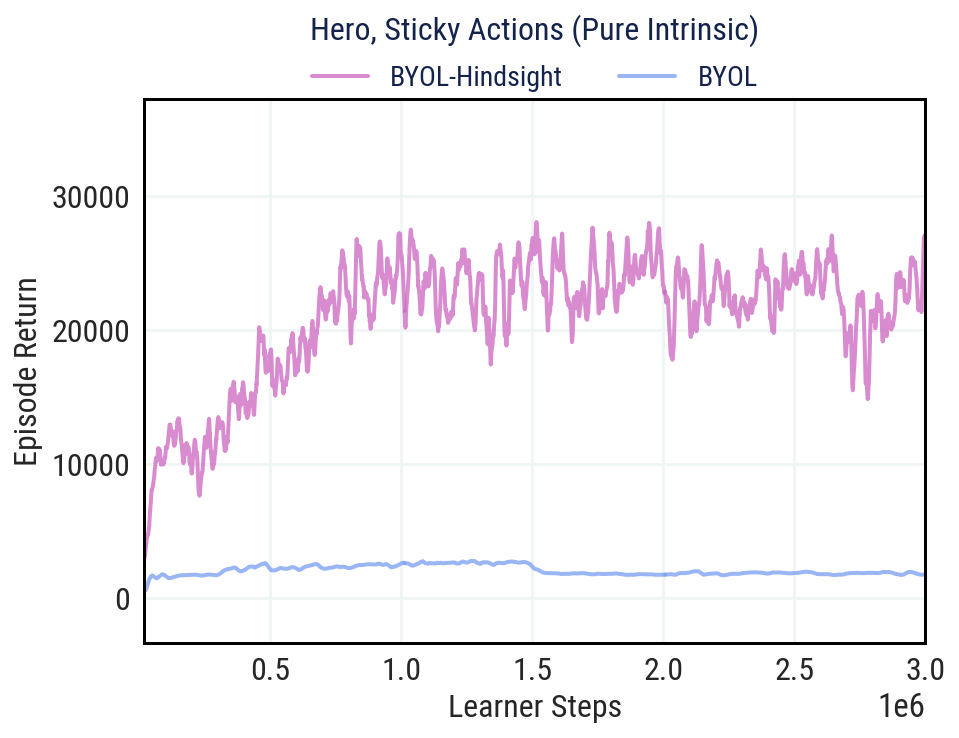}}
\vspace{-0.5em}

\hspace{-1px}
\subfloat[\tiny\textbf{Montez. (I)}]{
\includegraphics[height=0.146\linewidth, trim=0em 0em 0em 0em]
{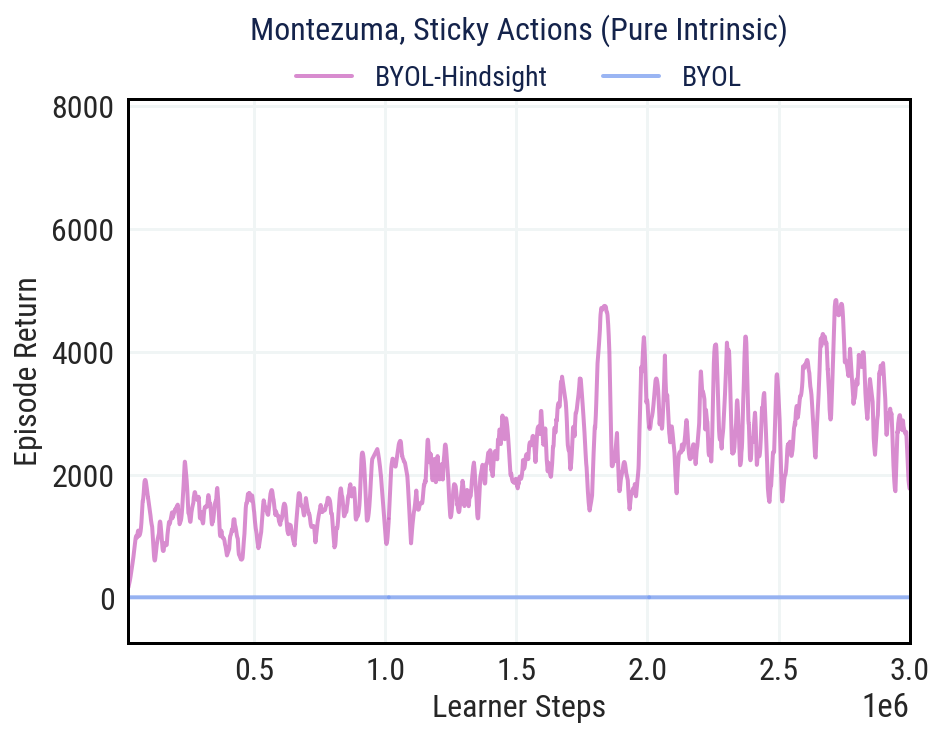}}
\hfill
\subfloat[\tiny\textbf{Pitfall (I)}]{
\includegraphics[height=0.146\linewidth, trim=0em 0em 0em 0em]
{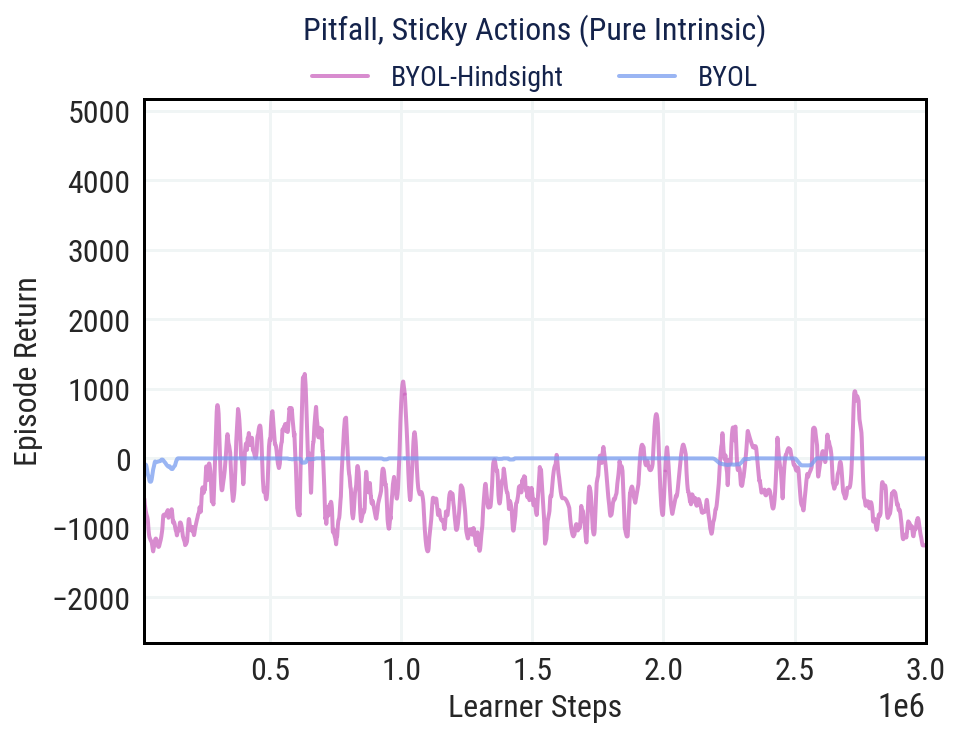}}
\hfill
\subfloat[\tiny\textbf{Priv. Eye (I)}]{
\includegraphics[height=0.146\linewidth, trim=0em 0em 0em 0em]
{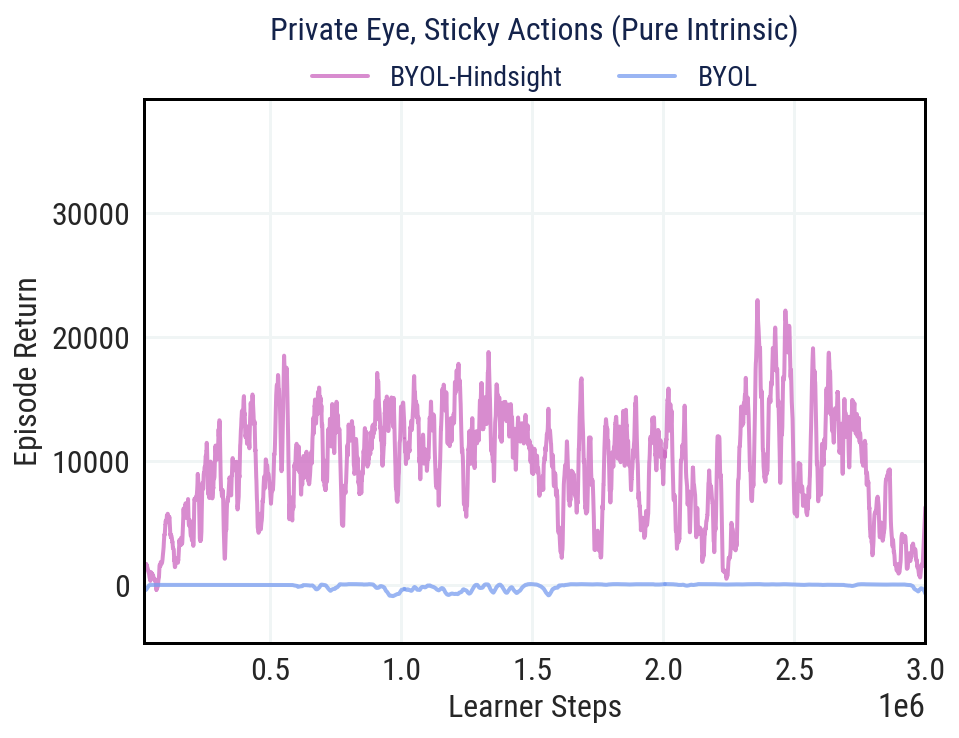}}
\hfill
\subfloat[\tiny\textbf{Solaris (I)}]{
\includegraphics[height=0.146\linewidth, trim=0em 0em 0em 0em]
{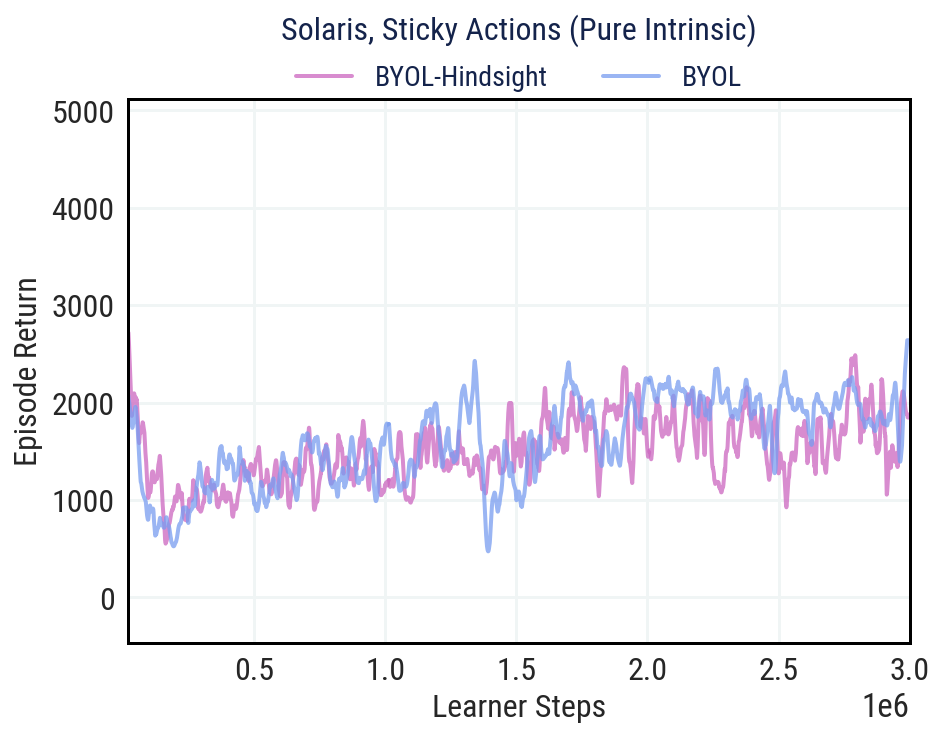}}
\hfill
\subfloat[\tiny\textbf{Venture (I)}]{
\includegraphics[height=0.146\linewidth, trim=0em 0em 0em 0em]
{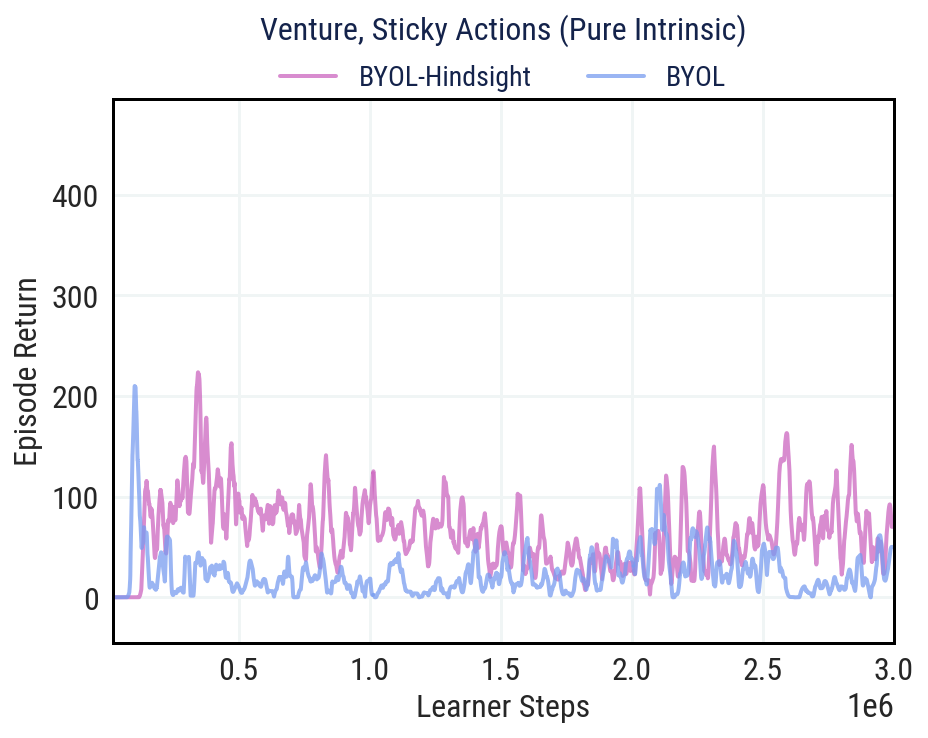}}
\vspace{-0.4em}
\caption{\small\dayum{\textit{Hard Exploration Games, with Sticky Actions}.
Performance measured by the sum of extrinsic rewards in an episode.
Intrinsic-only ``(I)'' results shown; Figure \ref{fig:atari:return:big} shows mixed ``(E+I)''.
}}
\label{fig:atari:return}
\vspace{-2.1em}
\end{figure}

\vspace{-0.35em}
\subsection{Bank Heist}\label{subsec:52}
\vspace{0.2em}

\dayum{
We use Atari with preprocessed grayscale 84$\times$84-pixel images as input \cite{bellemare2013arcade}.
In all settings, we consider both ``intrin-
sic-only'' (no extrinsic signal) and ``mixed'' (extrinsic + int-rinsic rewards) exploration regimes.
In Bank Heist, the goal is to rob as many banks as possible while avoiding the police.
}

\dayum{
\textbf{Stochasticity}~
First, Bank Heist is characterized by natur-ally-occurring stochastic traps (``Natural Traps''), as noted by prior work \cite{mavor2022stay}: It is impossible to predict where banks randomly regenerate and where bombs explode, thus a predictive error-based agent would simply endlessly enter and exit mazes while dropping bombs. Second, as an additional noise factor we use ``Sticky Actions'' \cite{machado2018revisiting} with stickiness 0.1.
}

\dayum{
\textbf{Results}~
See Figure \ref{fig:bank:sticky:return} (3M learner steps, 3 seeds).
Like in prior work, we measure the extrinsic reward per episode that the agent obtains, as a proxy for exploration ability.
In both the ``Natural Traps'' and ``Sticky Actions'' settings, and in both intrinsic-only and mixed regimes, \tttext{BYOL-Explore}'s progress is immediately derailed by the stochastic traps, whereas
\tttext{BYOL-Hindsight} achieves vastly better scores.
}

\vspace{-0.5em}
\subsection{Montezuma's Revenge}\label{subsec:53}

Playing Montezuma's Revenge requires learning complex dynamics including navigating around timed traps and moving enemies, and collecting keys to open doors in sequence.

\dayum{
\textbf{Stochasticity}~
First, Montezuma's Revenge is largely det-erministic, which forms a natural baseline for testing non-specificity in a challenging exploration setting. Second, we use ``Sticky Actions'' similar to above to add stochasticity.
}

\dayum{
\textbf{Results}~
See Figure \ref{fig:montezuma:sticky:room} (3M learner steps, 3 seeds).
Like in prior work, exploration is measured by the number of different dungeon rooms the agent manages to discover over its lifetime.
In ``Sticky Actions'', \tttext{BYOL-Explore} instantly flatlines in the intrinsic-only regime, and only does marginally better in the mixed regime. In contrast, \tttext{BYOL-Hindsight} explores most of the rooms in both regimes---which is an unprecedented result.
In the ``Non-Sticky Baseline'', in both regimes \tttext{BYOL-Hindsight} does as well as the original performance of \tttext{BYOL-Explore}, which verifies non-specificity.
See Appendix \ref{subapp:sticky} for evaluation results on episode return.
}

\vspace{-0.5em}
\subsection{Hard Exploration Games}\label{subsec:54}

\dayum{
We conduct broad-based experiments for the ten hardest exploration games in Atari, where stochasticity is introduced with sticky actions as above.
See Figure \ref{fig:atari:return} for results in the intrinsic-only ``(I)'' regime, and also Figure \ref{fig:atari:return:big} for the mixed ``(E+I)'' regime (3M learner steps, 1~seed). Like in prior work, we use extrinsic reward as a proxy for ``interesting behavior''.
In the large majority of cases, \tttext{BYOL-Hindsight} improves over \tttext{BYOL-Explore}, especially when the latter simply flatlines.
See Appendix \ref{subapp:hard} for additional results and analysis.
}

\subsection{Persistive Noise}\label{subsec:55}

\dayum{
While sticky actions is the standard protocol for adding stoc-hasticity, we further design an especially challenging setting by corrupting observations with an additive layer of 84$\times$84 pixel noise that persists across time as an action-triggerable random walk (Figure \ref{fig:pixelframes}). See Appendix \ref{subapp:persistive} for full details and results, again verifying the robustness of the algorithm.
}

\vspace{-0.5em}
\subsection{Temperature Sensitivity}\label{subsec:56}

\dayum{
The inner term in the contrastive loss (Objective \ref{obj:con}) shares a similar form with contrastive losses in unsupervised representation learning \cite{oord2018representation,tschannen2019mutual,he2020momentum,chen2020simple}, which admits a temperature parameter controlling the strength of penalties on negative samples \cite{wang2021understanding}. See Appendix \ref{subapp:temperature} for a sensitivity analysis of the temperature parameter on the effectiveness of the algorithm.
}

\vspace{-0.5em}
\subsection{Analysis of Invariance}\label{subsec:57}

\dayum{
For insight into invariance properties of the learned hindsight representations, see Appendix \ref{subapp:intrinsic} for analysis of intrinsic rewards over training; see Appendix \ref{subapp:outcome} for a comparison of prediction loss, reconstruction loss, and hindsight-only loss; and see Appendix \ref{subapp:representations} for visualizations of what information the hindsight representations may be encoding.
}
\vspace{-0.55em}
\section{Conclusion}\label{sec:6}

\dayum{
In this work, we studied the problem that stochasticity poses
to predictive error-based exploration.
Theoretically, we ref-
ined our notion of curiosity to separate (learnable) epistemic knowledge from (unlearnable) aleatoric variation.
Algorith-
mically, we proposed a method to learn (future-summari-
zing) representations of hindsight disentangled from (history
-summarizing) representations of context.
Practically, we arrived at a simple and scalable framework for
generating
(reducible) intrinsic rewards even in the presence of (irre-
ducible) stochastic traps---without
estimating the problematic entropy term at all.
Our perspective shares connections with counterfactuals in
policy evaluation \cite{buesing2018woulda,oberst2019counterfactual, lorberbom2021learning},
credit assignment \cite{harutyunyan2019hindsight,nota2021posterior,mesnard2021counterfactual}, invariance \cite{zhang2020learning,bica2021invariant,lu2022invariant}, and
fairness \cite{louizos2016variational,moyer2018invariant,lotfollahi2019conditional,foster2022contrastive}.
Future
work may study explicitly generative world models to map stochastic latents to outcomes,
as well as assessing the benefit of the approach in other stochastic domains such as NetHack.
See Appendix \ref{app:d} for an extended discussion.
}


\clearpage
\balance

\bibliographystyle{unsrt}
\bibliography{bib}


\clearpage
\onecolumn
\appendix


\icmltitle{\mytitle}

\icmlsetsymbol{equal}{*}
\begin{icmlauthorlist}
\icmlauthorAgain{Daniel Jarrett}{deep}
\icmlauthorAgain{Corentin Tallec}{deep}
\icmlauthorAgain{Florent Altch\'{e}}{deep}
\icmlauthorAgain{Thomas Mesnard}{deep}
\icmlauthorAgain{R\'{e}mi Munos}{deep}
\icmlauthorAgain{Michal Valko}{deep}
\end{icmlauthorlist}

\vspace{-1.25em}

\printAffiliationsAndNoticeAgain{}


\part{Supplementary Materials}
\localtableofcontents


\clearpage
\section{Proofs of Propositions}\label{app:a}

To simplify our notation, we remove subscripts such that $X,A,Y$ denotes the transition $X_{t},A_{t},X_{t+1}$, and $Z$ denotes the latent $Z_{t+1}$.
Then the environment's dynamics is given by $\tau(Y|x,a)$, the agent's policy is given by $\pi(A|x)$, and the induced state visitation is given by $\rho_{\pi}(X)$. The generator is denoted $p_{\theta}(Z|x,a,y)$, reconstructor $f_{\eta}(x,a,z)$, and critic $g_{\nu}(x,a,z)$.

\subsection{Pointwise Mutual Information}

We start by deriving several lemmas that will be useful, the first being a pointwise version of Barber and Agakov's variational lower bound on mutual information \cite{agakov2004algorithm,poole2019variational}:

\begin{relemma}[restate=thmba,name=Pointwise Barber-Agakov]\upshape\label{thm:ba}
Denote the pointwise mutual information:
\begin{align}
\text{PMI}_{\theta}(x,a;z)
\coloneqq
\log\frac{p_{\theta}(z|x,a)}{p_{\theta}(z)}
\end{align}
Then for any variational distribution $q$:
\begin{align}
\mathbb{E}_{Z\sim p_{\theta}(\cdot|x,a)}
\text{PMI}_{\theta}(x,a;Z)
\spacegeqq
\mathbb{E}_{Z\sim p_{\theta}(\cdot|x,a)}
\log\frac{q(Z|x,a)}{p_{\theta}(Z)}
\end{align}
\end{relemma}

\textit{Proof}. Starting from the left hand side:

\begin{align}
\mathbb{E}_{Z\sim p_{\theta}(\cdot|x,a)}
\text{PMI}_{\theta}(x,a;Z)
&
\spaceqq
\mathbb{E}_{Z\sim p_{\theta}(\cdot|x,a)}
\log\frac{p_{\theta}(Z|x,a)}{p_{\theta}(Z)}
\\
&
\spaceqq
\mathbb{E}_{Z\sim p_{\theta}(\cdot|x,a)}
\log\frac{p_{\theta}(Z|x,a)}{p_{\theta}(Z)}
+
\mathbb{E}_{Z\sim p_{\theta}(\cdot|x,a)}
\log\frac{q(Z|x,a)}{q(Z|x,a)}
\\
&
\spaceqq
\mathbb{E}_{Z\sim p_{\theta}(\cdot|x,a)}
\log\frac{q(Z|x,a)}{p_{\theta}(Z)}
+
D_{\text{KL}}
\big(
p_{\theta}(Z|x,a)\|q(Z|x,a)
\big)
\\
&
\spacegeqq
\mathbb{E}_{Z\sim p_{\theta}(\cdot|x,a)}
\log\frac{q(Z|x,a)}{p_{\theta}(Z)}
\end{align}

which completes the proof. \QED

Next, we define a generic contrastive expression with $K$\pix$-$\pix$1$ ``negative'' samples of $Z$, and show that taking its expectation with respect to those samples yields a valid (i.e. normalized) probability density:

\begin{relemma}[restate=thmnorm,name=Normalized Variational]\upshape\label{thm:norm}
Given independent samples $z_{1:K-1}$ from $p_{\theta}$, define:
\begin{align}
q(z|x,a,z_{1:K-1})
\coloneqq
\frac
{p_{\theta}(z)\cdot e^{g_{\nu}(x,a,z)}}
{\frac{1}{K}\left(e^{g_{\nu}(x,a,z)}+\sum_{i=1}^{K-1}e^{g_{\nu}(x,a,z_{i})}\right)}
\end{align}
then the following defines a normalized density:
\begin{align}
q(Z|x,a)
\coloneqq
\mathbb{E}_{Z_{1:K-1}\sim p_{\theta}^{K-1}}
q(Z|x,a,Z_{1:K-1})
\end{align}
\end{relemma}

\textit{Proof}. The expectation integrates to one:

\begin{align}
\int_{\mathcal{Z}}
q(z|x,a)
dz
&
\spaceqq
\int_{\mathcal{Z}}
\mathbb{E}_{Z_{1:K-1}\sim p_{\theta}^{K-1}}
\frac
{p_{\theta}(z)\cdot e^{g_{\nu}(x,a,z)}}
{\frac{1}{K}\left(e^{g_{\nu}(x,a,z)}+\sum_{i=1}^{K-1}e^{g_{\nu}(x,a,Z_{i})}\right)}
dz
\\
&
\spaceqq
\mathbb{E}_{\substack{Z\sim p_{\theta}\\Z_{1:K-1}\sim p_{\theta}^{K-1}}}
\frac
{e^{g_{\nu}(x,a,Z)}}
{\frac{1}{K}\left(e^{g_{\nu}(x,a,Z)}+\sum_{i=1}^{K-1}e^{g_{\nu}(x,a,Z_{i})}\right)}
\\
&
\spaceqq
K\cdot
\mathbb{E}_{Z_{1:K}\sim p_{\theta}^{K}}
\frac
{e^{g_{\nu}(x,a,Z_{1})}}
{\sum_{i=1}^{K}e^{g_{\nu}(x,a,Z_{i})}}
\\
&
\spaceqq
\mathbb{E}_{Z_{1:K}\sim p_{\theta}^{K}}
\frac
{\sum_{j=1}^{K}e^{g_{\nu}(x,a,Z_{j})}}
{\sum_{i=1}^{K}e^{g_{\nu}(x,a,Z_{i})}}
\\[2ex]
&
\spaceqq
1
\end{align}

which completes the proof. \QED

These two results allow us to show that the information $Z$ contains on a tuple $x,a$---with respect to the generator parameterized as $\theta$---is lower-bounded by the $x,a$-conditioned contrastive loss between ``positive'' samples $Z\sim p_{\theta}(\cdot|x,a)$ from the posterior and ``negative'' samples $Z\sim p_{\theta}$ from the prior:

\begin{relemma}[restate=thminfo,name=State-Action Lower Bound]\upshape\label{thm:info}
The $x,a$-wise mutual information satisfies:
\begin{align}
\mathbb{E}_{Z\sim p_{\theta}(\cdot|x,a)}
&
\text{PMI}_{\theta}(x,a;Z)
\nonumber
\\
&
\spacegeqq
\mathbb{E}_{\substack{Z\sim p_{\theta}(\cdot|x,a)\\Z_{1:K-1}\sim p_{\theta}^{K-1}}}
\log
\frac
{e^{g_{\nu}(x,a,Z)}}
{\frac{1}{K}\left(e^{g_{\nu}(x,a,Z)}+\sum_{i=1}^{K-1}e^{g_{\nu}(x,a,Z_{i})}\right)}
\end{align}
\end{relemma}

\textit{Proof}. Use Lemmas \ref{thm:ba} and \ref{thm:norm}, then Jensen's inequality:

\begin{align}
\mathbb{E}_{Z\sim p_{\theta}(\cdot|x,a)}
&
\text{PMI}_{\theta}(x,a;Z)
\nonumber
\\
&
\spacegeqq
\mathbb{E}_{Z\sim p_{\theta}(\cdot|x,a)}
\log\frac{q(Z|x,a)}{p_{\theta}(Z)}
\\
&
\spaceqq
\mathbb{E}_{Z\sim p_{\theta}(\cdot|x,a)}
\log\mathbb{E}_{Z_{1:K-1}\sim p_{\theta}^{K-1}}
\frac{q(Z|x,a,Z_{1:K-1})}{p_{\theta}(Z)}
\\
&
\spacegeqq
\mathbb{E}_{\substack{Z\sim p_{\theta}(\cdot|x,a)\\Z_{1:K-1}\sim p_{\theta}^{K-1}}}
\log\frac{q(Z|x,a,Z_{1:K-1})}{p_{\theta}(Z)}
\\
&
\spaceqq
\mathbb{E}_{\substack{Z\sim p_{\theta}(\cdot|x,a)\\Z_{1:K-1}\sim p_{\theta}^{K-1}}}
\log
\frac
{e^{g_{\nu}(x,a,Z)}}
{\frac{1}{K}\left(e^{g_{\nu}(x,a,Z)}+\sum_{i=1}^{K-1}e^{g_{\nu}(x,a,Z_{i})}\right)}
\end{align}

which completes the proof. \QED

Next, we show that our invariance loss (Objective \ref{obj:inv}) for a tuple $x,a,z$ is equal to the pointwise mutual information in the limit of infinitely large negative batches, assuming an optimal critic parameter:

\begin{relemma}[restate=thmopt,name=Pointwise Asymptotic Equality]\upshape\label{thm:opt}
Define the transition-wise contrastive loss:
\begin{align}
\mathcal{L}_{\theta,\nu}^{K,\text{con.}}(x,a,z)
\coloneqq
\mathbb{E}_{Z_{1:K-1}\sim p_{\theta}^{K-1}}
\log
\frac
{e^{g_{\nu}(x,a,z)}}
{\frac{1}{K}\left(e^{g_{\nu}(x,a,z)}+\sum_{i=1}^{K-1}e^{g_{\nu}(x,a,Z_{i})}\right)}
\end{align}
and the optimal critic parameter:
\begin{align}
\nu_{K}^{*}
\coloneqq
\underset{\nu}{\text{arg}\pix\text{max}}~~
\mathbb{E}_{\substack{X\sim\rho_{\pi}\\A\sim\pi(\cdot|X)\\Y\sim\tau(\cdot|X,A)\\Z\sim p_{\theta}(\cdot|X,A,Y)}}
\mathcal{L}_{\theta,\nu}^{K,\text{con.}}(X,A,Z)
\end{align}
Then $
\lim_{K\rightarrow\infty}
\mathcal{L}_{\theta,\nu_{K}^{*}}^{K,\text{con.}}(x,a,z)
=
\text{PMI}_{\theta}(x,a;z)
$.
\end{relemma}

\textit{Proof}. The \smash{$\mathbb{E}[\mathcal{L}_{\theta,\nu}^{K,\text{con.}}(X,A,Z)]$} term is just the InfoNCE loss between variables $Z$ and $X,A$, so we know that \smash{$\nu_{K}^{*}$} satisfies \smash{$
g_{\nu_{K}^{*}}(x,a,z)
=
\log\frac{p_{\theta}(z|x,a)}{p_{\theta}(z)}
+
c(x,a)
$}. Substituting this back into \smash{$\mathcal{L}_{\theta,\nu}^{K,\text{con.}}(x,a,z)$}:

\begin{align}
\lim_{K\rightarrow\infty}
&
\mathcal{L}_{\theta,\nu_{K}^{*}}^{K,\text{con.}}(x,a,z)
\\
&
=
\lim_{K\rightarrow\infty}
\mathbb{E}_{Z_{1:K-1}\sim p_{\theta}^{K-1}}
\log
\frac
{e^{g_{\nu_{K}^{*}}(x,a,z)}}
{\frac{1}{K}\left(e^{g_{\nu_{K}^{*}}(x,a,z)}+\sum_{i=1}^{K-1}e^{g_{\nu_{K}^{*}}(x,a,Z_{i})}\right)}
\\
&
=
\lim_{K\rightarrow\infty}
\mathbb{E}_{Z_{1:K-1}\sim p_{\theta}^{K-1}}
\log
\frac
{\frac{p_{\theta}(z|x,a)}{p_{\theta}(z)}}
{\frac{1}{K}\left(\frac{p_{\theta}(z|x,a)}{p_{\theta}(z)}+\sum_{i=1}^{K-1}\frac{p_{\theta}(Z_{i}|x,a)}{p_{\theta}(Z_{i})}\right)}
\\
&
=
\lim_{K\rightarrow\infty}
\mathbb{E}_{Z_{1:K-1}\sim p_{\theta}^{K-1}}
\left[
\log
\frac{p_{\theta}(z|x,a)}{p_{\theta}(z)}
-
\log
\frac{\frac{p_{\theta}(z|x,a)}{p_{\theta}(z)}+\sum_{i=1}^{K-1}\frac{p_{\theta}(Z_{i}|x,a)}{p_{\theta}(Z_{i})}}{K}
\right]
\\
&
=
\log
\frac{p_{\theta}(z|x,a)}{p_{\theta}(z)}
-
\lim_{K\rightarrow\infty}
\log
\frac{\frac{p_{\theta}(z|x,a)}{p_{\theta}(z)}+K-1}{K}
\\[3ex]
&
=
\text{PMI}_{\theta}(x,a;z)
\end{align}

which completes the proof. \QED

Lastly, recall the following basic relationship:

\begin{relemma}[restate=thmmut,name=Conditional Mutual Information]\upshape\label{thm:mut}
Conditioned on any $x,a$, we have that:
\begin{align}
\mathbb{I}_{\theta}
[Y;Z|x,a]
=
\mathbb{H}
[Y|x,a]
+
\mathbb{H}_{\theta}
[Y|x,a,Z]
\end{align}
\end{relemma}

\textit{Proof}. Starting from the left hand side:
\begin{align}
\mathbb{I}_{\theta}
[Y;Z|x,a]
&
\coloneqq
\mathbb{E}_{Z\sim p_{\theta}}
D_{\text{KL}}
\big(
p_{\theta}(Y|x,a,Z)\|\tau(Y|x,a)
\big)
\\[2ex]
&
\spaceqq
\mathbb{E}_{\substack{Z\sim p_{\theta}\\Y\sim p_{\theta}(\cdot|x,a,Z)}}
\log p_{\theta}(Y|x,a,Z)
-
\mathbb{E}_{\substack{Z\sim p_{\theta}\\Y\sim p_{\theta}(\cdot|x,a,Z)}}
\tau(Y|x,a)
\\
&
\spaceqq
-
\textstyle\int_{\mathcal{Z}}
p_{\theta}(z)
\mathbb{H}_{\theta}
[Y|x,a,z]
dz
-
\mathbb{E}_{\substack{Y\sim\tau(\cdot|x,a)\\Z\sim p_{\theta}(\cdot|x,a,Y)}}
\tau(Y|x,a)
\\
&
\spaceqq
\mathbb{H}
[Y|x,a]
-
\mathbb{H}_{\theta}
[Y|x,a,Z]
\end{align}
which completes the proof. \QED

\subsection{Optimistic Exploration}

In our structural causal model, by construction $Z$ captures all sources of noise---that is, there is no residual noise in each outcome $Y$. However, for the purposes of optimization, while learning $\eta$ we let the residual error be captured by a Gaussian ``log-likelihood'' (note that $\lambda$ plays the role of ``$2\sigma^{2}$''):
\begin{align}
\log p_{\eta}(Y|x,a,z)
\coloneqq
-
\frac{1}{2}\log(\lambda\pi)
-
\frac{1}{\lambda}
\big(
Y-f_{\eta}(x,a,z)
\big)^{2}
\end{align}
and note that $\theta$ also induces a log-likelihood of the ``ground-truth'' conditional:
\begin{align}
\log
p_{\theta}(Y|x,a,z)
\coloneqq
\log
\frac
{p_{\theta}(z|x,a,Y)\tau(Y|x,a)\pi(a,x)\rho_{\pi}(x)}
{\int_{\mathcal{Y}}p_{\theta}(z|x,a,y)\tau(y|x,a)\pi(a|x)\rho_{\pi}(x)dy}
\end{align}
Now, recall the reconstruction loss and (state-action) reconstruction bonus:
\begin{align}
\mathcal{L}_{\eta}^{\text{rec.}}(x,a,z,y)
\coloneqq
\big\|
y-f_{\eta}(x,a,z)
\big\|_{2}^{2}
\end{align}

\begin{align}
\mathcal{R}_{\theta,\eta}^{\text{rec.}}(x,a)
\coloneqq
\mathbb{E}_{\substack{Y\sim\tau(\cdot|x,a)\\Z\sim p_{\theta}(\cdot|x,a,Y)}}
\mathcal{L}_{\eta}^{\text{rec.}}(x,a,Z,Y)
\end{align}

as well as the invariance loss and (state-action) invariance bonus:

\begin{align}
\mathcal{L}_{\theta}^{\text{inv.}}(x,a,z)
\coloneqq
\text{PMI}_{\theta}(x,a;z)
\end{align}

\begin{align}
\mathcal{R}_{\theta}^{\text{inv.}}(x,a)
\coloneqq
\mathbb{E}_{\substack{Y\sim\tau(\cdot|x,a)\\Z\sim p_{\theta}(\cdot|x,a,Y)}}
\mathcal{L}_{\theta}^{\text{inv.}}(x,a,Z)
\end{align}

We now show that Theorem \ref{thm:overall} is true, which we restate using our subscript-less notation:

\begin{retheorem}[restate=resoverall,name=Optimistic Exploration]\upshape\label{res:overall}
Let $\lambda$ satisfy the inequality
\smash{$
\frac{1}{2}\log(\lambda\pi)
\leq
$}
\smash{$
\mathbb{H}_{\theta}
[Y|x,a,Z]
+
D_{\text{KL}}
\big(
p_{\theta}(Z|x,a)\|p_{\theta}(Z)
\big)
$},
where $\pi$ denotes here the mathematical constant (not the agent's policy). Then:

\begin{align}
\mathcal{R}_{\theta,\eta}(x,a)
\geq
D_{\text{KL}}
\big(
\tau(Y|x,a)\|\tau_{\theta,\eta}(Y|x,a)
\big)
\end{align}

where
\smash{$\tau_{\theta,\eta}(Y|x,a)$\pix$\coloneqq$\pix$\mathbb{E}_{Z\sim p_{\theta}}p_{\eta}(Y|x,a,Z)$}
denotes the learned world model.
Furthermore, assuming realizability, rewards vanish at optimal parameters \smash{$\theta^{*},\eta^{*}$}:
\begin{align}
\mathcal{R}_{\theta^{*},\eta^{*}}(x,a)=0
\quad
\forall x,a\in\text{supp}(\rho_{\pi})
\end{align}
\end{retheorem}

\textit{Proof}.
Use Definition \ref{def:hindsight}, then the constraint on $\lambda$, then Lemma \ref{thm:mut}:

\begin{align}
\mathcal{R}_{\theta,\eta}(x,a)
&
\coloneqq
\pix
\frac{1}{\lambda}
\pix
\mathcal{R}_{\theta,\eta}^{\text{rec.}}(x,a)
+
\mathcal{R}_{\theta}^{\text{inv.}}(x,a)
\\[1ex]
&
\spaceqq
\mathbb{E}_{\substack{Y\sim\tau(\cdot|x,a)\\Z\sim p_{\theta}(\cdot|x,a,Y)}}
\frac{1}{\lambda}
\big(
Y-f_{\eta}(x,a,Z)
\big)^{2}
+
\mathbb{E}_{Z\sim p_{\theta}(\cdot|x,a)}
\text{PMI}_{\theta}(x,a;Z)
\\[-0.5ex]
&
\spaceqq
\mathbb{E}_{\substack{Y\sim\tau(\cdot|x,a)\\Z\sim p_{\theta}(\cdot|x,a,Y)}}
\frac{1}{\lambda}
\big(
Y-f_{\eta}(x,a,Z)
\big)^{2}
+
D_{\text{KL}}
\big(
p_{\theta}(Z|x,a)\|p_{\theta}(Z)
\big)
\\[0.5ex]
&
\spacegeqq
-
\mathbb{E}_{\substack{Y\sim\tau(\cdot|x,a)\\Z\sim p_{\theta}(\cdot|x,a,Y)}}
\log p_{\eta}(Y|x,a,Z)
-
\mathbb{H}_{\theta}
[Y|x,a,Z]
\\[0.5ex]
&
\spaceqq
-
\mathbb{E}_{\substack{Y\sim\tau(\cdot|x,a)\\Z\sim p_{\theta}(\cdot|x,a,Y)}}
\log p_{\eta}(Y|x,a,Z)
+
\mathbb{I}_{\theta}
[Y;Z|x,a]
-
\mathbb{H}
[Y|x,a]
\\[0.5ex]
&
\spaceqq
-
\mathbb{E}_{\substack{Y\sim\tau(\cdot|x,a)\\Z\sim p_{\theta}(\cdot|x,a,Y)}}
\log p_{\eta}(Y|x,a,Z)
\text{\scriptsize~$\leftarrow$~\textbf{remaining} stochasticity}
\nonumber
\\
&~~~~\pix\pix\pix
+
\mathbb{E}_{Y\sim\tau(\cdot|x,a)}
D_{\text{KL}}
\big(
p_{\theta}(Z|x,a,Y)\|p_{\theta}(Z|x,a)
\big)
\text{\scriptsize~$\leftarrow$~\textbf{hindsight} information}
\nonumber
\\[0.5ex]
&~~~~\pix\pix\pix
-
\mathbb{E}_{Y\sim\tau(\cdot|x,a)}
\big[
-
\log\tau(Y|x,a)
\big]
\text{\scriptsize~$\leftarrow$~\textbf{total} stochasticity}
\\[1ex]
&
\spacegeqq
-
\mathbb{E}_{\substack{Y\sim\tau(\cdot|x,a)}}
\big[~
\mathbb{E}_{Z\sim p_{\theta}(\cdot|x,a,Y)}
\log p_{\eta}(Y|x,a,Z)
\nonumber
\\[0.5ex]
&~~~~\pix\pix\pix
-
D_{\text{KL}}
\big(
p_{\theta}(Z|x,a,Y)\|p_{\theta}(Z|x,a)
\big)
\nonumber
+
D_{\text{KL}}
\big(
p_{\theta}(Z|x,a,Y)\|p_{\eta}(Z|x,a,Y)
\big)
\big]
\nonumber
\\[0.75ex]
&~~~~\pix\pix\pix
+
\mathbb{E}_{Y\sim\tau(\cdot|x,a)}
\log\tau(Y|x,a)
\\[1.5ex]
&
\spaceqq
-
\mathbb{E}_{Y\sim\tau(\cdot|x,a)}
\log
\mathbb{E}_{Z\sim p_{\theta}}p_{\eta}(Y|x,a,Z)
+
\mathbb{E}_{Y\sim\tau(\cdot|x,a)}
\log\tau(Y|x,a)
\\[1.5ex]
&
\spaceqq
-
\mathbb{E}_{Y\sim\tau(\cdot|x,a)}
\log
\tau_{\theta,\eta}(Y|x,a)
+
\mathbb{E}_{Y\sim\tau(\cdot|x,a)}
\log\tau(Y|x,a)
\\[1.5ex]
&
\spaceqq
D_{\text{KL}}
\big(
\tau(Y|x,a)\|\tau_{\theta,\eta}(Y|x,a)
\big)
\end{align}

For the second part, we want to show that for the objective:

\begin{equation}
J(\theta,\eta;\lambda)
\coloneqq
\mathbb{E}_{\substack{X\sim\rho_{\pi}\\A\sim\pi(\cdot|X)}}
\left[
\pix
\frac{1}{\lambda}
\pix
\mathcal{R}_{\theta,\eta}^{\text{rec.}}(X,A)
+
\mathcal{R}_{\theta}^{\text{inv.}}(X,A)
\right]
\end{equation}

its optimal value is zero:
\begin{equation}
\underset{\theta,\eta}{\text{min}}~
J(\theta,\eta;\lambda)
=
0
\end{equation}

Take any MDP. By reparameterization, we know that there exists an equivalent graphical representation under which $Z$ is exogenous. Assuming realizability, let $\eta^{*}$ be such that $f_{\eta^{*}}=f$, and let $\theta^{*}$ be such that $p_{\theta^{*}}(Z|x,a,y)=p_{\eta^{*}}(Z|x,a,y)$ for any $x,a,y$. First, by construction we have that $Z\perp X,A$, so the mutual information between $Z$ and $X,A$ must be zero:

\begin{align}
\mathbb{E}_{\substack{X\sim\rho_{\pi}\\A\sim\pi(\cdot|X)}}
\mathcal{R}_{\theta}^{\text{inv.}}(X,A)
&
=
\mathbb{E}
\hspace{-3pt}
_{\substack{X\sim\rho_{\pi}\\A\sim\pi(\cdot|X)\\Z\sim p_{\theta}(\cdot|X,A)}}
\hspace{-3pt}
\text{PMI}_{\theta}(X,A;Z)
\\
&
=
\mathbb{I}_{\theta}[X,A;Z]
\\[3ex]
&
=
0
\end{align}

Second, by consistency of counterfactuals $f_{\eta^{*}}(x,a,Z)=y$ for any $Z$\pix$\sim$\pix$p_{\theta^{*}}(\cdot|x,a,y)$, so the reconstruction term is also zero, which completes the proof. \QED

The intuition is as follows: Assuming realizability, at convergence ``hindsight information'' and ``total stochasticity'' cancel (i.e. neither more nor less), and the ``remaining stochasticity'' term goes to zero.

\subsection{Optimal Invariance}

We now show that Theorem \ref{thm:inv} is true, which we restate using our subscript-less notation:

\begin{retheorem}[restate=resinv,name=Optimal Invariance]\upshape\label{res:inv}
The contrastive bonus lower-bounds the (ideal) invariance bonus for any pair $x,a$:

\begin{align}
\mathcal{R}_{\theta,\nu}^{K,\text{con.}}(x,a)
\leq
\mathcal{R}_{\theta}^{\text{inv.}}(x,a)
\end{align}

Furthermore, assuming realizability, for optimal critic parameter $\nu_{K}^{*}
\coloneqq
\text{arg}\pix\text{max}_{\nu}~
\mathbb{E}
_{X,A\sim\rho_{\pi}}
\mathcal{R}_{\theta,\nu}^{K,\text{con.}}(X,A)
$ the bound is asymptotically tight (in the batch size $K\rightarrow\infty$):
\begin{align}
\lim_{K\rightarrow\infty}
\mathcal{R}_{\theta,\nu_{K}^{*}}^{K,\text{con.}}&(x,a)
=
\mathcal{R}_{\theta}^{\text{inv.}}(x,a)
\end{align}

\end{retheorem}

\textit{Proof}. Use Lemma \ref{thm:info} for the first part:

\begin{align}
\mathcal{R}_{\theta,\nu}^{K,\text{con.}}(x,a)
&
\coloneqq
\mathbb{E}_{Z\sim p_{\theta}(\cdot|x,a)}
\mathcal{L}_{\theta,\nu}^{K,\text{con.}}(x,a,Z)
\\
&
\spaceqq
\mathbb{E}_{\substack{Z\sim p_{\theta}(\cdot|x,a)\\Z_{1:K-1}\sim p_{\theta}^{K-1}}}
\log
\frac
{e^{g_{\nu}(x,a,Z)}}
{\frac{1}{K}\left(e^{g_{\nu}(x,a,Z)}+\sum_{i=1}^{K-1}e^{g_{\nu}(x,a,Z_{i})}\right)}
\\
&
\spaceleqq
\mathbb{E}_{Z\sim p_{\theta}(\cdot|x,a)}
\text{PMI}_{\theta}(x,a;Z)
\\[3ex]
&
\eqqcolon
\mathcal{R}_{\theta}^{\text{inv.}}(x,a)
\end{align}

and use Lemma \ref{thm:opt} for the second part:

\begin{align}
\lim_{K\rightarrow\infty}
\mathcal{R}_{\theta,\nu_{K}^{*}}^{K,\text{con.}}(x,a)
&
\coloneqq
\lim_{K\rightarrow\infty}
\mathbb{E}_{Z\sim p_{\theta}(\cdot|x,a)}
\mathcal{L}_{\theta,\nu_{K}^{*}}^{K,\text{con.}}(x,a,Z)
\\[1.5ex]
&
\spaceqq
\mathbb{E}_{Z\sim p_{\theta}(\cdot|x,a)}
\lim_{K\rightarrow\infty}
\mathcal{L}_{\theta,\nu_{K}^{*}}^{K,\text{con.}}(x,a,Z)
\\[1.5ex]
&
\spaceqq
\mathbb{E}_{Z\sim p_{\theta}(\cdot|x,a)}
\text{PMI}_{\theta}(x,a;Z)
\\[2.5ex]
&
\eqqcolon
\mathcal{R}_{\theta}^{\text{inv.}}(x,a)
\end{align}
which completes the proof. \QED
\newpage

\section{Further Experiment Results}\label{app:b}

\subsection{Sticky Actions}\label{subapp:sticky}

\begin{figure*}[h!]
\vspace{-0.75em}
\centering
\makebox[1.0\textwidth][c]{
\hspace{-12px}
\subfloat[\textbf{Sticky Actions (Intrinsic Only)}]{
\includegraphics[height=0.19\linewidth, trim=0em 0em 0em 0em]
{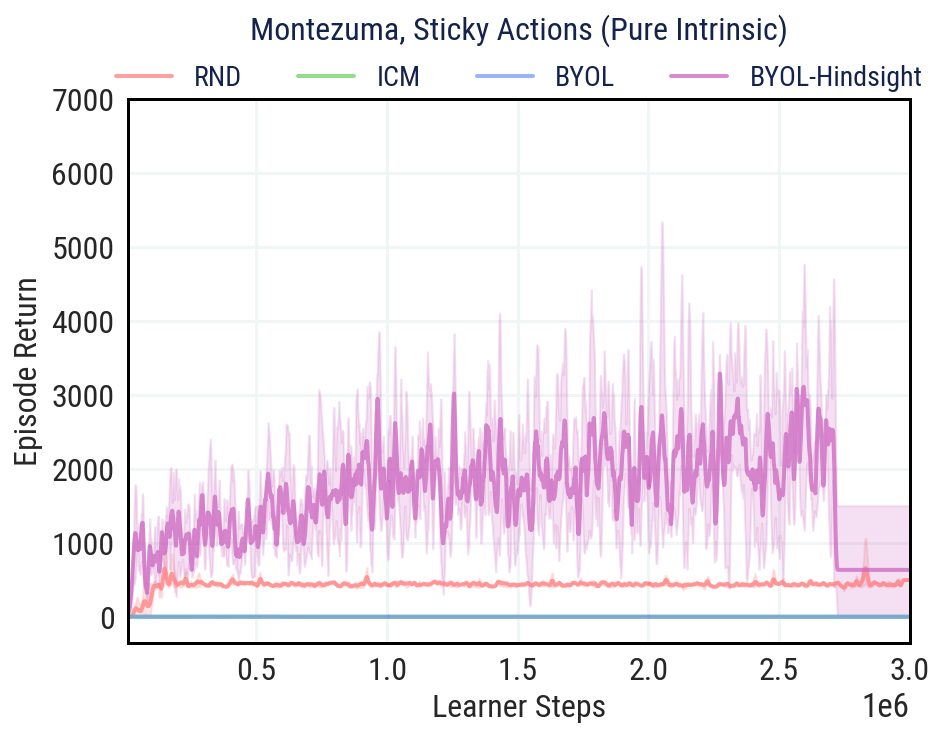}}
\hfill
\subfloat[\textbf{Sticky Actions (Ext. + Intrinsic)}]{
\includegraphics[height=0.19\linewidth, trim=0em 0em 0em 0em]
{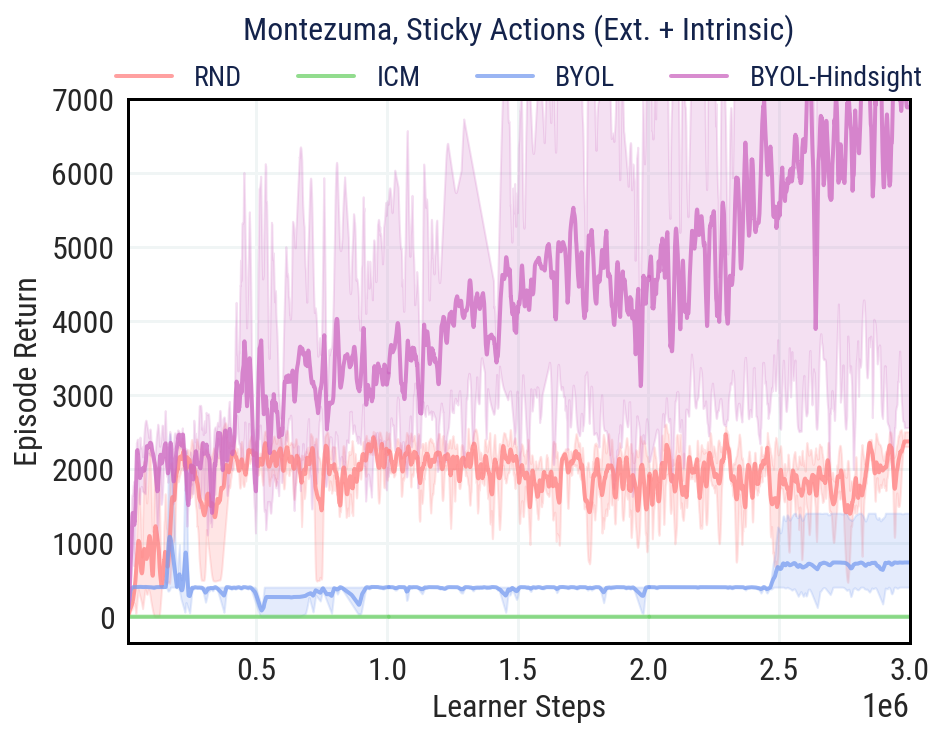}}
\hfill
\subfloat[\textbf{Non-Sticky Baseline (Intrinsic Only)}]{
\includegraphics[height=0.19\linewidth, trim=0em 0em 0em 0em]
{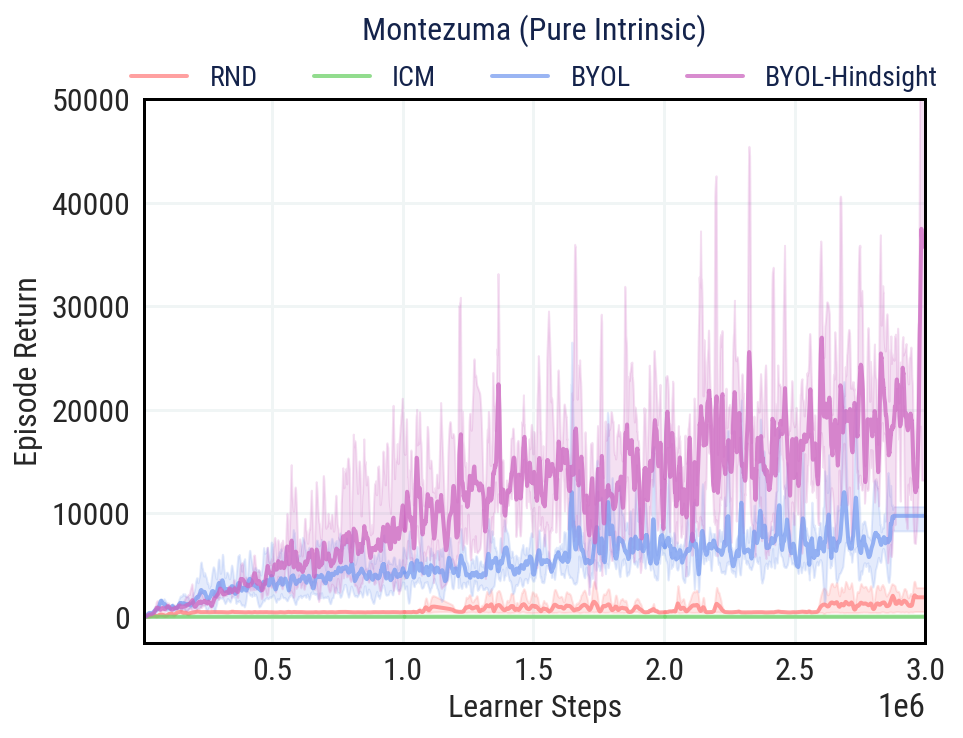}}
\hfill
\subfloat[\textbf{Non-Sticky Baseline (Ext. + Intrinsic)}]{
\includegraphics[height=0.19\linewidth, trim=0em 0em 0em 0em]
{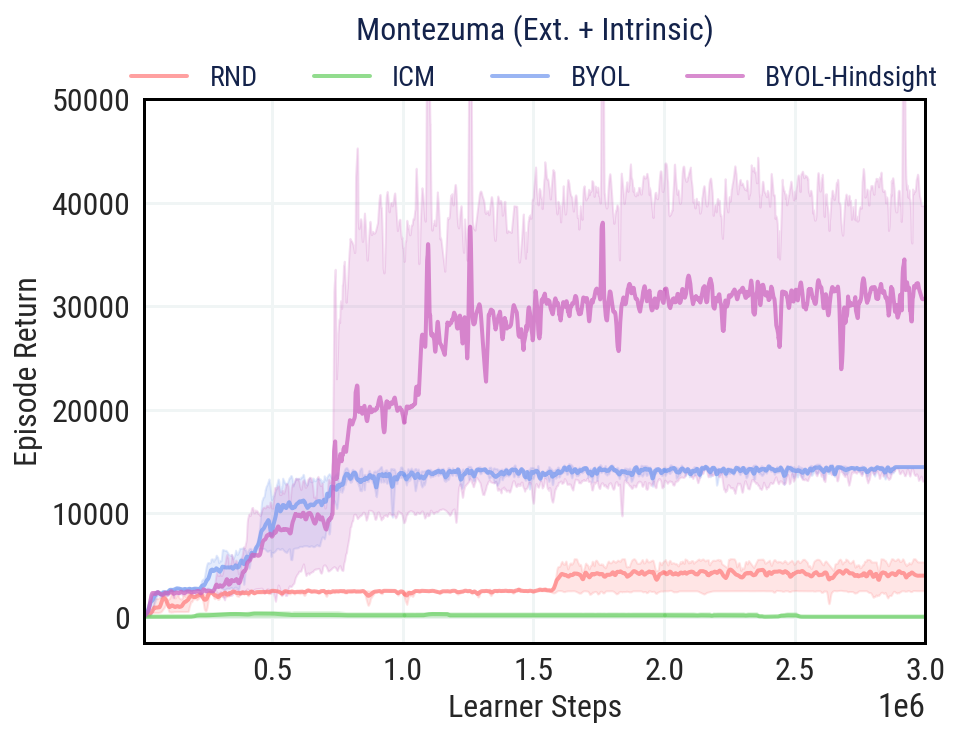}}
}
\vspace{-0.5em}
\caption{\small\dayum{\textit{Montezuma's Revenge, with Sticky Actions and Non-Sticky Baseline}.
Performance measured by extrinsic rewards in an episode.
}}
\label{fig:montezuma:sticky:return}
\vspace{-0.25em}
\end{figure*}

Figure \ref{fig:montezuma:sticky:room} showed exploration as measured by the number of different dungeon rooms the agent manages to discover. For completeness, Figure \ref{fig:montezuma:sticky:return} here also shows a comparison using the extrinsic reward that the agent obtains as a proxy.
The conclusions are similar: In ``Sticky Actions'', \tttext{BYOL-Explore} instantly flatlines in the intrinsic-only regime, and only does marginally better in the mixed regime. In contrast, \tttext{BYOL-Hindsight} achieves much higher scores in both regimes.
Moreover, in the ``Non-Sticky Baseline'', in both regimes \tttext{BYOL-Hindsight} actually manages to do even better than the original performance of \tttext{BYOL-Explore}.

\subsection{Persistive Noise}\label{subapp:persistive}

While sticky actions is the standard protocol for adding stochasticity, we further design an especially challenging form of stochasticity, as follows.
First, recall that each observation $O_{t}$ in our Atari environment is a grayscale 84$\times$84-pixel image. In the ``Persistive Noise'' setting, observations are corrupted by an additive layer of 84$\times$84-pixel noise that persists across time, with each pixel evolving randomly according to distributions that depend on the actions selected by the agent at each time step. Specifically, the value of each pixel $(i,j)\in\mathbb{N}_{<84}^{2}$ of the (final) observation $\tilde{O}_{t}$ is computed as follows:
\begin{equation}
\tilde{O}_{t}^{(i,j)}
\coloneqq
O_{t}^{(i,j)}+U_{t}^{(i,j)}\mod 256
\end{equation}

where $U_{t}$ is the persistive layer of pixel noise:
\begin{equation}
U_{t}^{(i,j)}=U_{t-1}^{(i,j)}+\epsilon_{t}^{(i,j)}\mod 50
\end{equation}

and noise steps are sampled as:
\begin{alignat}{2}
&\epsilon_{t}^{(i,j)}\sim\text{Uniform}\{-1,1\}\quad
&&\text{if}\quad\tttext{key}(a_{t-1})~\text{is odd}
\\
&\epsilon_{t}^{(i,j)}\sim\text{Uniform}\{-11,11\}\quad
&&\text{if}\quad\tttext{key}(a_{t-1})~\text{is even}
\end{alignat}

where $\tttext{key}(a)$ is the numerical key code associated with action $a$. See Figure \ref{fig:pixelframes} for example frames generated by this process.

\begin{figure*}[b!]
\vspace{-1.75em}
\centering
\makebox[1.0\textwidth][c]{
\hspace{-12px}
\subfloat[\textbf{Persistive Noise (Intrinsic Only)}]{
\includegraphics[height=0.19\linewidth, trim=0em 0em 0em 0em]
{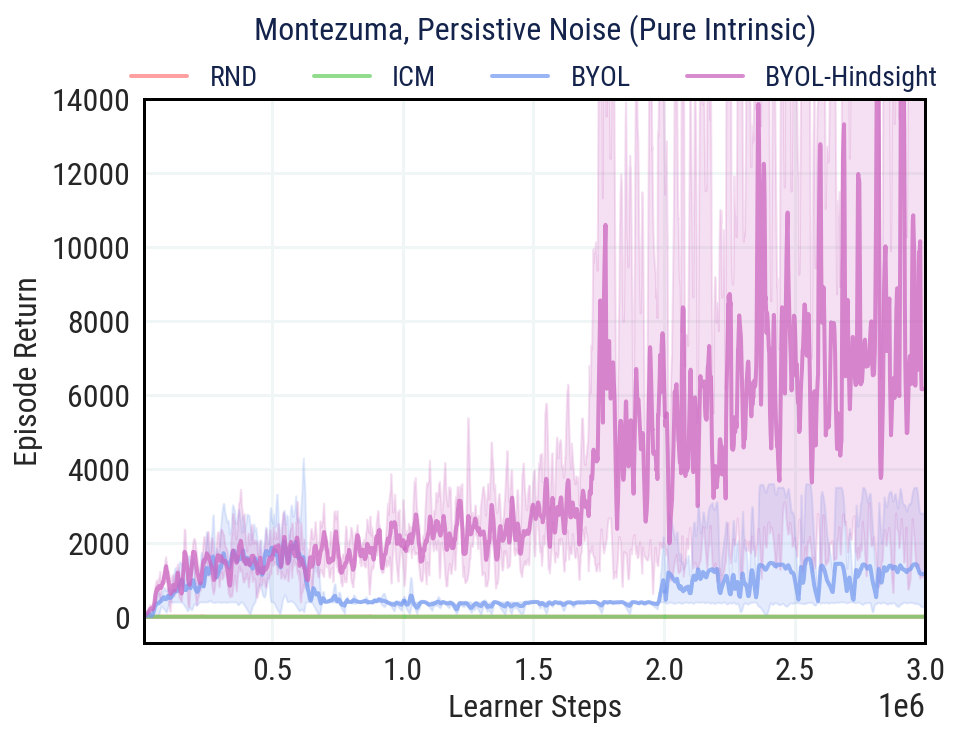}}
\hfill
\subfloat[\textbf{Persistive Noise (Ext. + Intrinsic)}]{
\includegraphics[height=0.19\linewidth, trim=0em 0em 0em 0em]
{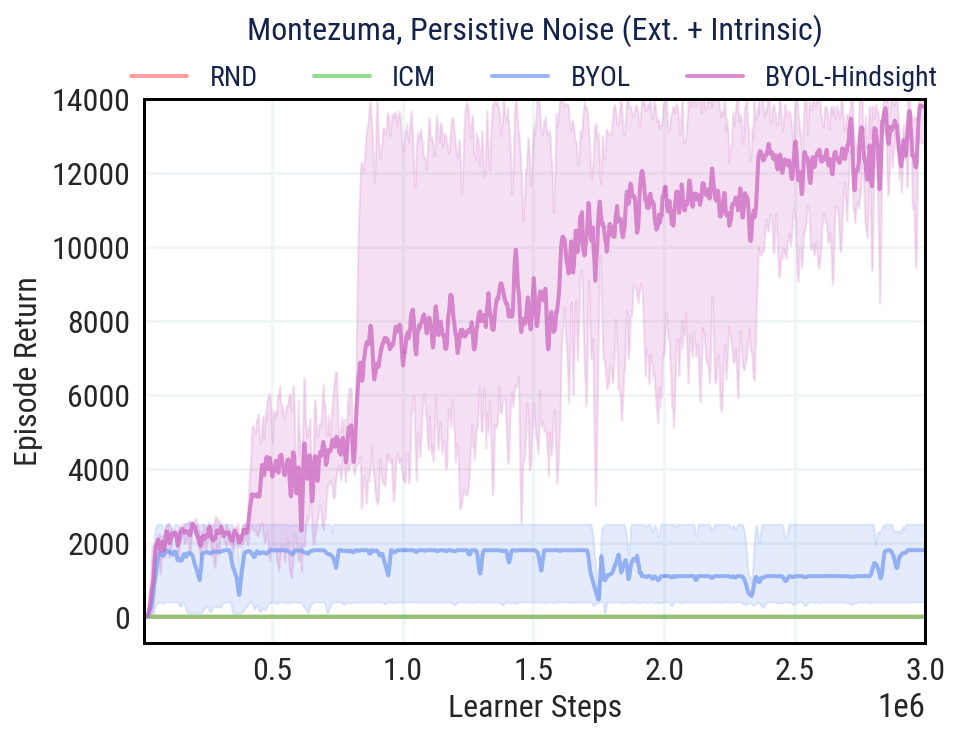}}
\hfill
\subfloat[\textbf{Non-Noisy Baseline (Intrinsic Only)}]{
\includegraphics[height=0.19\linewidth, trim=0em 0em 0em 0em]
{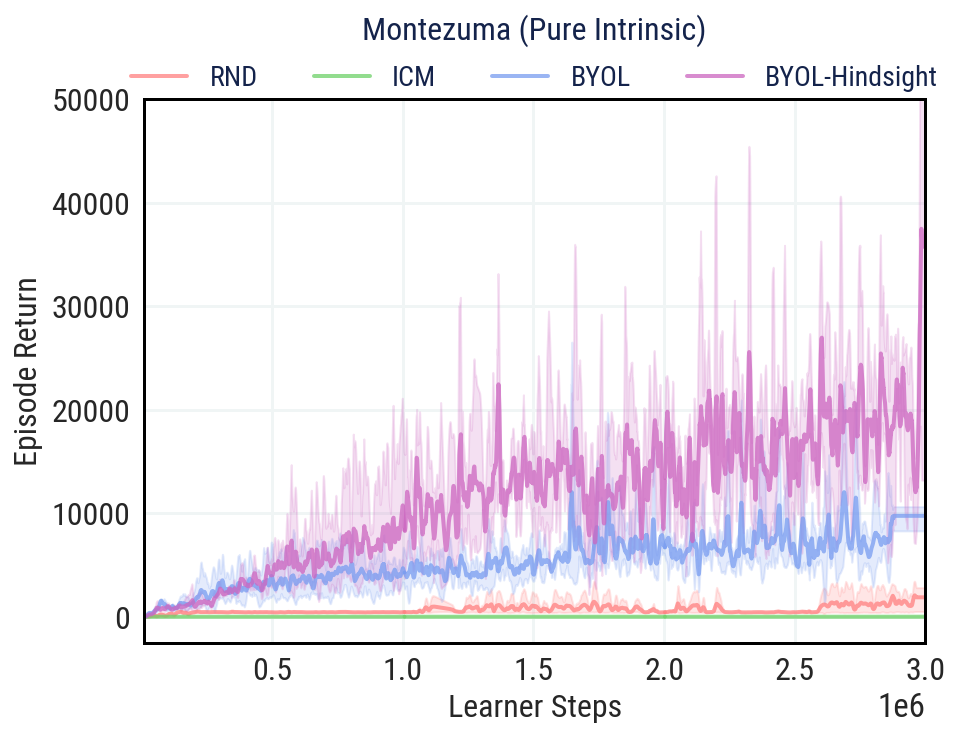}}
\hfill
\subfloat[\textbf{Non-Noisy Baseline (Ext. + Intrinsic)}]{
\includegraphics[height=0.19\linewidth, trim=0em 0em 0em 0em]
{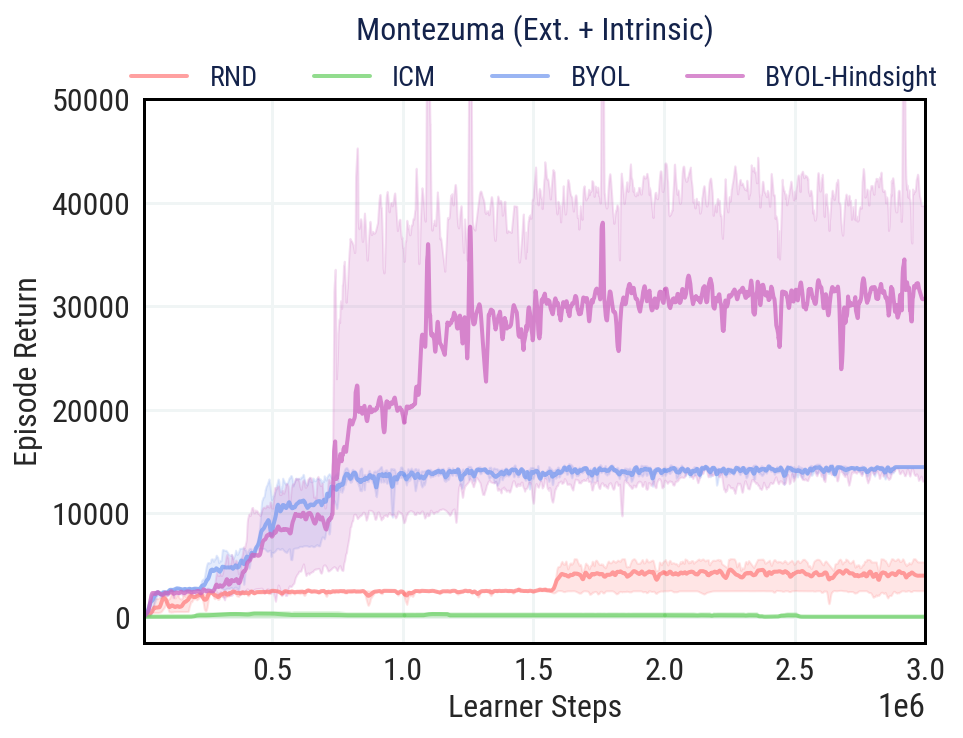}}
}
\vspace{-0.5em}
\caption{\small\dayum{\textit{Montezuma's Revenge, with Persistive Noise / Non-Noisy Baseline}.
Performance measured by extrinsic rewards in an episode.
}}
\label{fig:montezuma:pixel:return}
\vspace{-0.5em}
\end{figure*}

\begin{figure*}[h!]
\centering
\makebox[1.0\textwidth][c]{
\hspace{-7px}
\subfloat[\textbf{Persistive Noise (Intrinsic Only)}]{
\includegraphics[height=0.195\linewidth, trim=0em 0em 0em 0em]
{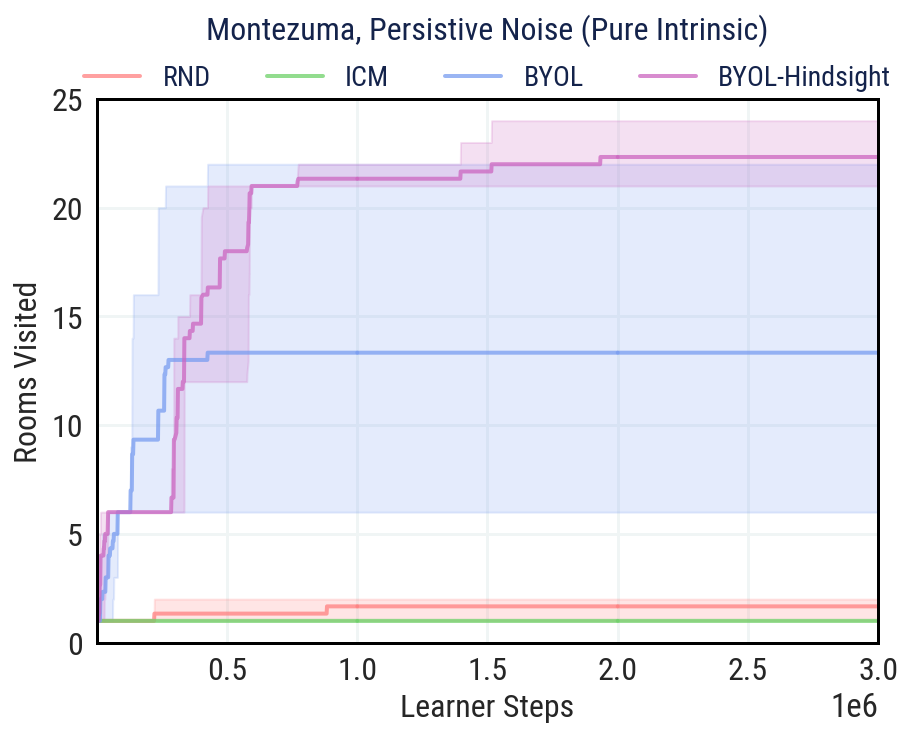}}
\hfill
\subfloat[\textbf{Persistive Noise (Ext. + Intrinsic)}]{
\includegraphics[height=0.195\linewidth, trim=0em 0em 0em 0em]
{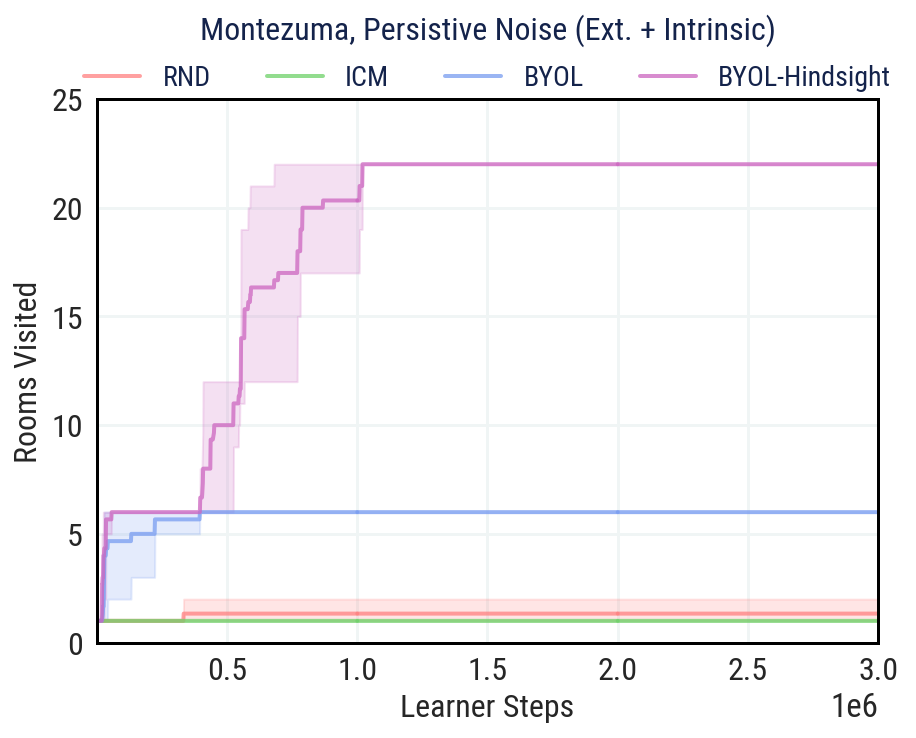}}
\hfill
\subfloat[\textbf{Non-Noisy Baseline (Intrinsic Only)}]{
\includegraphics[height=0.195\linewidth, trim=0em 0em 0em 0em]
{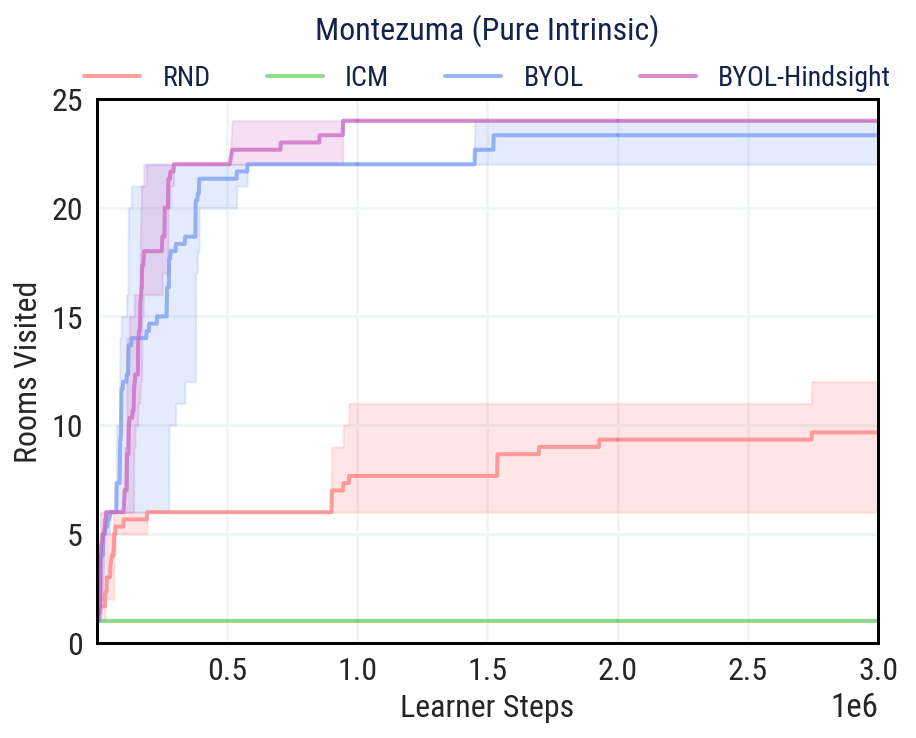}}
\hfill
\subfloat[\textbf{Non-Noisy Baseline (Ext. + Intrinsic)}]{
\includegraphics[height=0.195\linewidth, trim=0em 0em 0em 0em]
{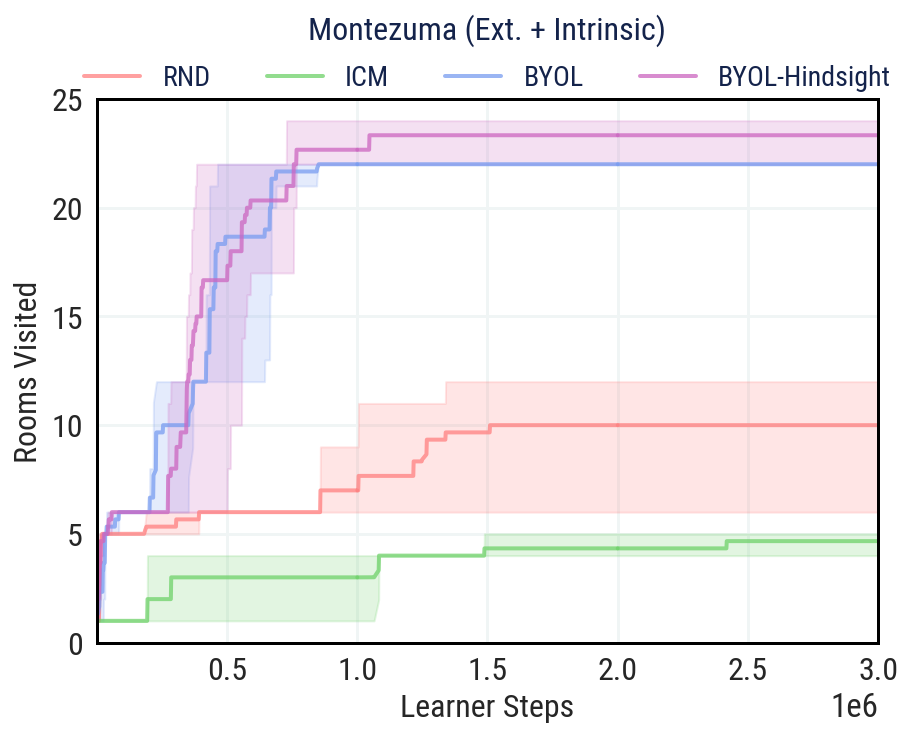}}
}
\vspace{-0.5em}
\caption{\small\dayum{\textit{Montezuma's Revenge, with Persistive Noise / Non-Noisy Baseline}.
Performance measured by rooms reached (episodic setting).
}}
\label{fig:montezuma:pixel:room}
\vspace{-0.75em}
\end{figure*}

Note that this setting is particularly challenging: While prior works have experimented with pixel-level noise (e.g. \cite{pathak2017curiosity,guo2022byol}), they have either designed noise that is not additive (e.g. there are separate channels of pixels to the original frame) or not persistive (e.g. each time step's noise does not depend on the previous time step's noise)---which means it is in principle easy for a learned representation to simply ``ignore'' the noise. This is not possible in our setting.

Figures \ref{fig:montezuma:pixel:return} and \ref{fig:montezuma:pixel:room} show results (3M learner steps, 3 seeds), in terms of episode return as well as rooms visited. In addition to ``Persistive Noise'' as just described, the results for the vanilla environment are shown again for reference (``Non-Noisy Baseline''). The conclusion is as suspected: \tttext{BYOL-Explore} suffers greatly from the presence of this stochasticity, whereas \tttext{BYOL-Hindsight} is more resilient to it.

\subsection{Temperature Sensitivity}\label{subapp:temperature}

The inner term in the contrastive loss (Objective \ref{obj:con}) shares a similar form with contrastive losses in unsupervised representation learning \cite{oord2018representation,tschannen2019mutual,he2020momentum,chen2020simple}, which admits a temperature (hyper-)parameter controlling the strength of penalties on negative samples \cite{wang2021understanding}. A valid question is: How sensitive is \tttext{BYOL-Hindsight} to the specific choice of value for this parameter?

\begin{figure*}[h!]
\vspace{-0.75em}
\centering
\makebox[1.0\textwidth][c]{
\hspace{-7px}
\subfloat[\textbf{Sticky Actions (Intrinsic Only)}]{
\includegraphics[height=0.195\linewidth, trim=0em 0em 0em 0em]
{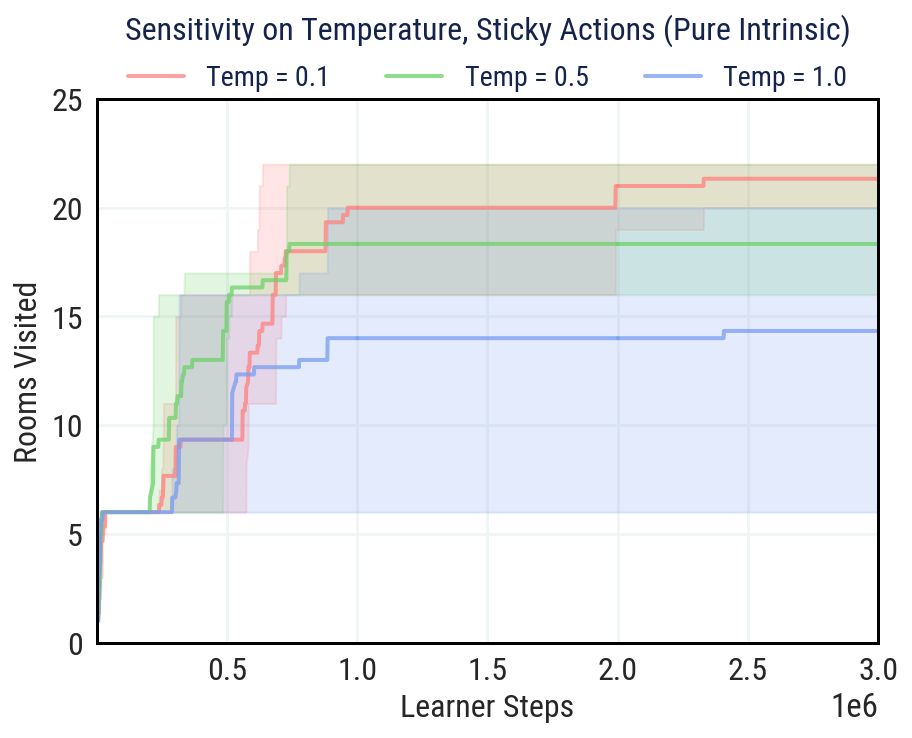}}
\hfill
\subfloat[\textbf{Sticky Actions (Ext. + Intrinsic)}]{
\includegraphics[height=0.195\linewidth, trim=0em 0em 0em 0em]
{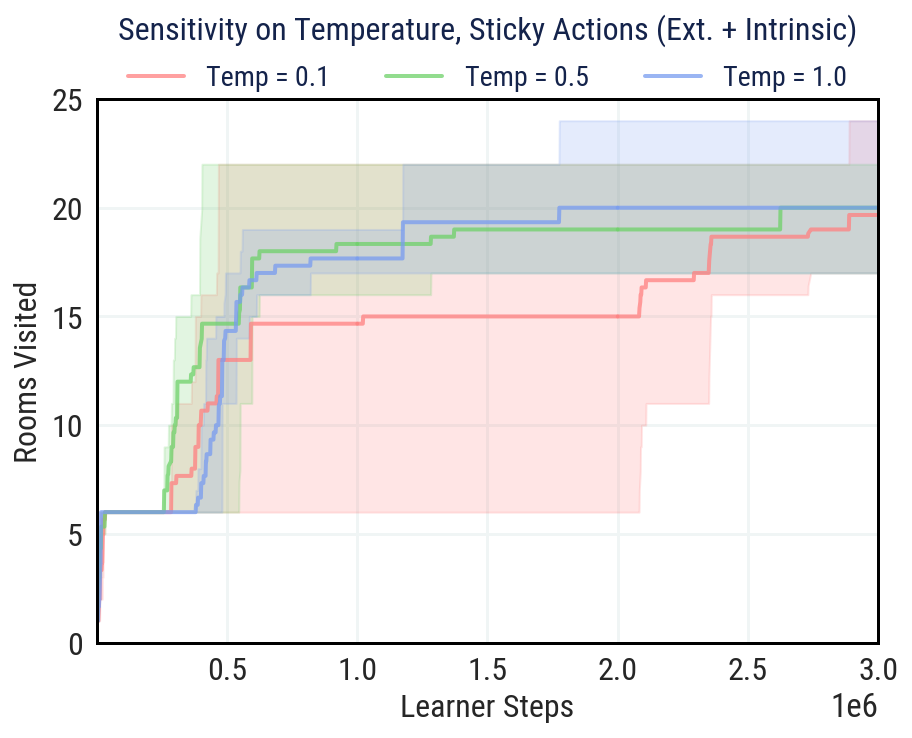}}
\hfill
\subfloat[\textbf{Non-Sticky Baseline (Intrinsic Only)}]{
\includegraphics[height=0.195\linewidth, trim=0em 0em 0em 0em]
{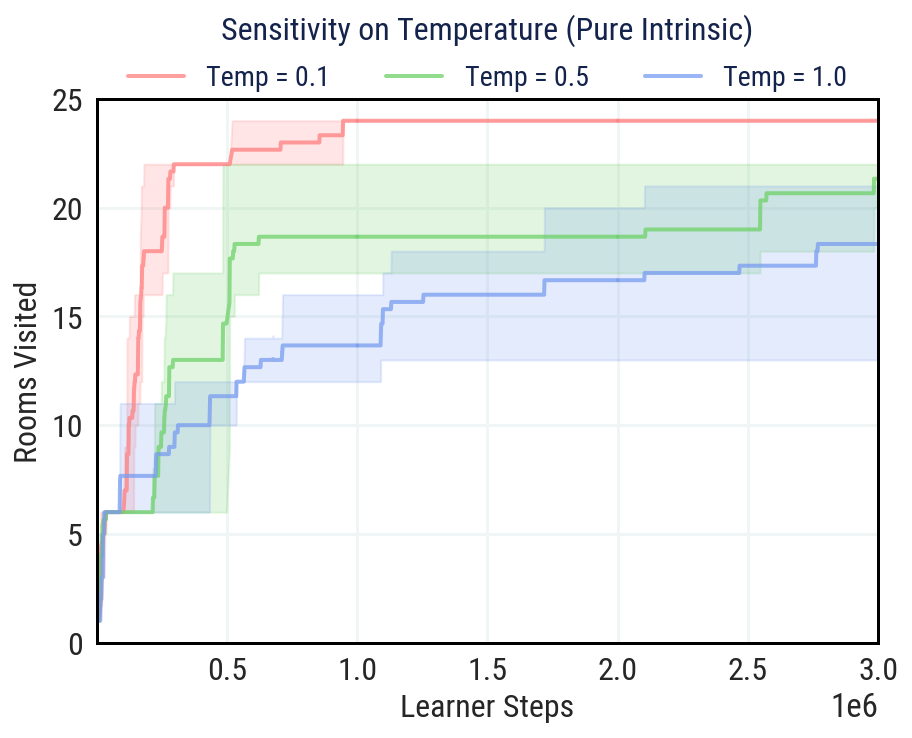}}
\hfill
\subfloat[\textbf{Non-Sticky Baseline (Ext. + Intrinsic)}]{
\includegraphics[height=0.195\linewidth, trim=0em 0em 0em 0em]
{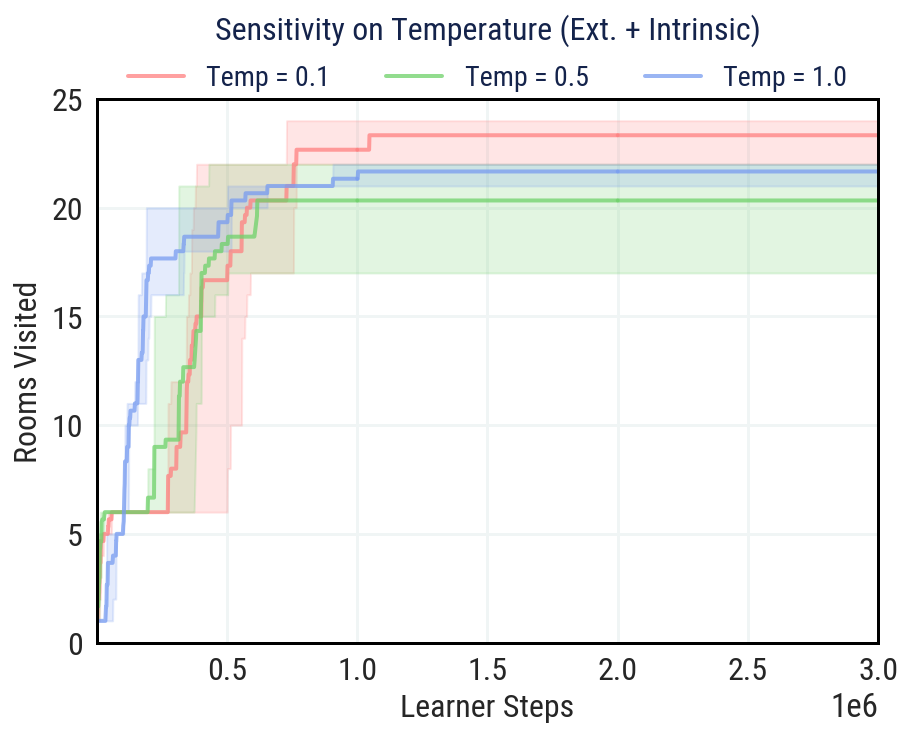}}
}
\vspace{-0.5em}
\caption{\small\dayum{\textit{Temperature Sensitivity (Montezuma's Revenge, with Sticky Actions and Non-Sticky Baseline)}.
Rooms reached (episodic setting).
}}
\label{fig:montezuma:temperature:room}
\vspace{-0.25em}
\end{figure*}

\begin{figure*}[h!]
\vspace{-0.75em}
\centering
\makebox[1.0\textwidth][c]{
\hspace{-12px}
\subfloat[\textbf{Sticky Actions (Intrinsic Only)}]{
\includegraphics[height=0.19\linewidth, trim=0em 0em 0em 0em]
{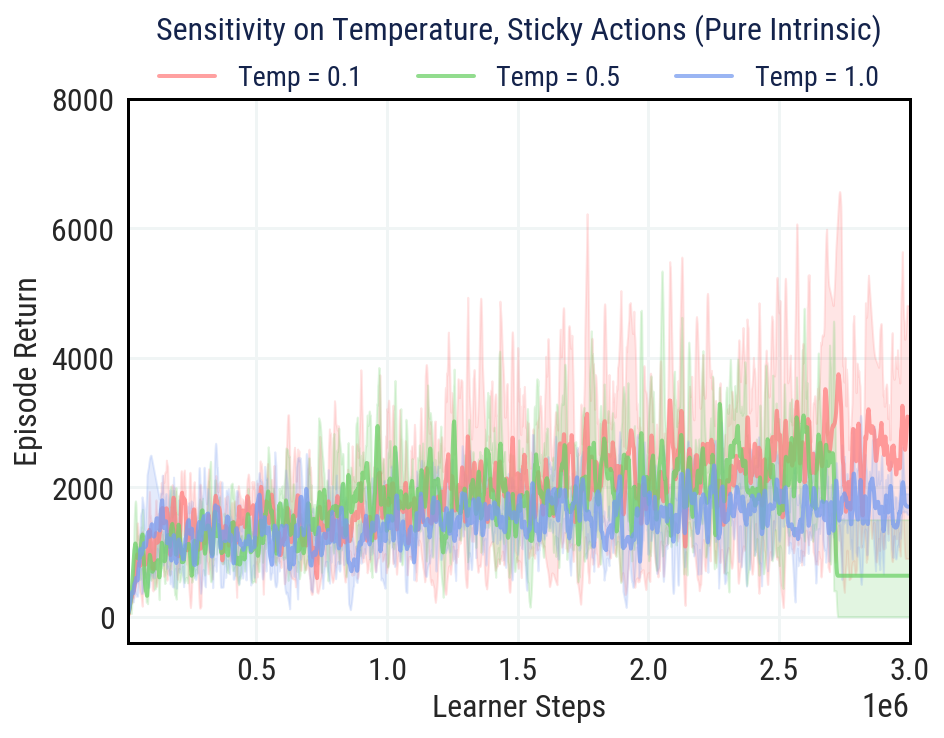}}
\hfill
\subfloat[\textbf{Sticky Actions (Ext. + Intrinsic)}]{
\includegraphics[height=0.19\linewidth, trim=0em 0em 0em 0em]
{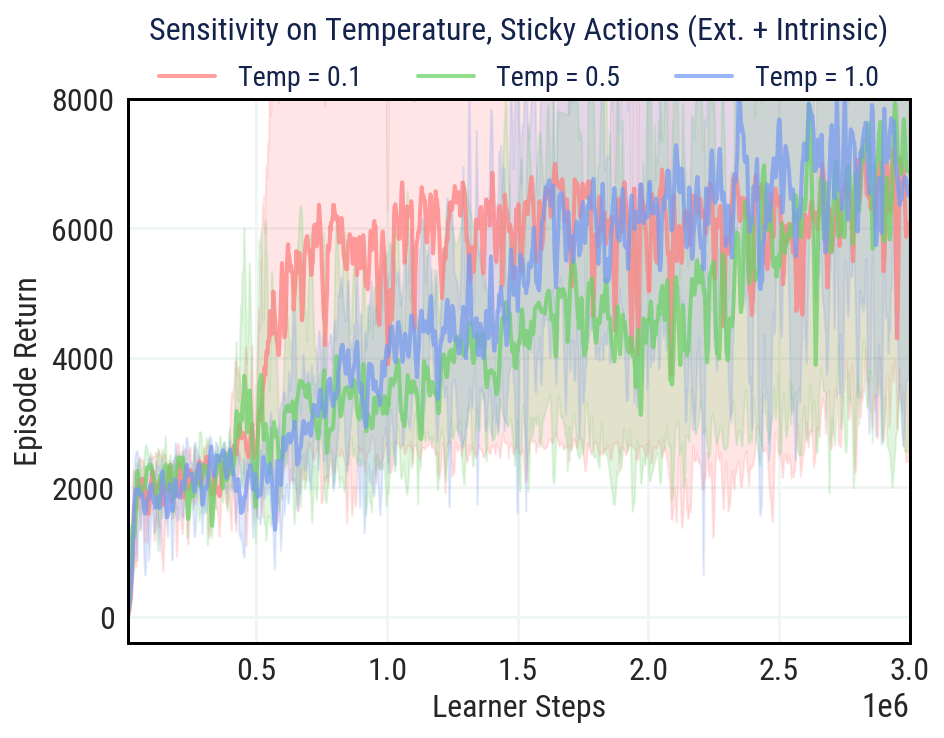}}
\hfill
\subfloat[\textbf{Non-Sticky Baseline (Intrinsic Only)}]{
\includegraphics[height=0.19\linewidth, trim=0em 0em 0em 0em]
{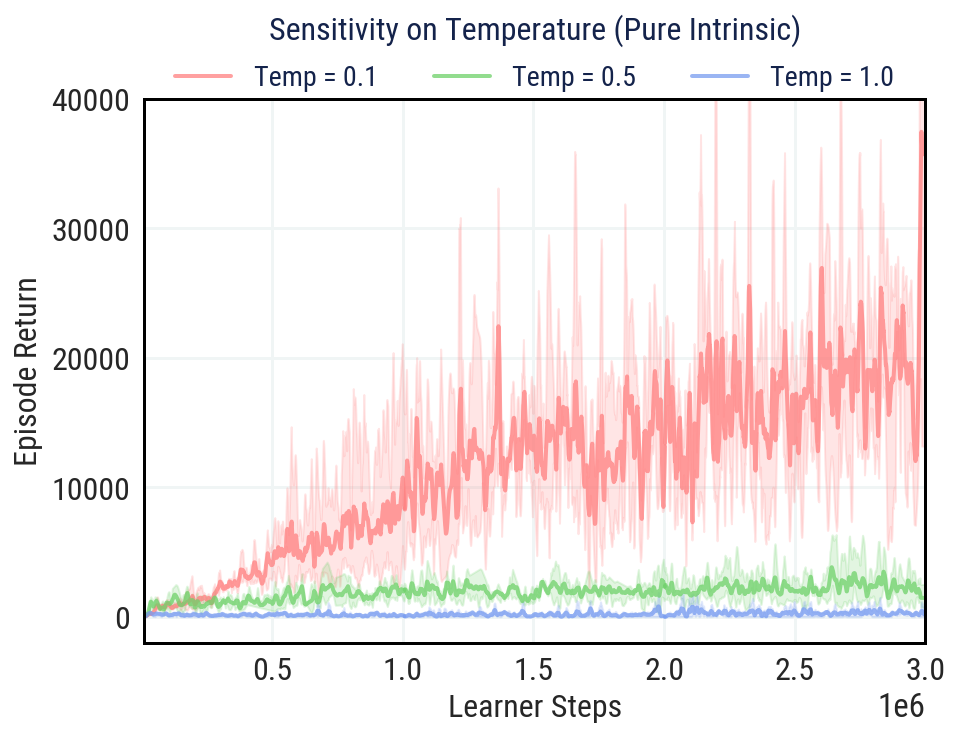}}
\hfill
\subfloat[\textbf{Non-Sticky Baseline (Ext. + Intrinsic)}]{
\includegraphics[height=0.19\linewidth, trim=0em 0em 0em 0em]
{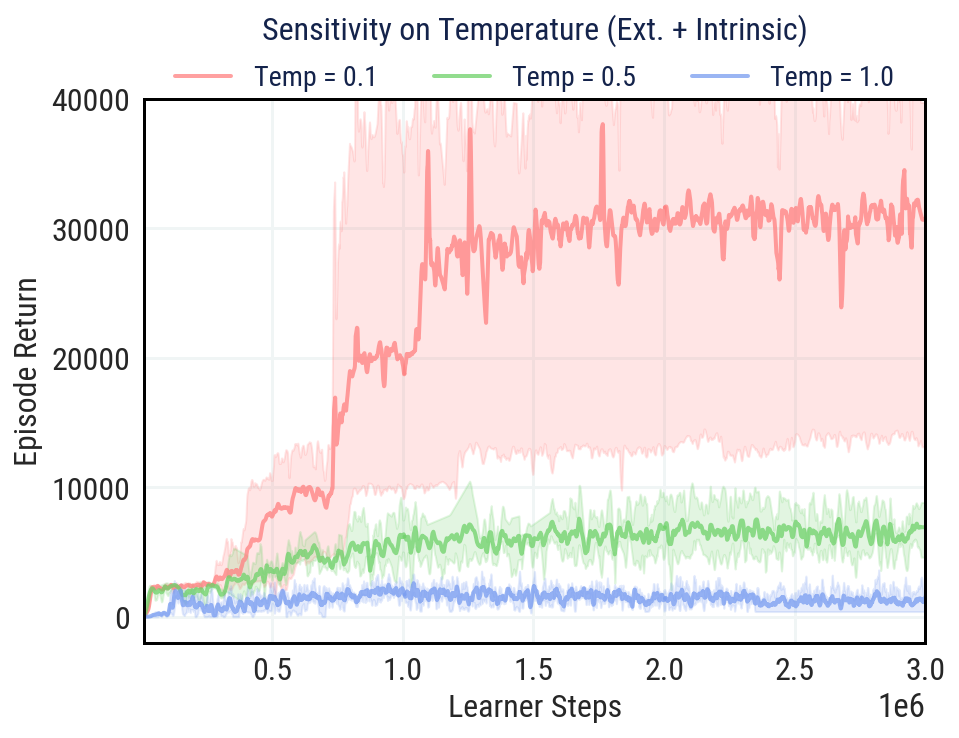}}
}
\vspace{-0.5em}
\caption{\small\dayum{\textit{Temperature Sensitivity (Montezuma's Revenge, with Sticky Actions and Non-Sticky Baseline)}.
Extrinsic rewards in an episode.
}}
\label{fig:montezuma:temperature:return}
\vspace{-0.25em}
\end{figure*}

Figures \ref{fig:montezuma:temperature:return} and \ref{fig:montezuma:temperature:room} show results (3M learner steps, 3 seeds), in terms of episode return as well as rooms visited, for ``Sticky Actions'' and ``Non-Sticky Baseline''. Interestingly, the performance of \tttext{BYOL-Hindsight} in the ``Sticky Actions'' setting is only mildly sensitive to the choice of temperature. In the ``Non-Sticky Baseline'' setting, sensitivity is most acute in the intrinsic-only exploration regime: Lower temperature performs better than higher temperature. This makes sense, because in a largely deterministic environment, a lower temperature provides a stronger incentive for invariances to be enforced, whereas a higher temperature may allow more leakage of information from $X,A$ into $Z$, which may diminish exploration.

\subsection{Intrinsic Rewards}\label{subapp:intrinsic}

The derivation of Theorem \ref{res:overall} makes it clear that ``hindsight information'' and ``total stochasticity'' can be misaligned in two ways: First, hindsight may capture less information than in the ``true'' latent (which means the reconstruction error is not driven to zero). Second, hindsight may capture more information than in the ``true'' latent (which means it leaks some information about the current state and action).

\begin{figure*}[h!]
\vspace{-0.75em}
\centering
\makebox[1.0\textwidth][c]{
\hspace{-12px}
\subfloat[\textbf{Sticky Actions (Intrinsic Only)}]{
\includegraphics[height=0.191\linewidth, trim=0em 0em 0em 0em]
{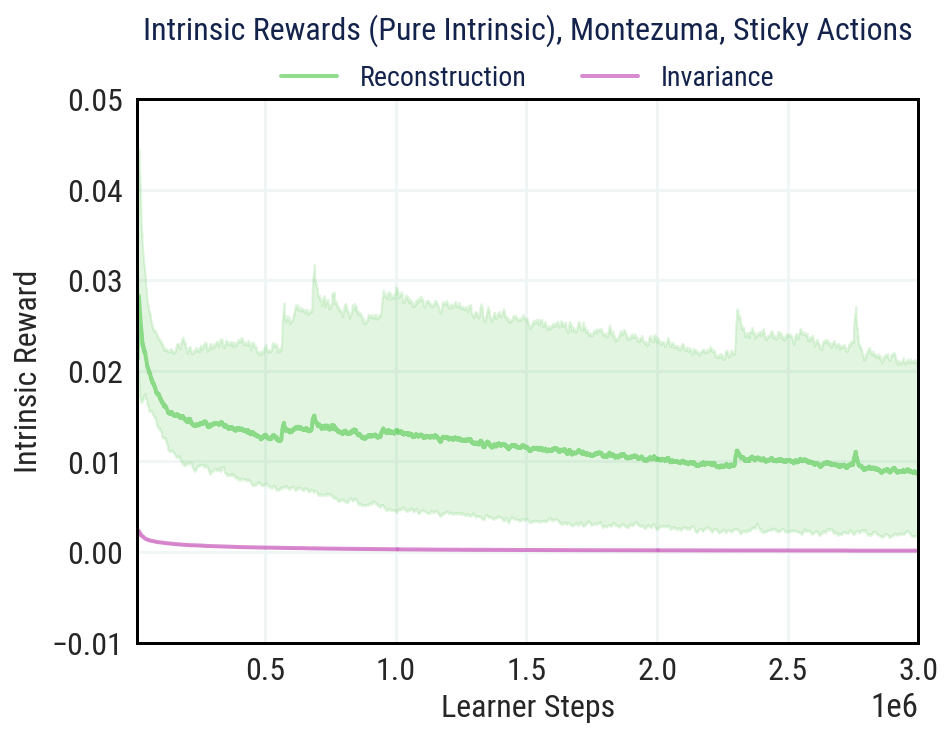}}
\hfill
\subfloat[\textbf{Sticky Actions (Ext. + Intrinsic)}]{
\includegraphics[height=0.191\linewidth, trim=0em 0em 0em 0em]
{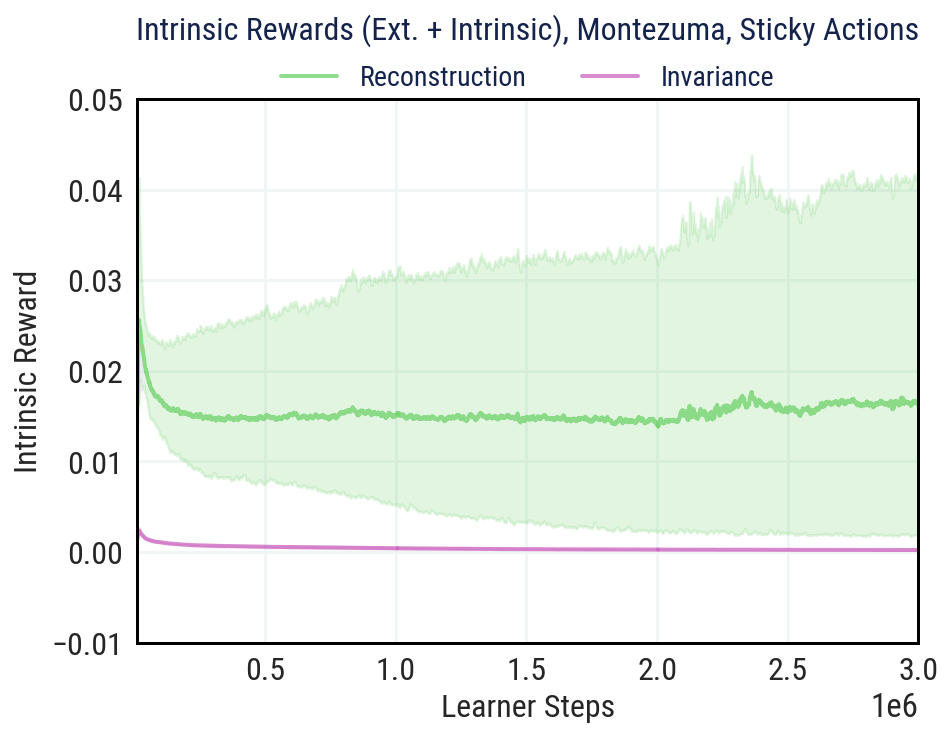}}
\hfill
\subfloat[\textbf{Non-Sticky Baseline (Intrinsic Only)}]{
\includegraphics[height=0.191\linewidth, trim=0em 0em 0em 0em]
{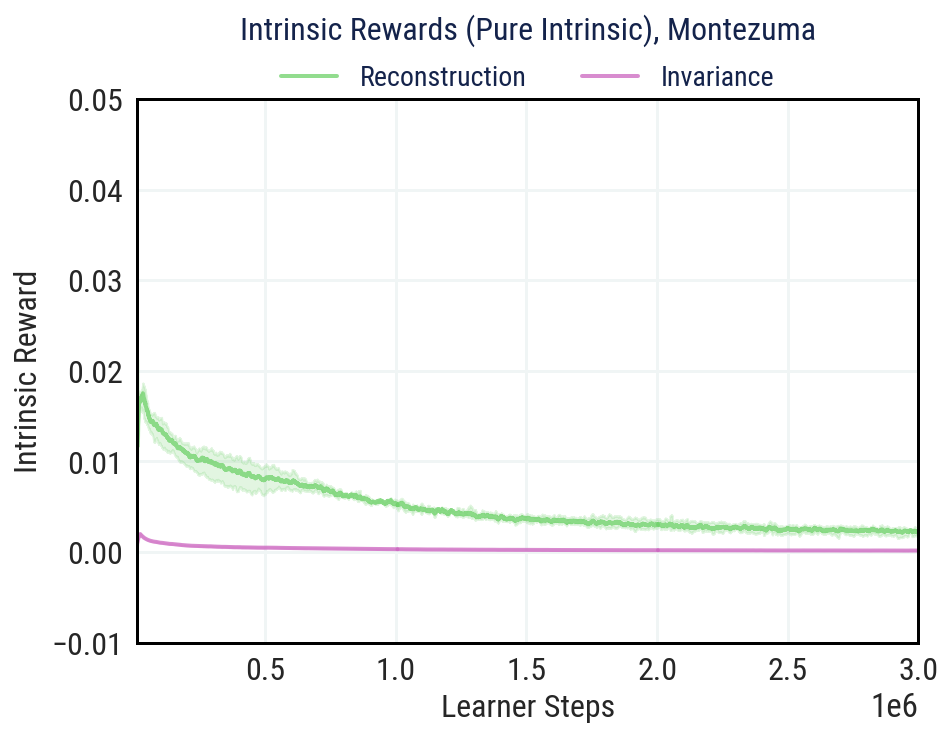}}
\hfill
\subfloat[\textbf{Non-Sticky Baseline (Ext. + Intrinsic)}]{
\includegraphics[height=0.191\linewidth, trim=0em 0em 0em 0em]
{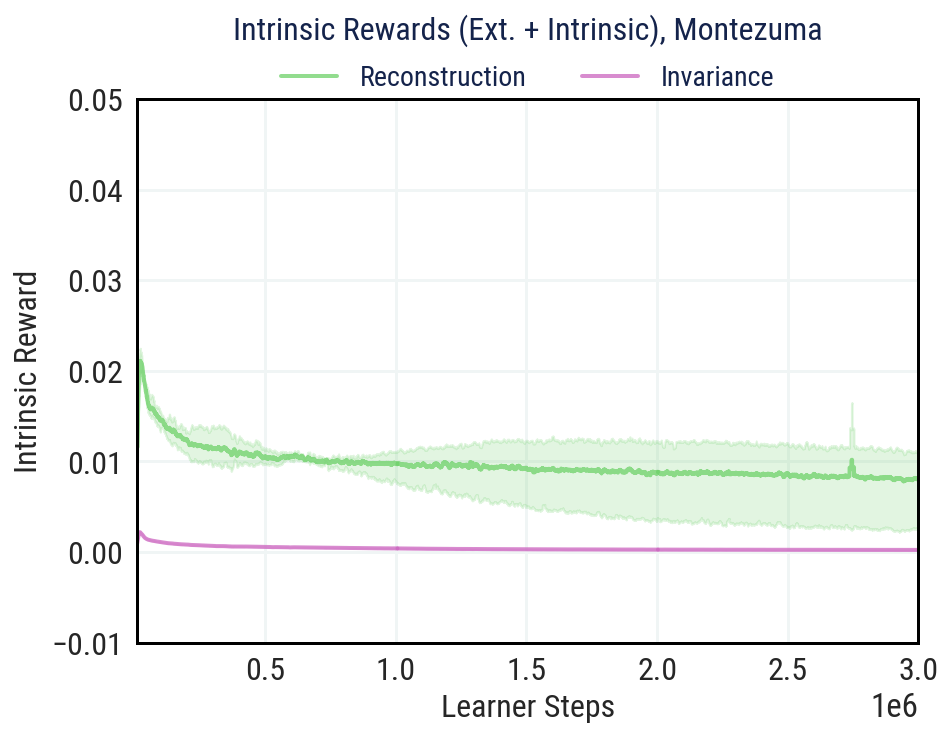}}
}
\vspace{-0.5em}
\caption{\small\dayum{\textit{Intrinsic Rewards (Montezuma's Revenge, with Sticky Actions and Non-Sticky Baseline)}.
Reconstruction and invariance bonuses.
}}
\label{fig:montezuma:intrinsic}
\vspace{-0.25em}
\end{figure*}

Regarding the former, Figure \ref{fig:montezuma:intrinsic} shows that---for both noise settings and both exploration regimes---the reconstruction bonus indeed converges to very small values, but not exactly zero, which is consistent with the fact that $Z$ does not perfectly capture the entirety of what is unpredictable. (This may be due to a variety of usual factors, such as non-realizability and errors in optimization and estimation of expectations).

Regarding the latter, there are several ways to assess how well the invariance constraint between $Z$ and $X,A$ may be enforced. Firstly, we may observe the value of the invariance loss. Figure \ref{fig:montezuma:intrinsic} also shows the invariance bonus over time: We observe that---to the best of the critic's ability to tell---the invariance constraint appears relatively well-enforced. Of course, this is certainly not a perfect measure, as it depends on how discriminative the critic is, to begin with.

\subsection{Outcome Losses}\label{subapp:outcome}

Secondly, we can also gauge the amount of informational overlap between $Z$ and $X,A$ as follows: In addition to learning the function $f_{\eta}(X,A,Z)$ (viz. ``Reconstruction Loss''), consider training additional predictors to predict $Y$, but only using states and actions as input as in the usual forward prediction (viz. ``Prediction Loss''), as well as only using the learned hindsight vectors as input (viz. ``Hindsight-only Loss'').

\begin{figure*}[h!]
\vspace{-0.75em}
\centering
\makebox[1.0\textwidth][c]{
\hspace{-9px}
\subfloat[\textbf{Sticky Actions (Intrinsic Only)}]{
\includegraphics[height=0.195\linewidth, trim=0em 0em 0em 0em]
{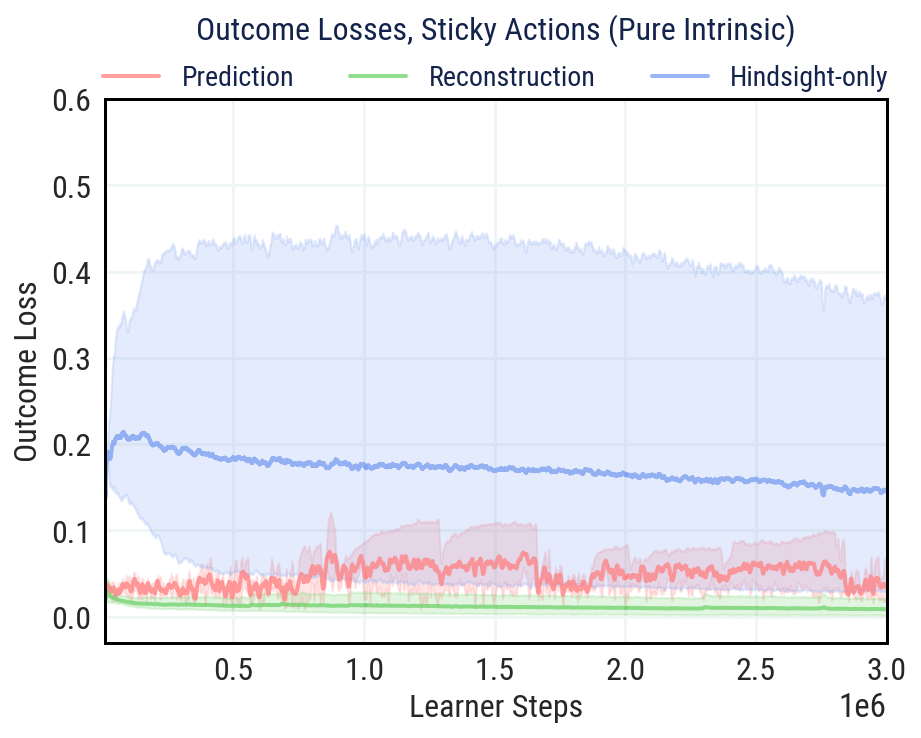}}
\hfill
\subfloat[\textbf{Sticky Actions (Ext. + Intrinsic)}]{
\includegraphics[height=0.195\linewidth, trim=0em 0em 0em 0em]
{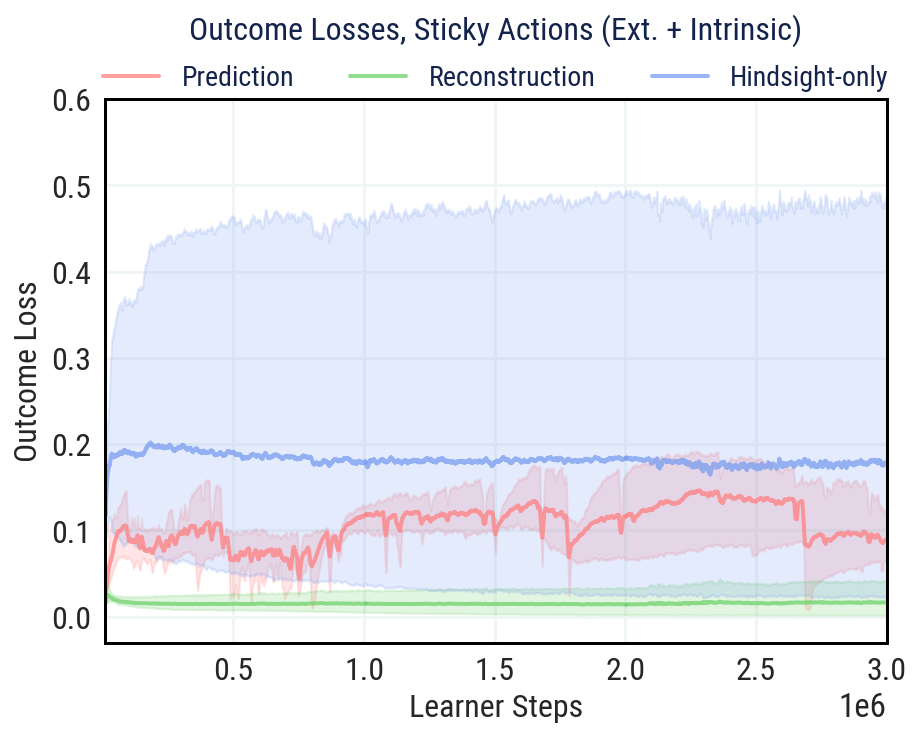}}
\hfill
\subfloat[\textbf{Non-Sticky Baseline (Intrinsic Only)}]{
\includegraphics[height=0.195\linewidth, trim=0em 0em 0em 0em]
{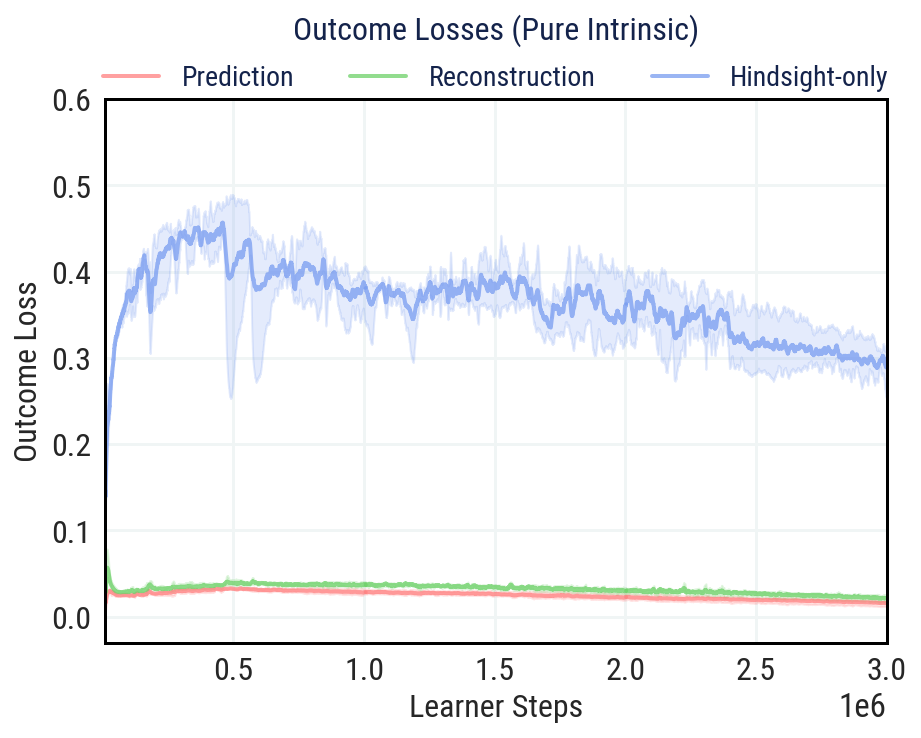}}
\hfill
\subfloat[\textbf{Non-Sticky Baseline (Ext. + Intrinsic)}]{
\includegraphics[height=0.195\linewidth, trim=0em 0em 0em 0em]
{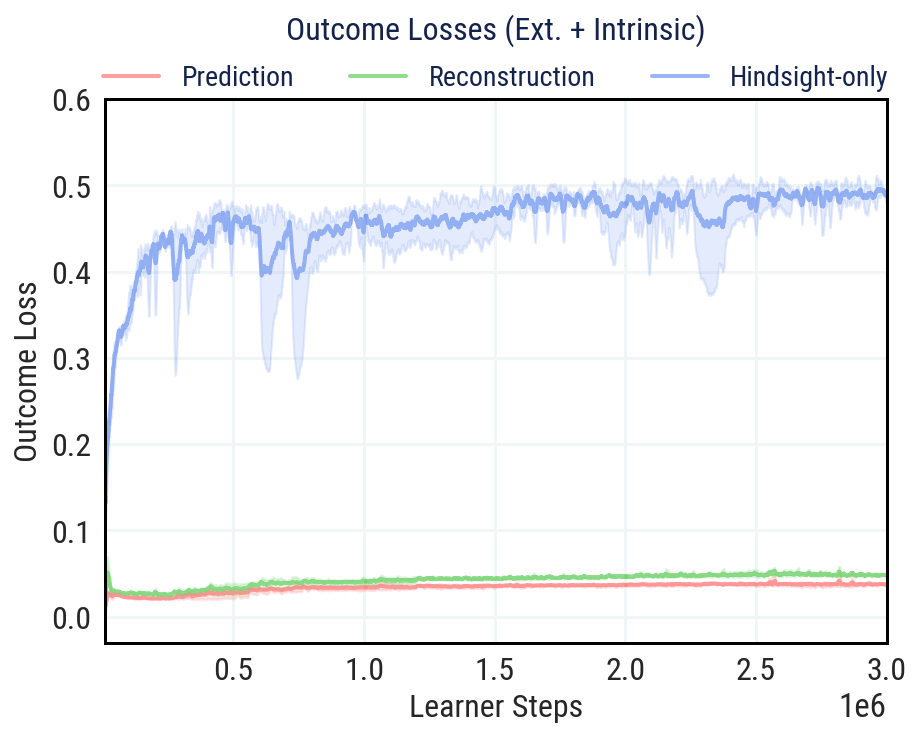}}
}
\vspace{-0.5em}
\caption{\small\dayum{\textit{Prediction, Reconstruction, and Hindsight-only Losses (Montezuma's Revenge, with Sticky Actions and Non-Sticky Baseline)}.
}}
\label{fig:montezuma:outcome}
\vspace{-0.25em}
\end{figure*}

From Figure \ref{fig:montezuma:outcome}, we observe in the ``Sticky Actions'' setting that the hindsight-only error (i.e. using only $Z$ as input) is the highest; prediction error (i.e. using only $X,A$) is lower, but not as low as reconstruction error (i.e. using $X,A,Z$ as input). In the ``Non-Sticky Baseline'' setting, as expected the prediction and reconstruction errors are roughly equal, since the environment is largely deterministic, so adding whatever hindsight vectors as input will not confer any benefit in modeling outcomes.
For reference, the variance of target vectors is around $~0.4$.
These observations are consistent with the fact that leakage indeed occurs, but much of it is regularized away by the invariance constraint. Precisely, from the loss comparison we see that $Z$ alone contains strictly less information than in $X,A$ for predicting $Y$, and that their union $X,A,Z$ contains the most information for predicting $Y$.
(Note that much information about each room is statically determined by the room number, thus even a tiny amount if leakage may be sufficient to determine large portions of the future).

Finally, we stress that care is required when interpreting these curves due to the bootstrapped nature of the latent space in \tttext{BYOL-Hindsight}, as well as the fact that the dataset on which the predictor/reconstructor, generator, and critic are trained is endogenous to the rollout policy trained on the basis of the intrinsic reward.

\subsection{Hindsight Information}\label{subapp:representations}

\begin{figure*}[h!]
\vspace{-0.75em}
\subfloat[$o_{t-3}$; $a_{t-4}=R$]{
{\transparent{0.9}
\includegraphics[height=0.167\linewidth, trim=0em 0em 0em 0em]
{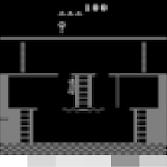}}}
\hfill
\subfloat[$o_{t-2}$; $a_{t-3}=R$]{
{\transparent{0.9}
\includegraphics[height=0.167\linewidth, trim=0em 0em 0em 0em]
{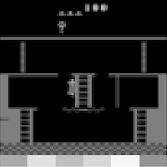}}}
\hfill
\subfloat[$o_{t-1}$; $a_{t-2}=R$]{
{\transparent{0.9}
\includegraphics[height=0.167\linewidth, trim=0em 0em 0em 0em]
{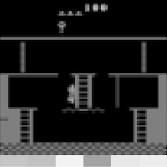}}}
\hfill
\subfloat[$o_{t}$; $a_{t-1}=L$]{
{\transparent{0.9}
\includegraphics[height=0.167\linewidth, trim=0em 0em 0em 0em]
{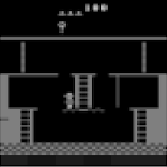}}}
\hfill
\subfloat[$o_{t+1}$; $a_{t}=U$]{
{\transparent{0.9}
\includegraphics[height=0.167\linewidth, trim=0em 0em 0em 0em]
{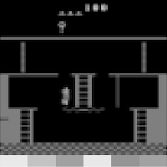}}}
\caption{\small\dayum{\textit{Example Input State ($x_{t}$\pix$\coloneqq$\pix$o_{t-3:t}$) and Target State ($x_{t+1}$\pix$\coloneqq$\pix$o_{t+1}$)}. For each frame, the prior (patch-sampling) action is shown.}}
\label{fig:viz:obs}
\vspace{-0.25em}
\end{figure*}

Thirdly, we can visually inspect for informational overlap in a simulation setting, as follows: In a rollout dataset $\rho_{\pi}$, define input states as the most recent four frames, $x_{t}\coloneqq o_{t-3:t}$, and define target states as the next frame $x_{t+1}\coloneqq o_{t+1}$. Then we learn representations $z_{t+1}$ as usual---that is, by optimizing the objective \smash{$
\text{min}_{\theta,\eta}
\text{max}_{\nu}
~
\mathbb{E}_{X_{t},A_{t}\sim\pi(\cdot|X_{t})}
\mathcal{R}_{\theta,\eta,\nu}^{K}(X_{t},A_{t})$}. 
As our source of stochasticity, we use a strip of large ``patches'' at the bottom of each frame, whose grayscale values are random variables that depend on the action selected by the agent in the prior time step. Importantly, each action may induce a different distribution of patch values for the strip that appears in the next frame---that is, the most natural parameterization involves a structural causal model with directed edges from $A_{t}$ to $Z_{t+1}$. See Figure \ref{fig:viz:obs} for a representative example of such a sequence of observations. (Note that this artificial-noise setting is similar to the noisy pixels we used above in Appendix \ref{subapp:persistive}, but these noise patches are larger and so more discernible than noise pixels for visual inspection).

\dayum{
Then we learn the following five functions: (a) ``Identity'', i.e. $\hat{x}_{t+1}$\pix$\coloneqq$\pix$h(x_{t+1})$; (b) ``Prediction'', i.e. $\hat{x}_{t+1}$\pix$\coloneqq$\pix$h(x_{t},a_{t})$; (c) ``Reconstruction'', i.e. $\hat{x}_{t+1}$\pix$\coloneqq$\pix$h(x_{t},a_{t},z_{t+1})$; (d) ``Hindsight-Only'', i.e. $\hat{x}_{t+1}$\pix$\coloneqq$\pix$h(z_{t+1})$; and (e) ``Hindsight-and-Action-Only'', i.e. $\hat{x}_{t+1}$\pix$\coloneqq$\pix$h(a_{t},z_{t+1})$. Given a new input state $x_{t}$ and action $a_{t}$ (and hindsight $z_{t+1}$ from the generator), we obtain the outputs $\hat{x}_{t+1}$ from these functions, and inspect the pixel-wise differences (i.e. target state errors) from ground-truths $x_{t+1}$.
}

\begin{figure*}[h!]
\vspace{-0.75em}
\subfloat[$\hat{x}_{t+1}$\pix$\coloneqq$\pix$h(x_{t+1})$]{
{\transparent{1.0}
\includegraphics[height=0.167\linewidth, trim=0em 0em 0em 0em]
{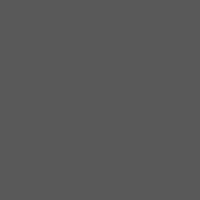}}}
\hfill
\subfloat[$\hat{x}_{t+1}$\pix$\coloneqq$\pix$h(x_{t},a_{t})$]{
\includegraphics[height=0.167\linewidth, trim=0em 0em 0em 0em]
{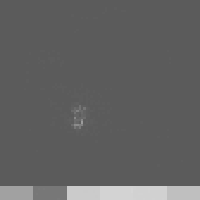}}
\hfill
\subfloat[$\hat{x}_{t+1}$\pix$\coloneqq$\pix$h(x_{t},a_{t},z_{t+1})$]{
\includegraphics[height=0.167\linewidth, trim=0em 0em 0em 0em]
{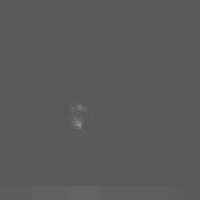}}
\hfill
\subfloat[$\hat{x}_{t+1}$\pix$\coloneqq$\pix$h(z_{t+1})$]{
\includegraphics[height=0.167\linewidth, trim=0em 0em 0em 0em]
{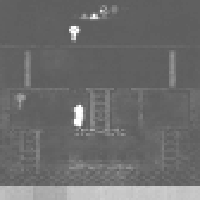}}
\hfill
\subfloat[$\hat{x}_{t+1}$\pix$\coloneqq$\pix$h(a_{t},z_{t+1})$]{
\includegraphics[height=0.167\linewidth, trim=0em 0em 0em 0em]
{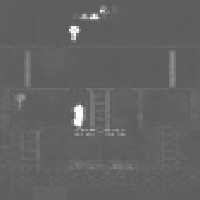}}
\caption{\small\textit{Visualizing Target State Errors}. Identity, Prediction, Reconstruction, Hindsight-Only, and Hindsight-and-Action-Only Errors.}
\label{fig:viz:con}
\vspace{-0.25em}
\end{figure*}

\dayum{
See Figure \ref{fig:viz:con} for a representative example (corresponding to the example in Figure \ref{fig:viz:obs}). Note that the first three results are straightforward and as expected: (a) The ``Identity'' function appears to be learned very well. (b) The ``Prediction'' function also appears to learn to predict the main content of the next frame quite well, but is completely unable to predict the random pixel strip at the bottom of the frame. (c) The ``Reconstruction'' function appears to learn to reconstruct both the main content of the next frame, as well as the random pixel strip. Importantly, we may now ask: What if $z_{t+1}$ has simply learned to copy $x_{t+1}$, or otherwise leaked information from $x_{t},a_{t}$? The next two results reassure us that this is unlikely: (d) The ``Hindsight-Only'' function appears to map quite poorly into both the main content and the pixel strip (e.g. the main character is never even there), which tells us that $z_{t+1}$ alone does not have great overlap with $x_{t},a_{t}$. Moreover, (e) The ``Hindsight-and-Action-Only'' function appears to map quite poorly into the main content, but now the pixel strip is actually modeled very well, which is consistent with the fact that $z_{t+1}$ does capture latent information about stochasticity that relies on $a_{t}$ for resolution.
}

\subsection{Hard Exploration Games}\label{subapp:hard}

\begin{figure*}[h!]
\vspace{-1.5em}
\centering
\makebox[1.0\textwidth][c]{
\hspace{-12px}
\subfloat[\tiny\textbf{Alien (I)}]{
\includegraphics[height=0.19\linewidth, trim=0em 0em 0em 0em]
{results/I_Manuscript/b_Return/3_A_Return/alien_int}}
\hfill
\subfloat[\tiny\textbf{Bank (I)}]{
\includegraphics[height=0.19\linewidth, trim=0em 0em 0em 0em]
{results/I_Manuscript/b_Return/3_A_Return/bank_int}}
\hfill
\subfloat[\tiny\textbf{Freeway (I)}]{
\includegraphics[height=0.19\linewidth, trim=0em 0em 0em 0em]
{results/I_Manuscript/b_Return/3_A_Return/freeway_int}}
\hfill
\subfloat[\tiny\textbf{Gravitar (I)}]{
\includegraphics[height=0.19\linewidth, trim=0em 0em 0em 0em]
{results/I_Manuscript/b_Return/3_A_Return/gravitar_int}}
}
\vspace{-0.6em}

\makebox[\textwidth][c]{
\hspace{-12px}
\subfloat[\tiny\textbf{Hero (I)}]{
\includegraphics[height=0.19\linewidth, trim=0em 0em 0em 0em]
{results/I_Manuscript/b_Return/3_A_Return/hero_int}}
\hfill
\subfloat[\tiny\textbf{Montezuma (I)}]{
\includegraphics[height=0.19\linewidth, trim=0em 0em 0em 0em]
{results/I_Manuscript/b_Return/3_A_Return/montezuma_int}}
\hfill
\subfloat[\tiny\textbf{Pitfall (I)}]{
\includegraphics[height=0.19\linewidth, trim=0em 0em 0em 0em]
{results/I_Manuscript/b_Return/3_A_Return/pitfall_int}}
\hfill
\subfloat[\tiny\textbf{Private Eye (I)}]{
\includegraphics[height=0.19\linewidth, trim=0em 0em 0em 0em]
{results/I_Manuscript/b_Return/3_A_Return/private_int}}
}
\vspace{-0.6em}

\makebox[\textwidth][c]{
\hspace{-12px}
\subfloat[\tiny\textbf{Solaris (I)}]{
\includegraphics[height=0.19\linewidth, trim=0em 0em 0em 0em]
{results/I_Manuscript/b_Return/3_A_Return/solaris_int}}
\hfill
\subfloat[\tiny\textbf{Venture (I)}]{
\includegraphics[height=0.19\linewidth, trim=0em 0em 0em 0em]
{results/I_Manuscript/b_Return/3_A_Return/venture_int}}
\hfill
\subfloat[\tiny\textbf{Alien (E+I)}]{
\includegraphics[height=0.19\linewidth, trim=0em 0em 0em 0em]
{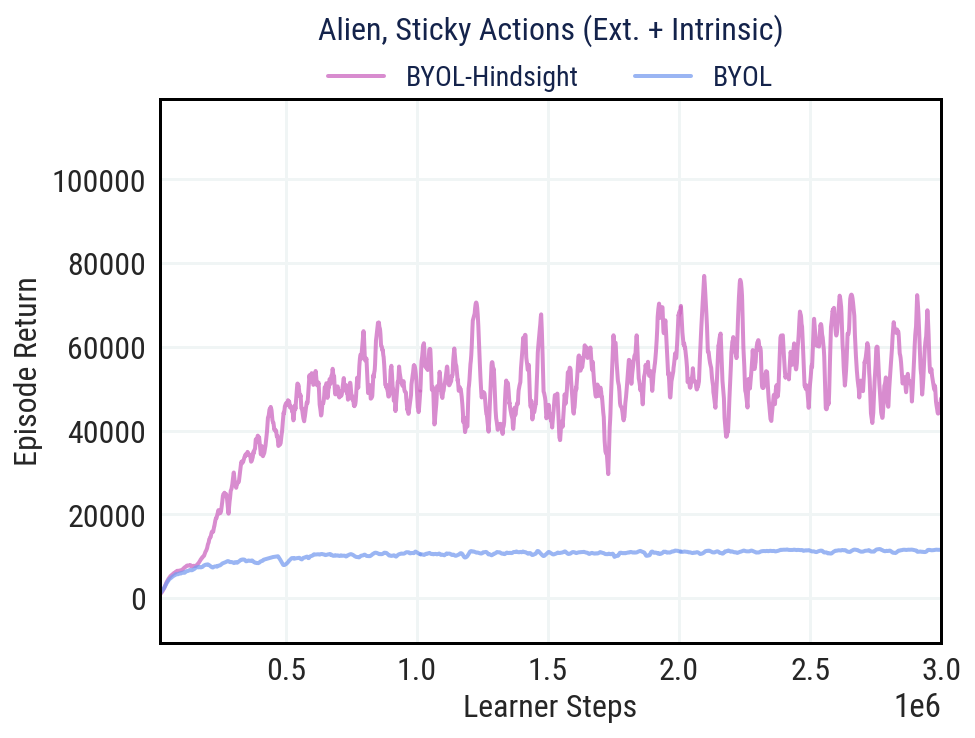}}
\hfill
\subfloat[\tiny\textbf{Bank (E+I)}]{
\includegraphics[height=0.19\linewidth, trim=0em 0em 0em 0em]
{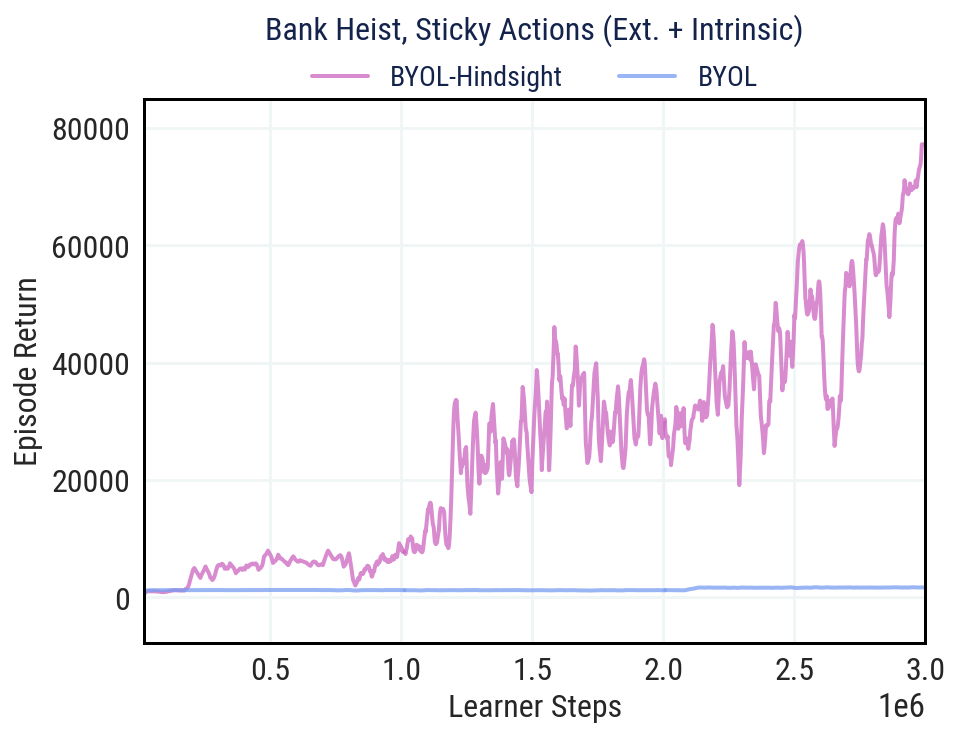}}
}
\vspace{-0.6em}

\makebox[\textwidth][c]{
\hspace{-12px}
\subfloat[\tiny\textbf{Freeway (E+I)}]{
\includegraphics[height=0.19\linewidth, trim=0em 0em 0em 0em]
{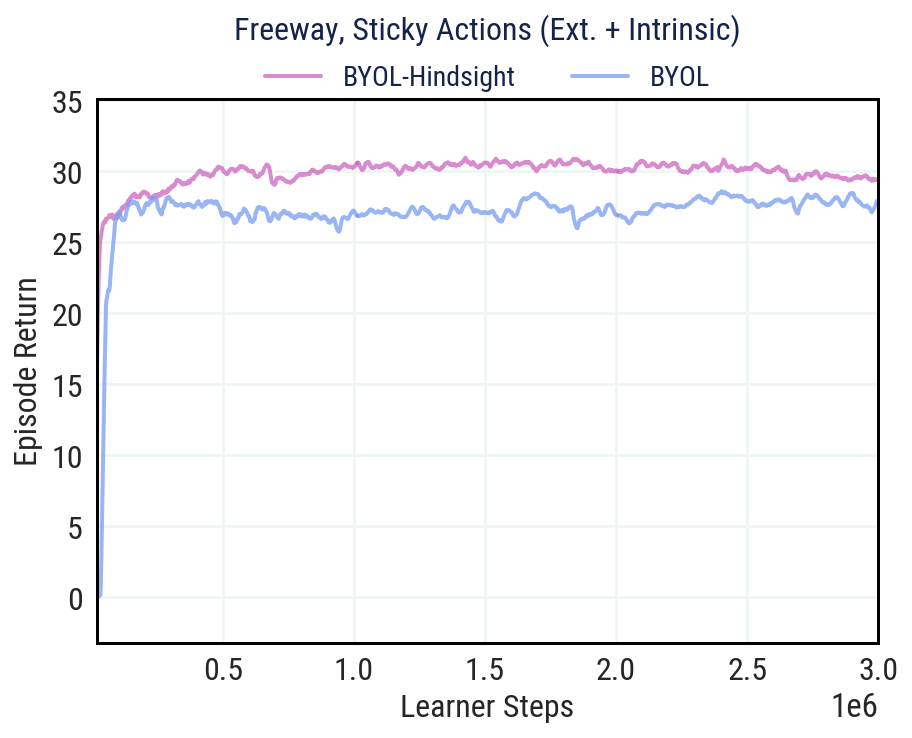}}
\hfill
\subfloat[\tiny\textbf{Gravitar (E+I)}]{
\includegraphics[height=0.19\linewidth, trim=0em 0em 0em 0em]
{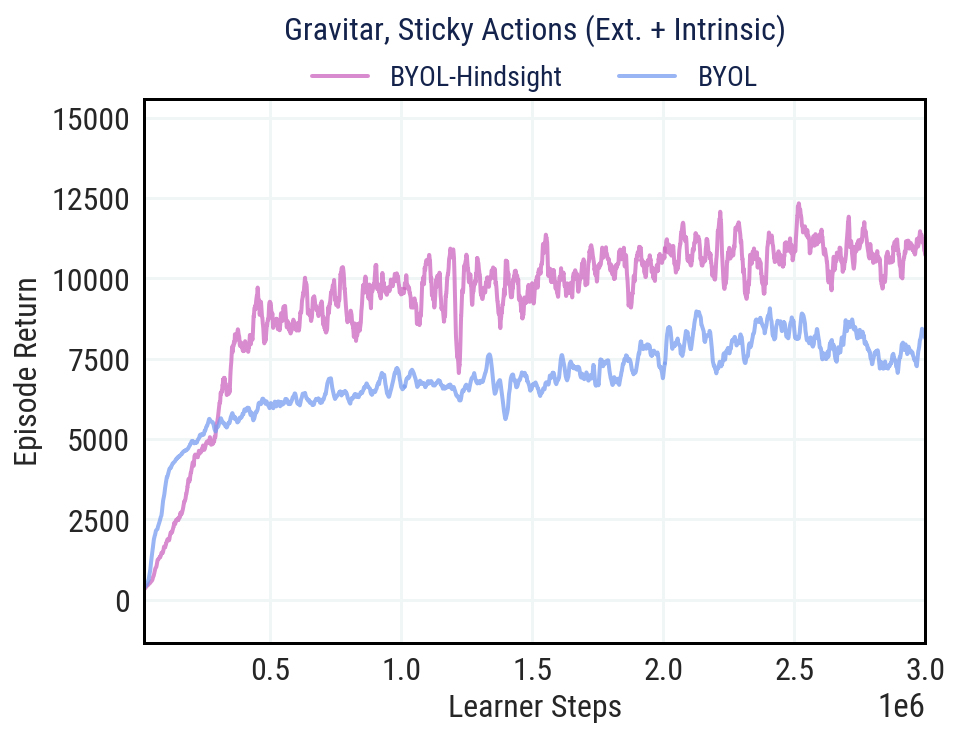}}
\hfill
\subfloat[\tiny\textbf{Hero (E+I)}]{
\includegraphics[height=0.19\linewidth, trim=0em 0em 0em 0em]
{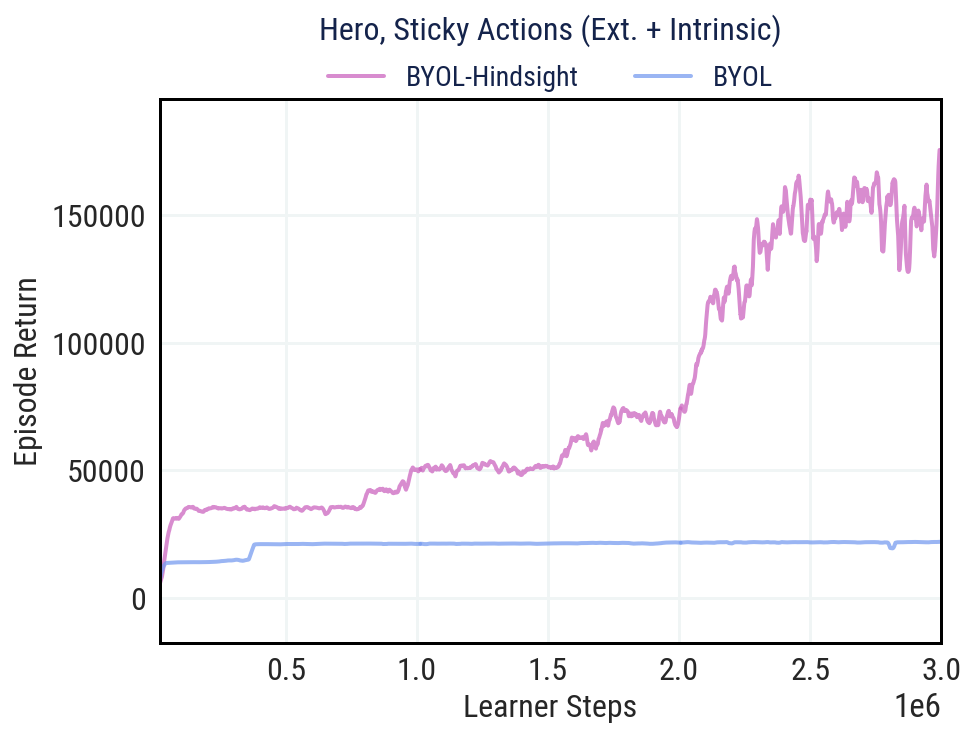}}
\hfill
\subfloat[\tiny\textbf{Montezuma (E+I)}]{
\includegraphics[height=0.19\linewidth, trim=0em 0em 0em 0em]
{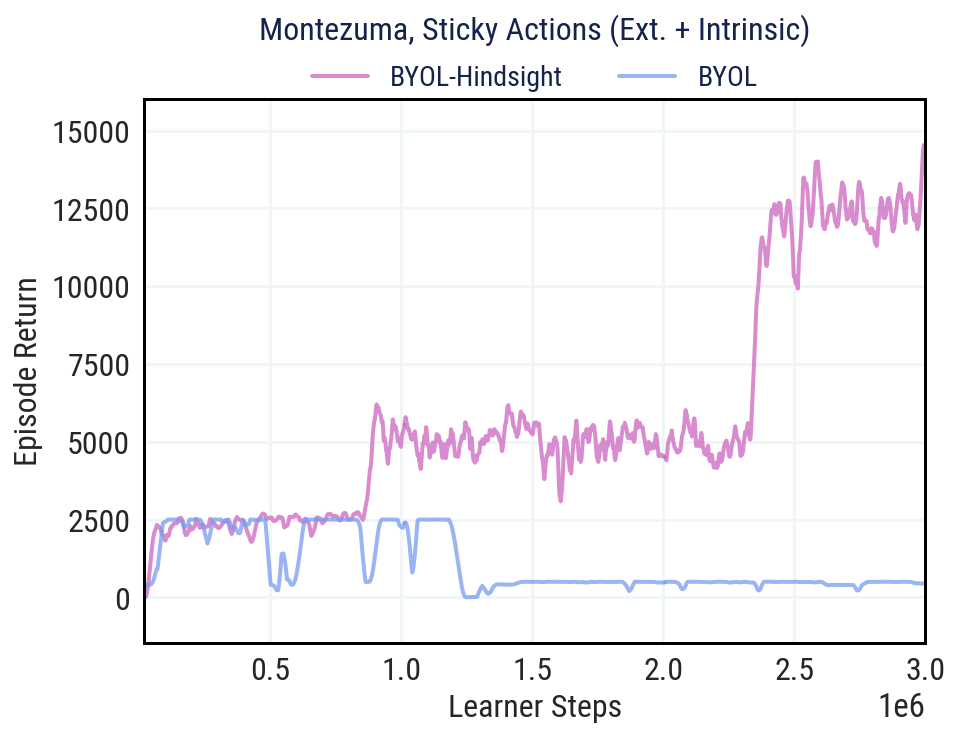}}
}
\vspace{-0.6em}

\makebox[\textwidth][c]{
\hspace{-12px}
\subfloat[\tiny\textbf{Pitfall (E+I)}]{
\includegraphics[height=0.19\linewidth, trim=0em 0em 0em 0em]
{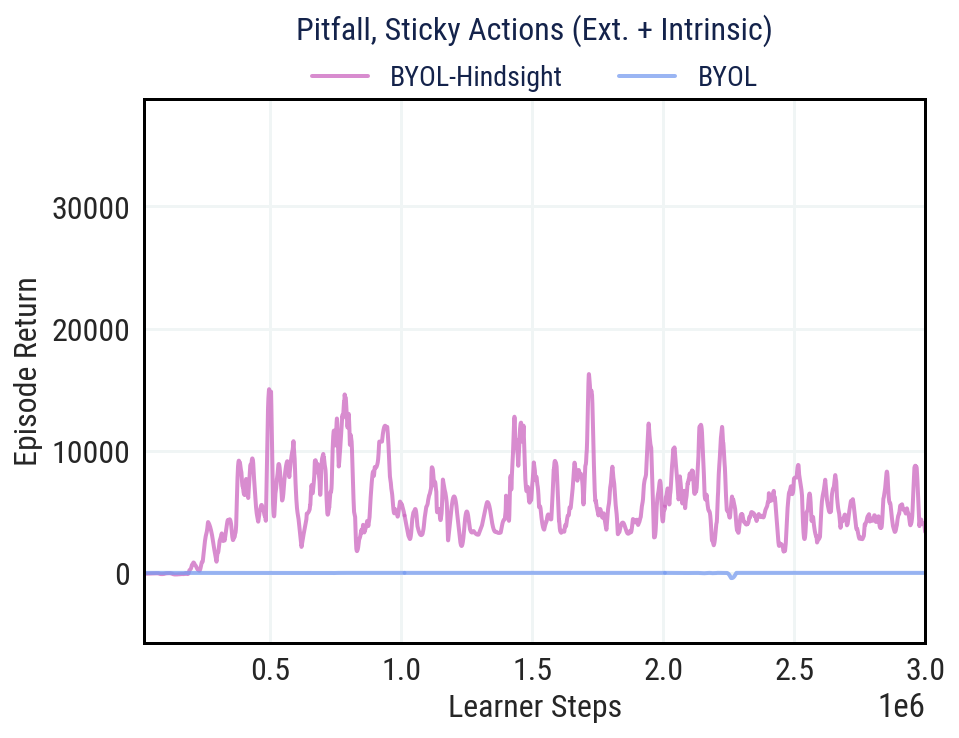}}
\hfill
\subfloat[\tiny\textbf{Private Eye (E+I)}]{
\includegraphics[height=0.19\linewidth, trim=0em 0em 0em 0em]
{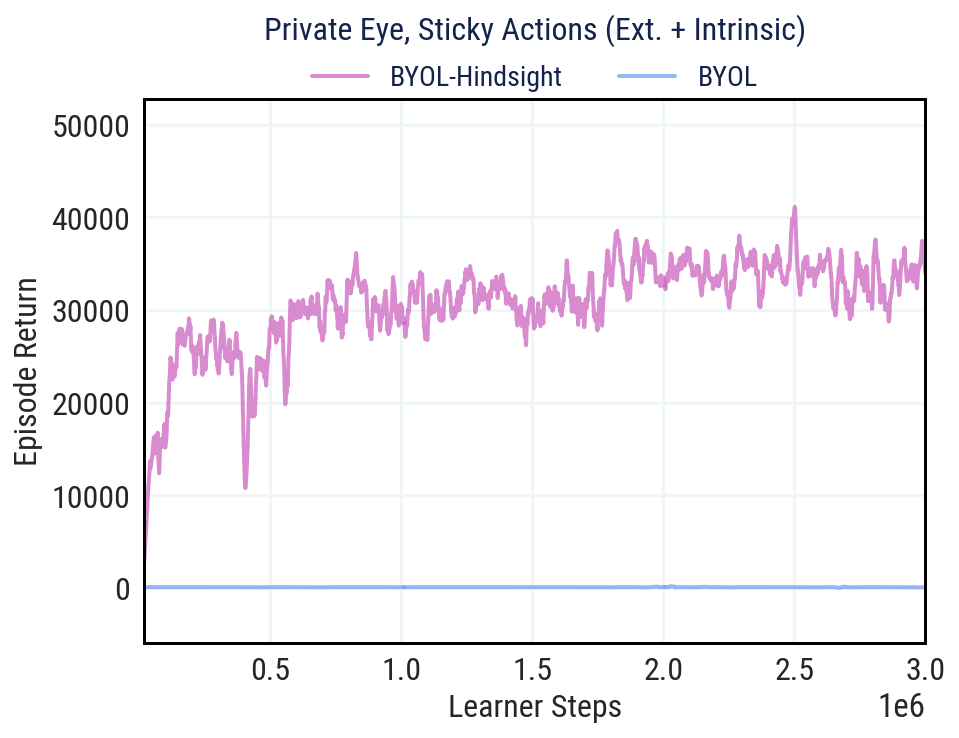}}
\hfill
\subfloat[\tiny\textbf{Solaris (E+I)}]{
\includegraphics[height=0.19\linewidth, trim=0em 0em 0em 0em]
{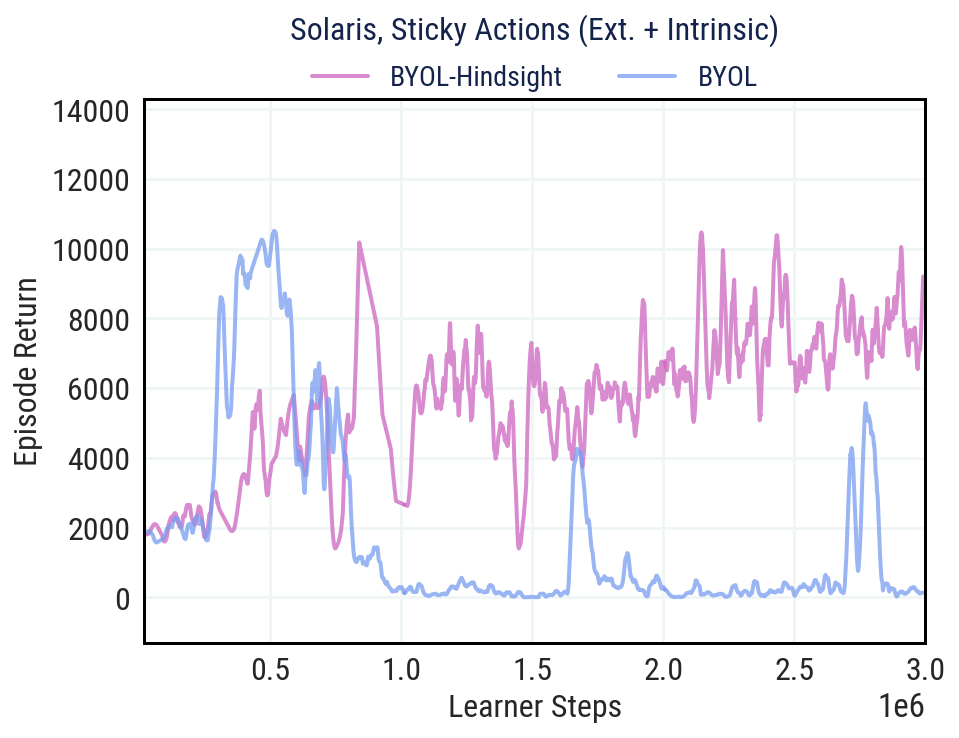}}
\hfill
\subfloat[\tiny\textbf{Venture (E+I)}]{
\includegraphics[height=0.19\linewidth, trim=0em 0em 0em 0em]
{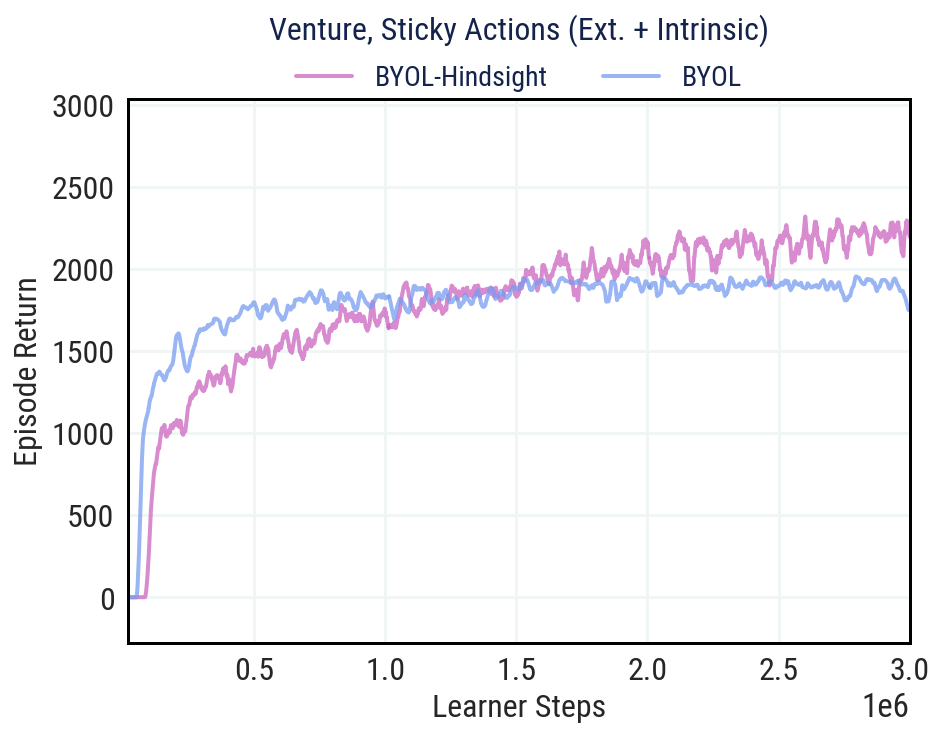}}
}
\vspace{-0.6em}

\caption{\small\dayum{\textit{Hard Exploration Games, with Sticky Actions}.
Performance measured by the sum of extrinsic rewards obtained in an episode.
}}
\label{fig:atari:return:big}
\vspace{-0.5em}
\end{figure*}

\dayum{
Figure \ref{fig:atari:return} only showed results for Atari hard-exploration games in the intrinsic-only ``(I)'' exploration regime, due to space constraints. Figure \ref{fig:atari:return:big} shows larger plots for those, as well as corresponding results for the mixed ``(E+I)'' exploration regime.
}

\begin{figure*}[h!]
\vspace{-0.75em}
\centering
\makebox[1.0\textwidth][c]{
\hspace{-12px}
\subfloat[\tiny\textbf{Alien (I)}]{
\includegraphics[height=0.19\linewidth, trim=0em 0em 0em 0em]
{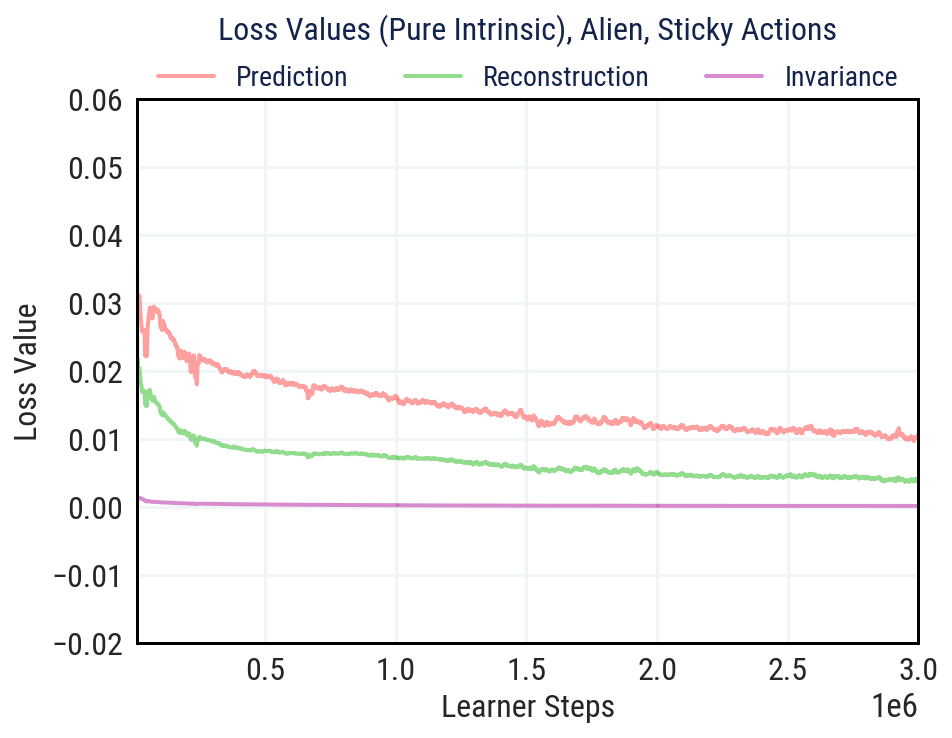}}
\hfill
\subfloat[\tiny\textbf{Bank (I)}]{
\includegraphics[height=0.19\linewidth, trim=0em 0em 0em 0em]
{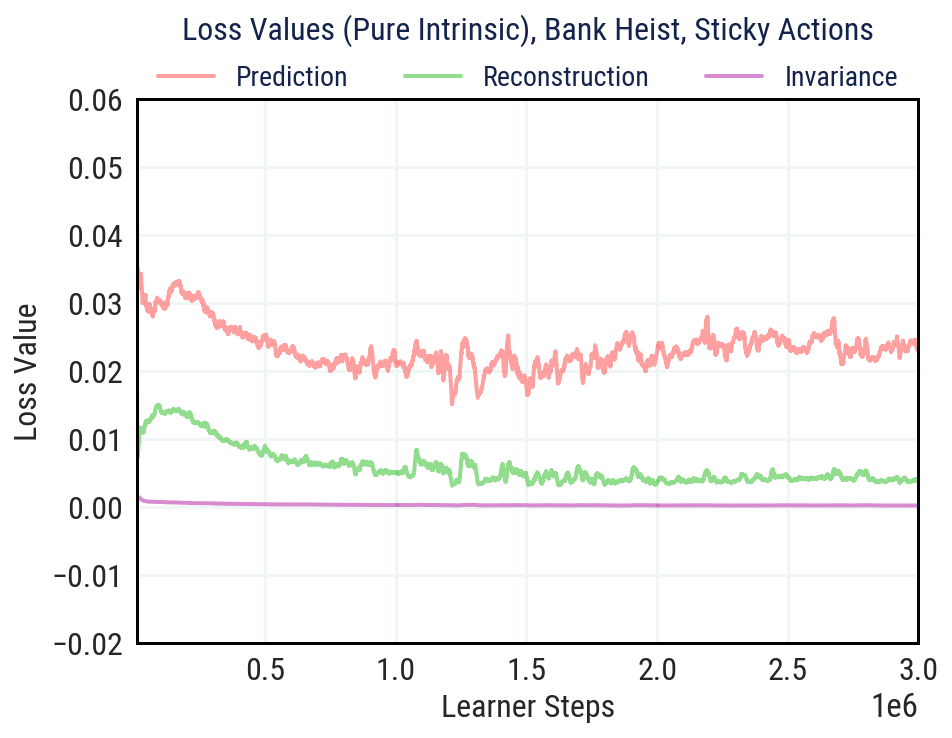}}
\hfill
\subfloat[\tiny\textbf{Freeway (I)}]{
\includegraphics[height=0.19\linewidth, trim=0em 0em 0em 0em]
{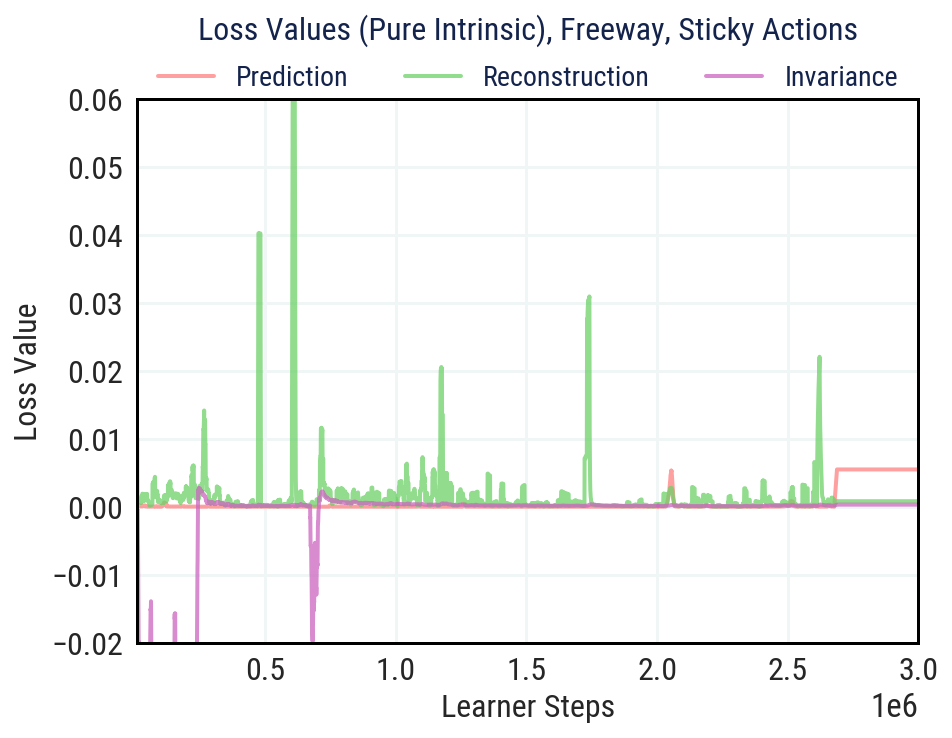}}
\hfill
\subfloat[\tiny\textbf{Gravitar (I)}]{
\includegraphics[height=0.19\linewidth, trim=0em 0em 0em 0em]
{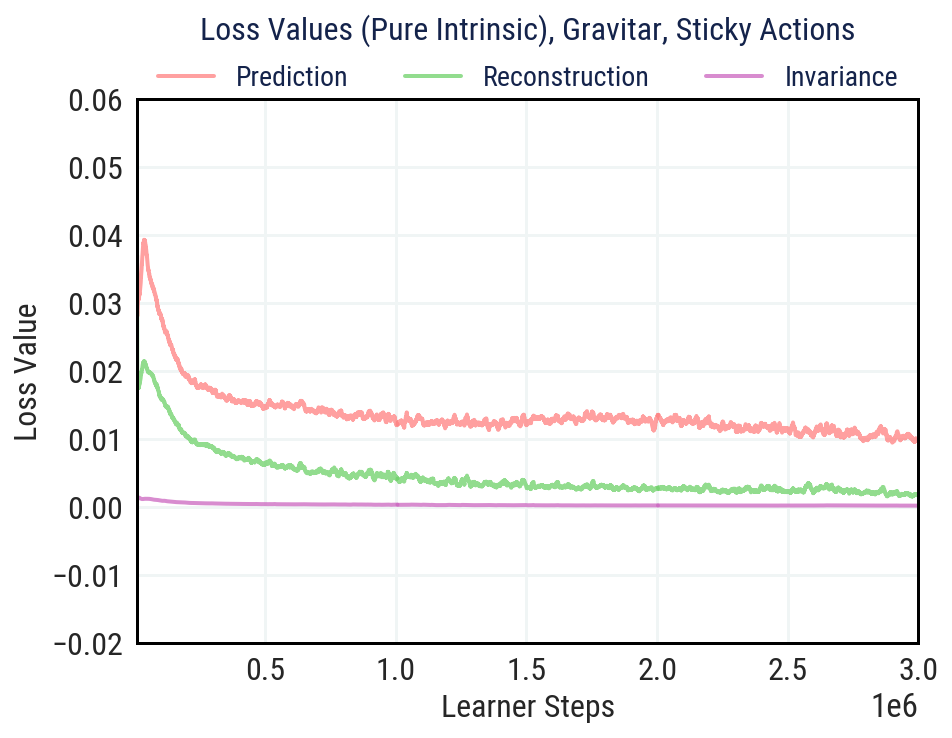}}
}
\vspace{-0.25em}

\makebox[\textwidth][c]{
\hspace{-12px}
\subfloat[\tiny\textbf{Hero (I)}]{
\includegraphics[height=0.19\linewidth, trim=0em 0em 0em 0em]
{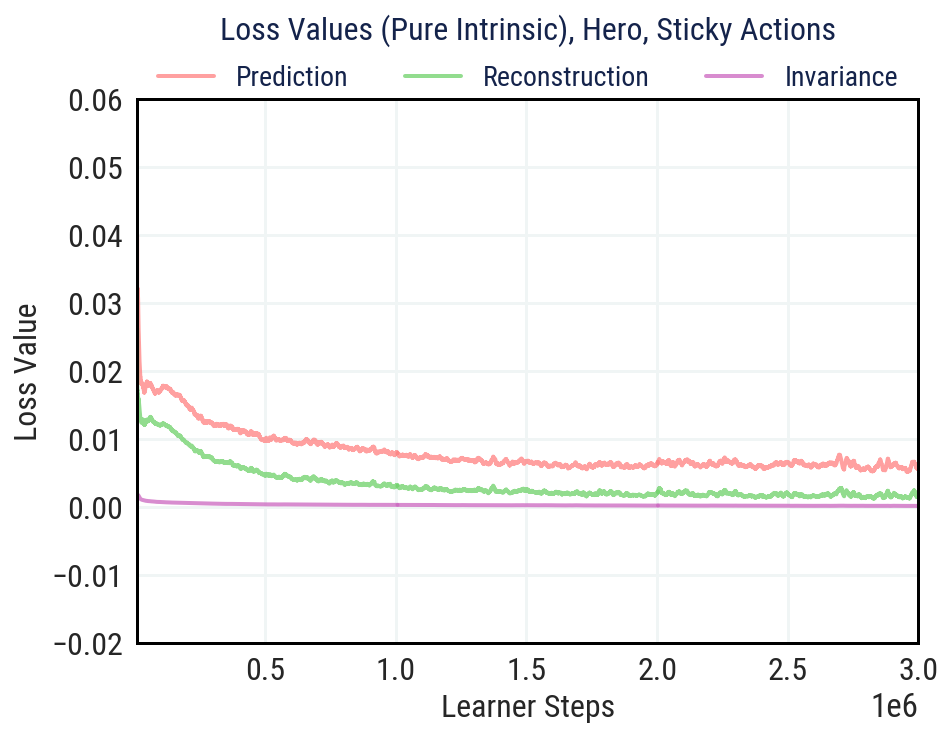}}
\hfill
\subfloat[\tiny\textbf{Pitfall (I)}]{
\includegraphics[height=0.19\linewidth, trim=0em 0em 0em 0em]
{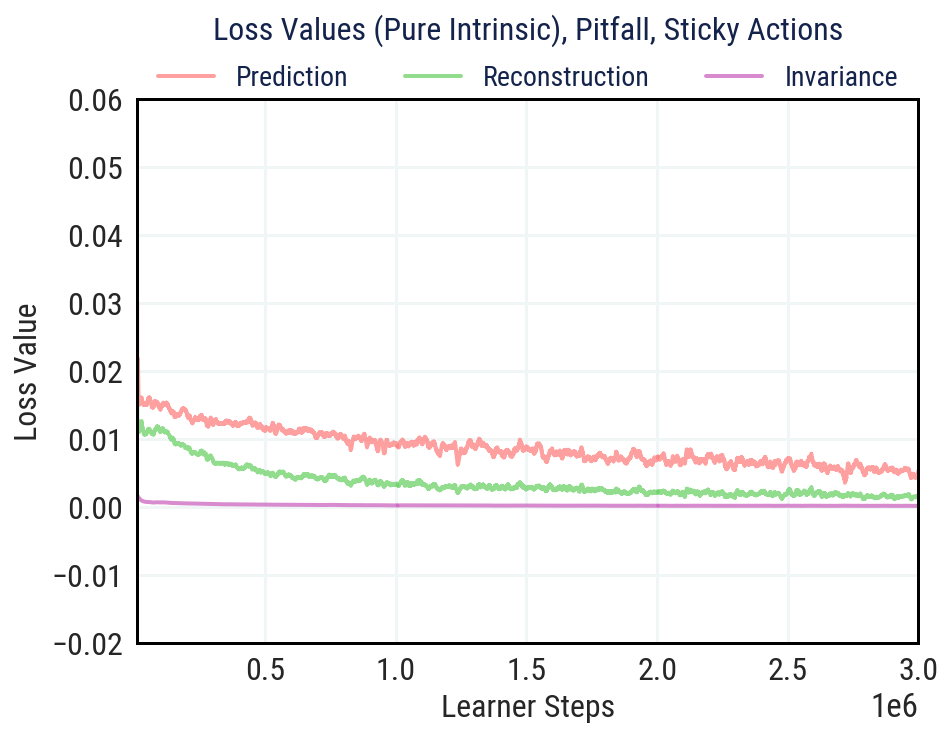}}
\hfill
\subfloat[\tiny\textbf{Private Eye (I)}]{
\includegraphics[height=0.19\linewidth, trim=0em 0em 0em 0em]
{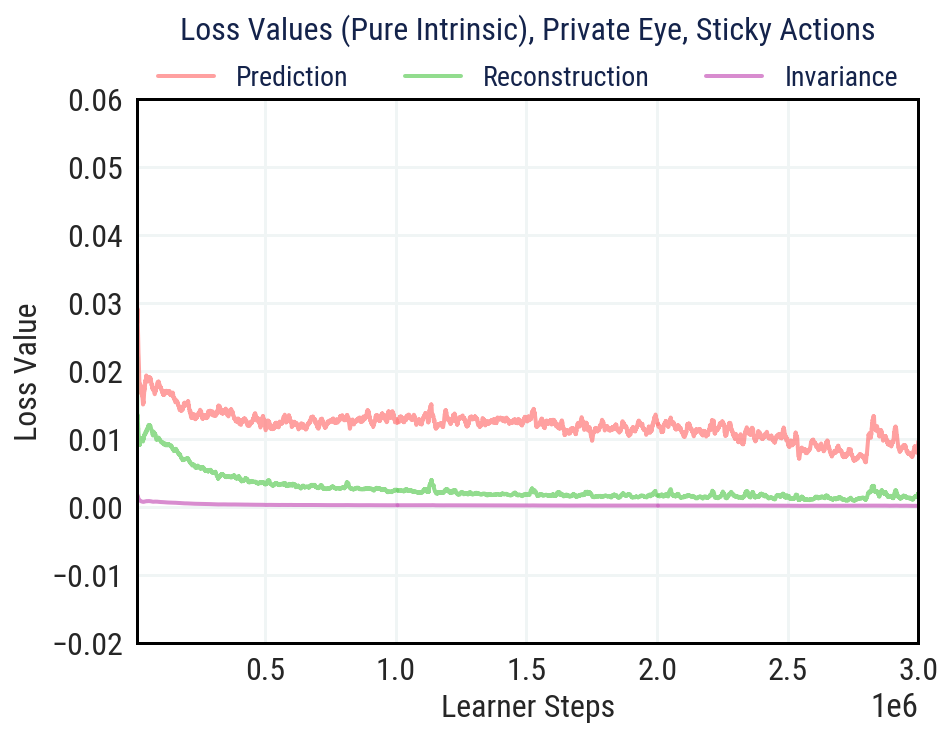}}
\hfill
\subfloat[\tiny\textbf{Qbert (I)}]{
\includegraphics[height=0.19\linewidth, trim=0em 0em 0em 0em]
{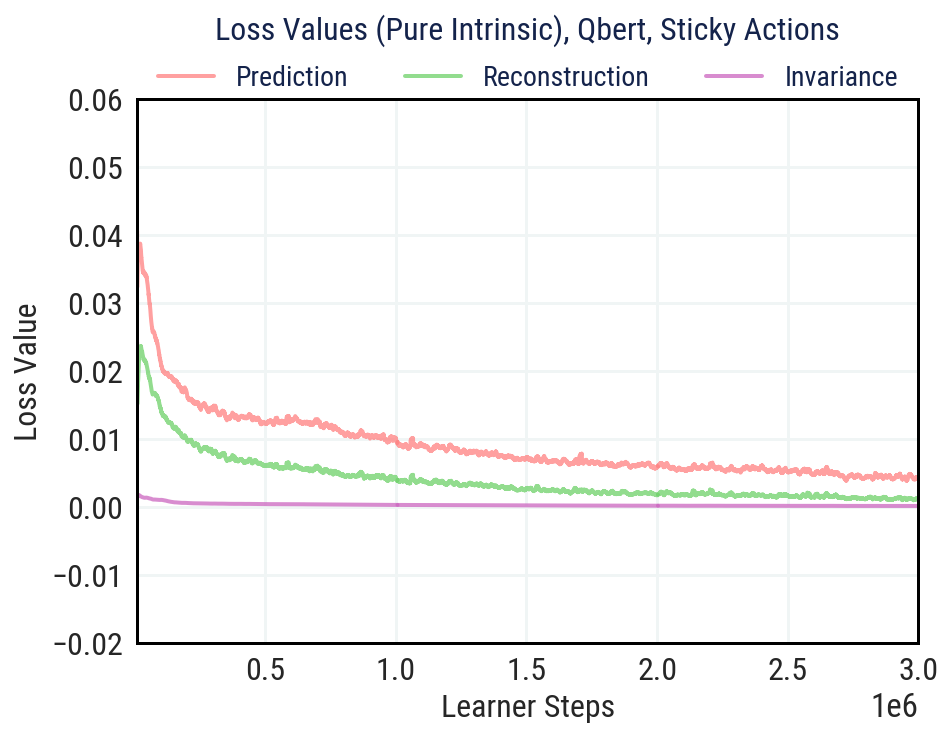}}

}
\vspace{-0.25em}

\makebox[\textwidth][c]{
\hspace{-12px}
\subfloat[\tiny\textbf{Solaris (I)}]{
\includegraphics[height=0.19\linewidth, trim=0em 0em 0em 0em]
{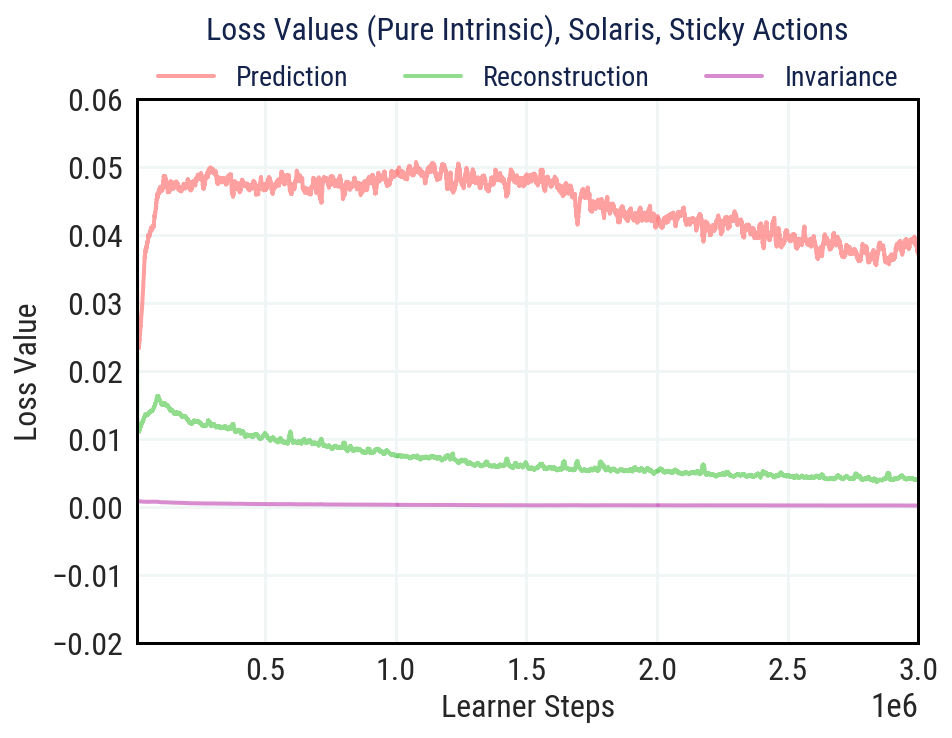}}
\hfill
\subfloat[\tiny\textbf{Venture (I)}]{
\includegraphics[height=0.19\linewidth, trim=0em 0em 0em 0em]
{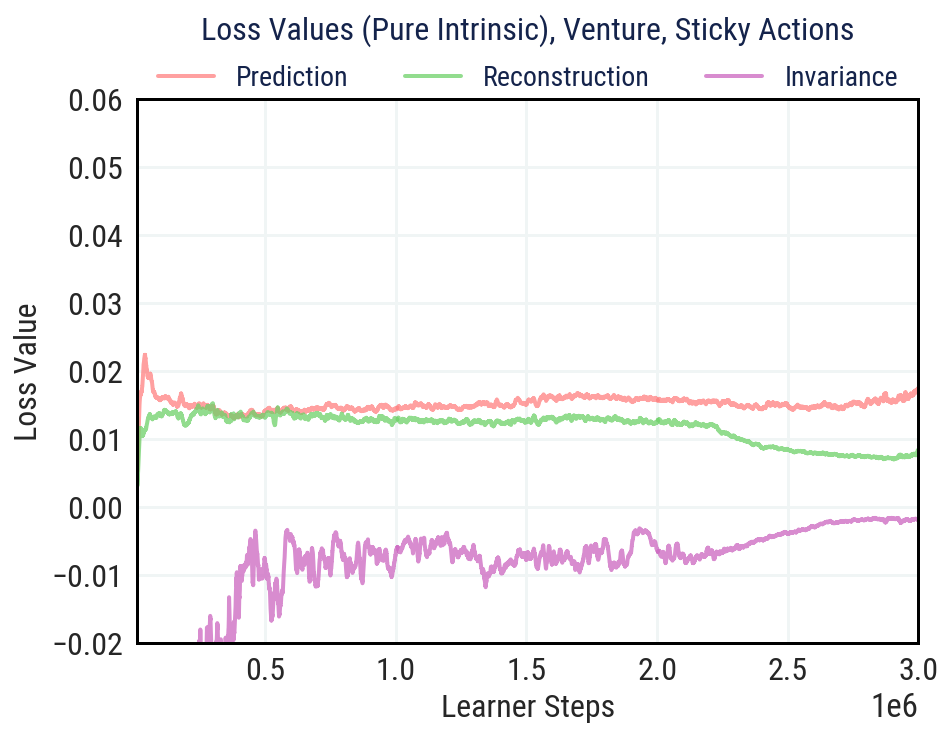}}
\hfill
\subfloat[\tiny\textbf{Alien (E+I)}]{
\includegraphics[height=0.19\linewidth, trim=0em 0em 0em 0em]
{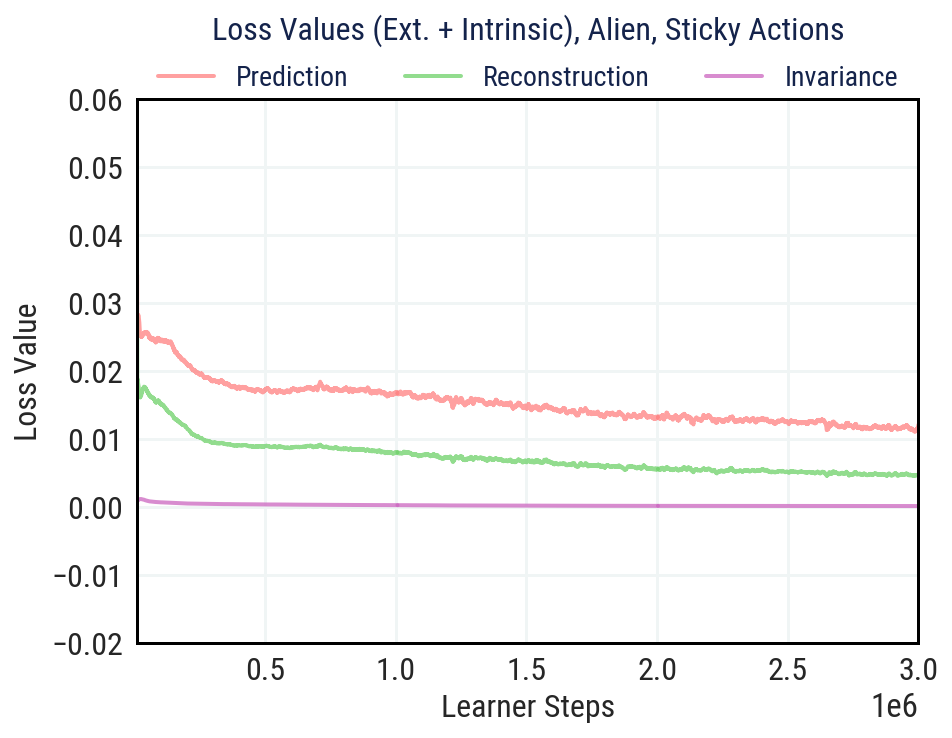}}
\hfill
\subfloat[\tiny\textbf{Bank (E+I)}]{
\includegraphics[height=0.19\linewidth, trim=0em 0em 0em 0em]
{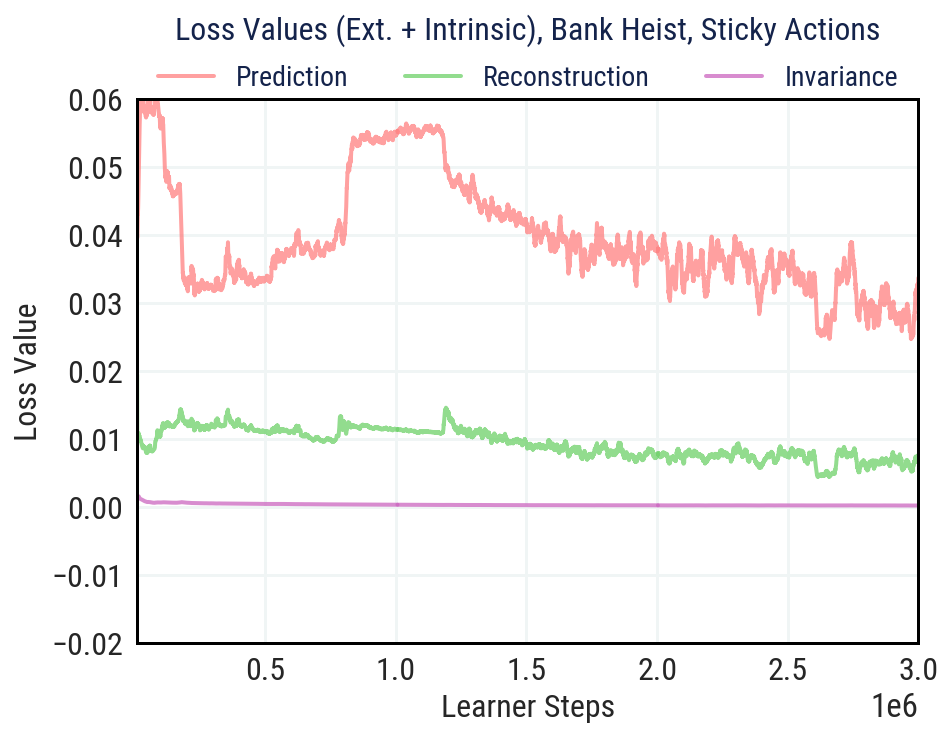}}
}
\vspace{-0.25em}

\makebox[\textwidth][c]{
\hspace{-12px}
\subfloat[\tiny\textbf{Freeway (E+I)}]{
\includegraphics[height=0.19\linewidth, trim=0em 0em 0em 0em]
{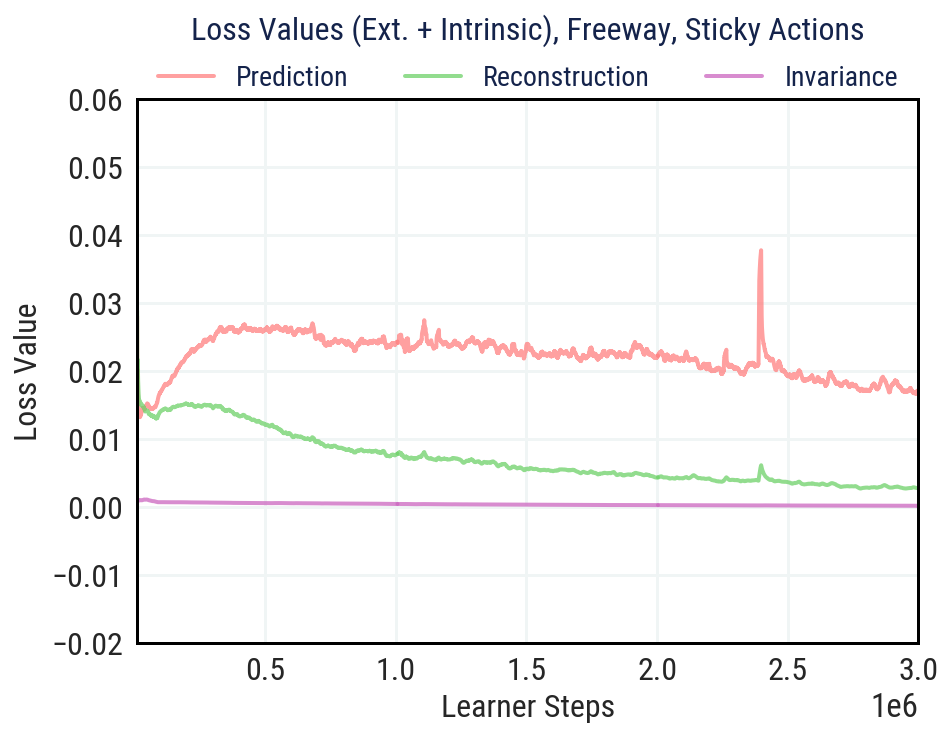}}
\hfill
\subfloat[\tiny\textbf{Gravitar (E+I)}]{
\includegraphics[height=0.19\linewidth, trim=0em 0em 0em 0em]
{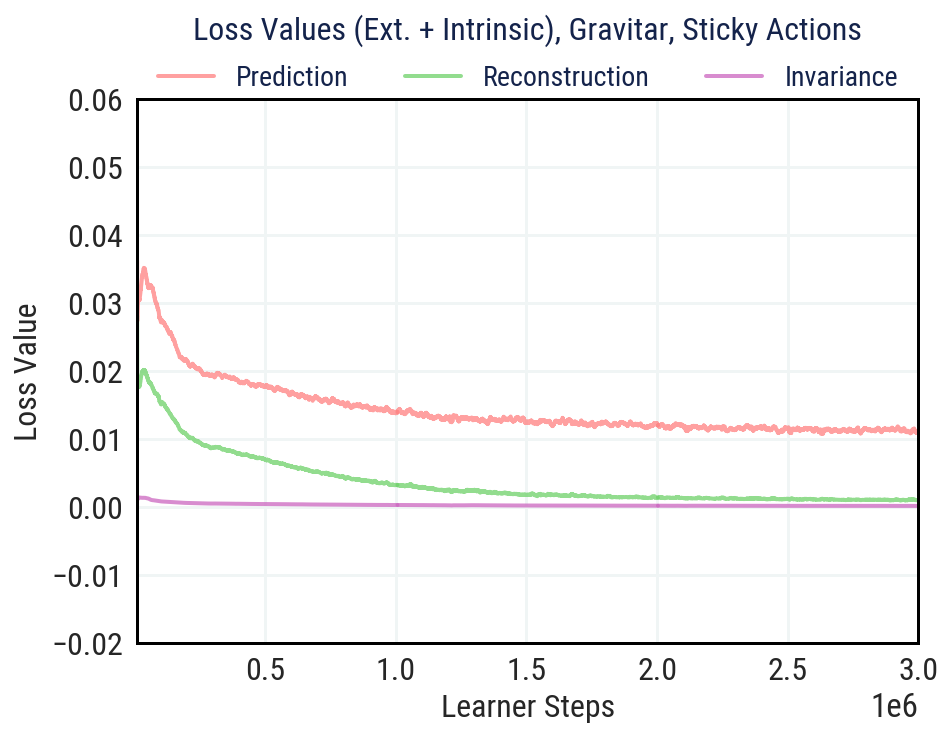}}
\hfill
\subfloat[\tiny\textbf{Hero (E+I)}]{
\includegraphics[height=0.19\linewidth, trim=0em 0em 0em 0em]
{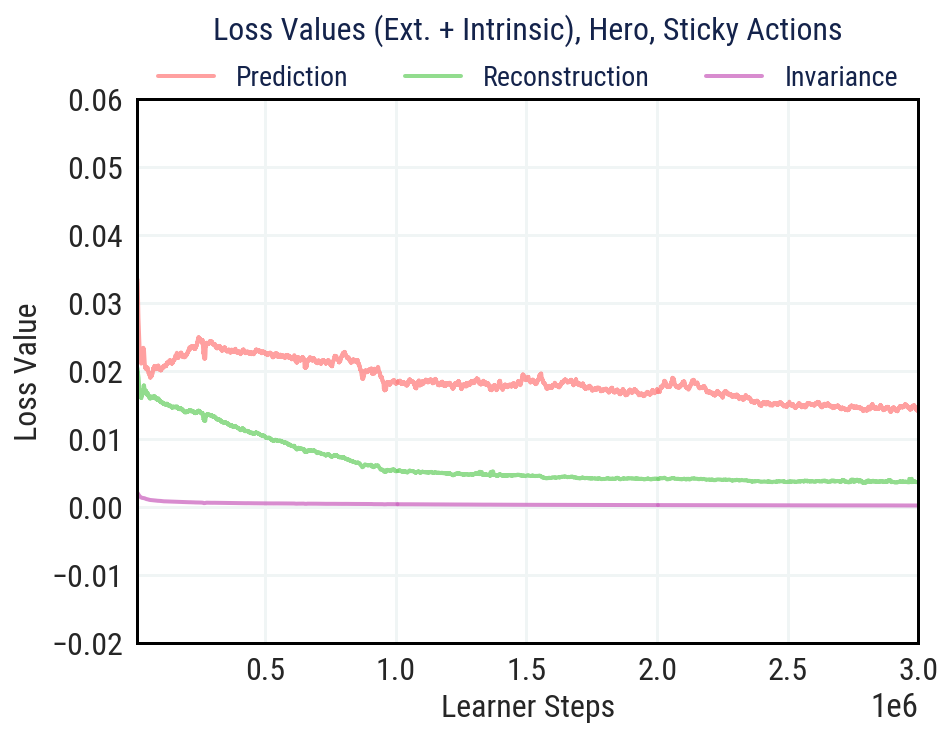}}
\hfill
\subfloat[\tiny\textbf{Pitfall (E+I)}]{
\includegraphics[height=0.19\linewidth, trim=0em 0em 0em 0em]
{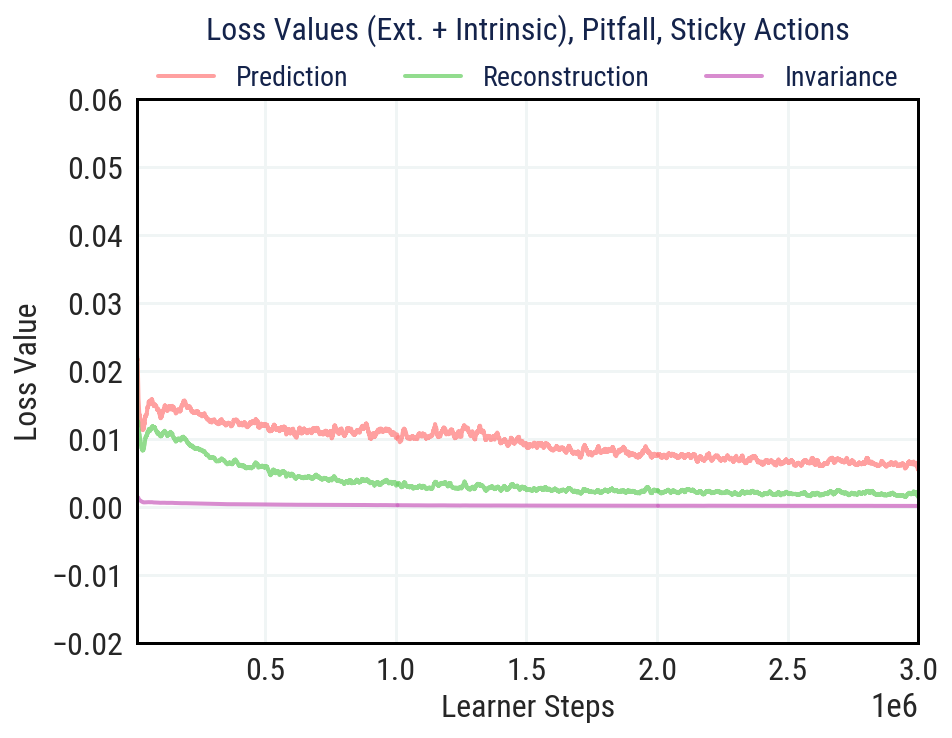}}
}
\vspace{-0.25em}

\makebox[\textwidth][c]{
\hspace{-12px}
\subfloat[\tiny\textbf{Private Eye (E+I)}]{
\includegraphics[height=0.19\linewidth, trim=0em 0em 0em 0em]
{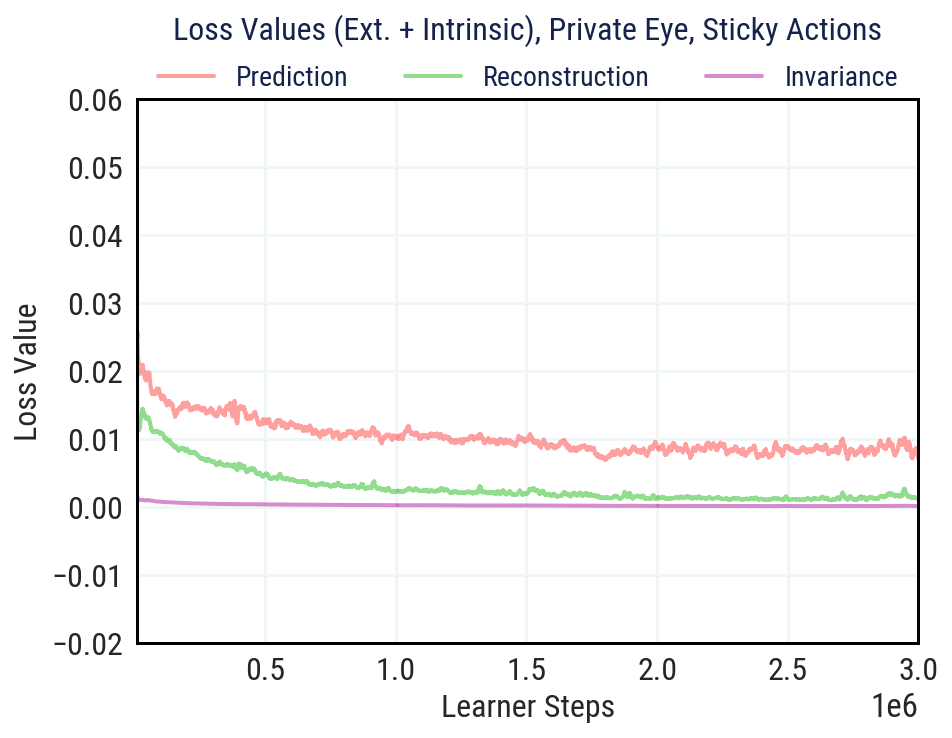}}
\hfill
\subfloat[\tiny\textbf{Qbert (E+I)}]{
\includegraphics[height=0.19\linewidth, trim=0em 0em 0em 0em]
{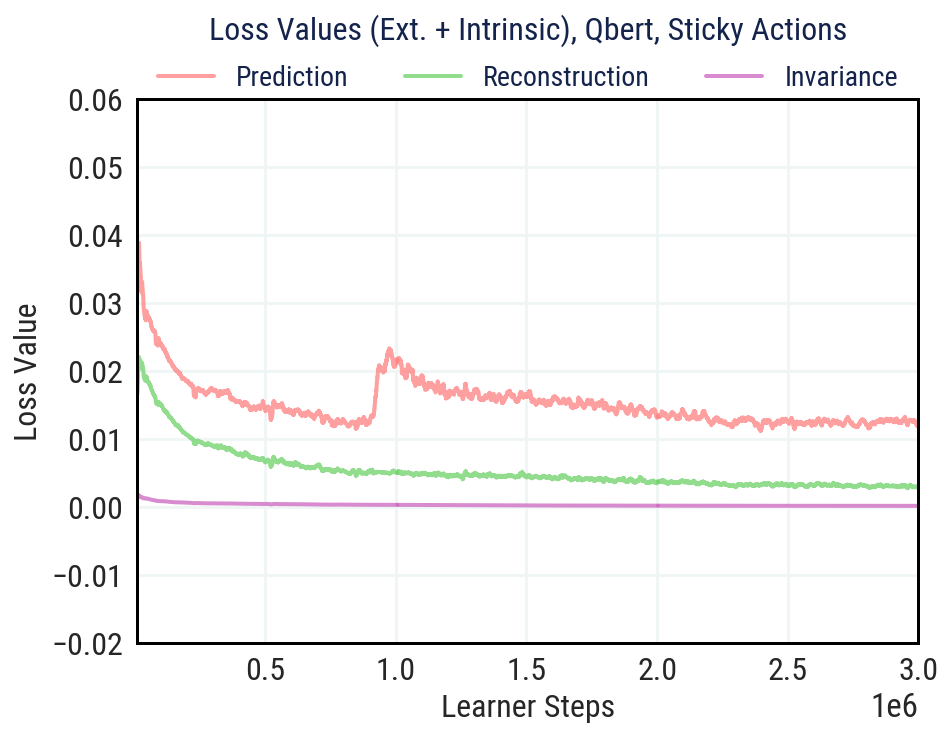}}
\hfill
\subfloat[\tiny\textbf{Solaris (E+I)}]{
\includegraphics[height=0.19\linewidth, trim=0em 0em 0em 0em]
{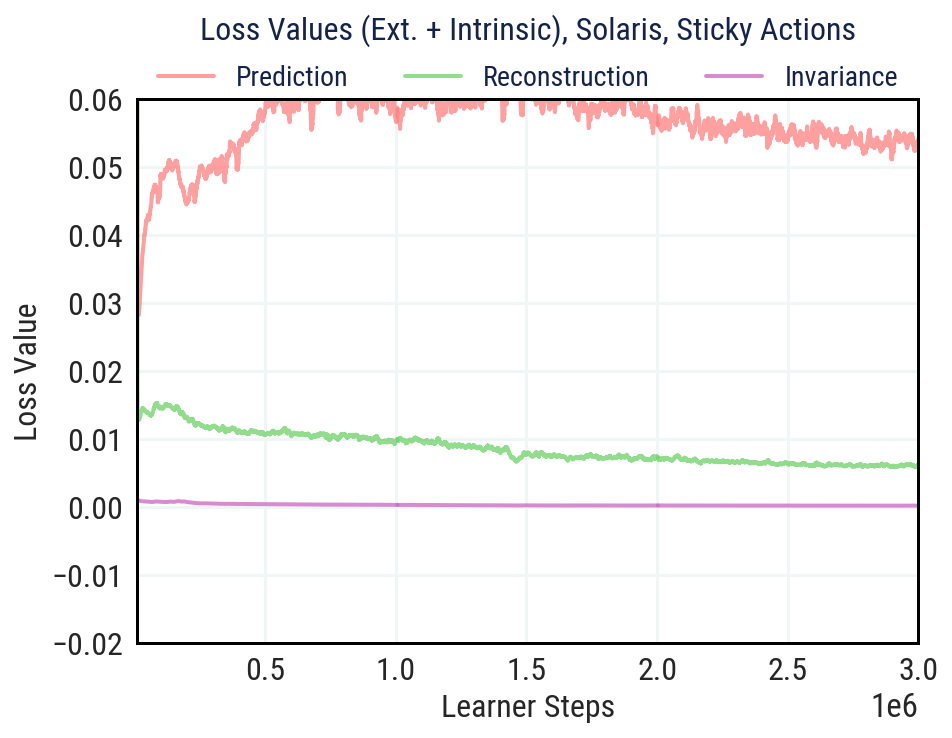}}
\hfill
\subfloat[\tiny\textbf{Venture (E+I)}]{
\includegraphics[height=0.19\linewidth, trim=0em 0em 0em 0em]
{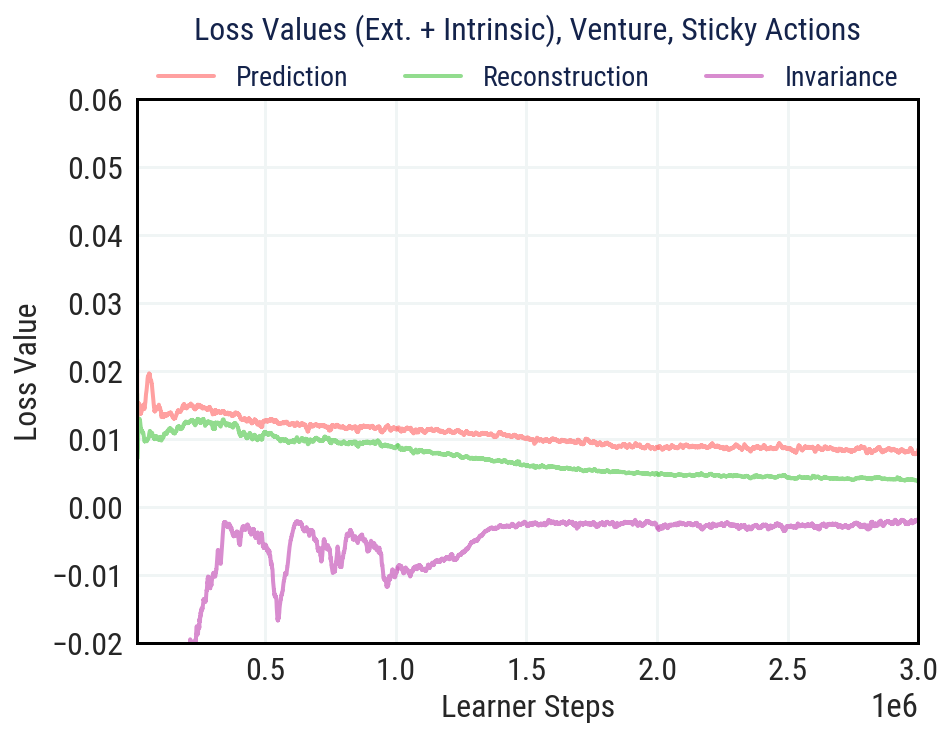}}
}
\vspace{-0.325em}

\caption{\small\dayum{\textit{Loss Values (Hard Exploration Games, with Sticky Actions)}.
\tttext{BYOL-Hindsight} prediction, reconstruction, invariance losses.
}}
\label{fig:atari:loss:big}
\vspace{-0.75em}
\end{figure*}

\dayum{
For additional insight into loss dynamics, Figure \ref{fig:atari:loss:big} shows the behavior of \tttext{BYOL-Hindsight}'s reconstruction and invariance losses. For comparison, a predictor is also trained to measure the usual forward prediction loss. We observe that the losses generally behave consistently with our hypothesis: Prediction losses are higher than reconstruction losses due to stochasticity.
}

\subsection{Additional Remarks on Results}\label{subapp:remarks}

\dayum{
\textbf{Bank Heist}~~ \textit{In Figure 7, why are the returns higher for natural traps (intrinsic only) when compared with natural traps (intrinsic + extrinsic)?}

First, note in Figure 7(a)--(b) that for both the "intrinsic-only" and "mixed" regimes, the agent is still improving over time, except that the latter is improving more slowly. This phenomenon is actually not necessarily surprising. 

Consider training an agent in the "intrinsic-only" and "mixed" regimes. Indeed, as training proceeds over time, we may generally expect that returns obtained in the latter eventually surpass returns obtained in the former---because it has access to the extrinsic reward signal itself. However, the key word here is "\textit{eventually}": In general, it is not always true that an agent in the "mixed" regime improves \textit{more quickly} than in the "intrinsic-only" regime. How quickly their returns improve with/without extrinsic rewards depends entirely on the specifics of the environment, including how the extrinsic rewards are distributed, as well as how their magnitudes compare to the intrinsic rewards from exploration. For example, if there are many opportunities for earning small extrinsic rewards, then in the "mixed" regime the agent may spend more time chasing after those extrinsic rewards during training, which may slow down their exploration of further parts of the environment that could actually yield more rewards later on.

The Bank Heist environment is characterized by this, which offers a plausible explanation for the observed result: The game consists of a series of different cities that can be entered and exited in sequence. In each city, extrinsic rewards are obtained by robbing banks (by running over them) and also by destroying police cars (by dropping dynamite onto them). In each city, police cars respawn over time, and banks respawn when police cars are destroyed, so there are many opportunities for small extrinsic rewards to be earned, unless the agent is caught or runs out of fuel. Now, in the "intrinsic-only" regime, we expect that the agent is rewarded greatly for entering each new city (which loads an entirely different maze onto the screen). In order to keep exploring new cities, the agent must learn to \textit{survive}, which requires learning to destroy police cars and rob banks to refuel on exit. So, pure exploration in Bank Heist is already aligned with episode returns. On the other hand, in the "mixed" regime, the agent is \textit{additionally} incentivized to maximize the number of banks robbed and police cars destroyed in any given city, because the number of police cars and banks robbed nonlinearly compound the extrinsic rewards earned in that city. However, doing so increases the risk of being caught or running out of fuel, which may slow down learning.

In Figure 7(a)--(b), we see that the "mixed" agent starts earning non-trivial returns earlier than the "intrinsic-only" agent: At 1.2m learner steps, almost 20,000 is earned by the former, compared to 2,000 by the latter. Subsequently, the former improves more slowly than the latter. But given the above discussion, this result is consistent with the observation that the "mixed" agent spends more time within each city trying to destroy more police cars and rob more banks, more so than the "intrinsic-only" agent, which is only incentivized to do so "sufficiently" for surviving to explore more cities---thereby incidentally earning more rewards in further cities as a side-effect. Again, note that we do expect the "mixed" agent to eventually surpass the "intrinsic-only" agent, but this may take much longer to happen (i.e. beyond the 3m learner steps in the experiment). Of course, the precise dynamics of learning depends on the coefficient that combines the intrinsic and extrinsic rewards. In this work, we do not tune this mixing coefficient for each individual environment for optimal learning speed, because our focus is instead simply on demonstrating a positive benefit of \textit{hindsight} for curiosity-driven exploration.

(Note that this phenomenon is not observed for the Sticky Actions setting, which makes sense: Combined with the fact that dynamite explodes at unpredictable times, the fact that actions are sticky means that precise timing in destroying police cars is impossible in this setting, moreover exiting the maze to refuel and visit a new city may require several drive-bys to succeed, therefore staying too long in a city becomes very dangerous. So in this case, the policy has extrinsic incentive to exit and refuel to new cities earlier than before, which counteracts the effect described above. This gives a plausible explanation for the observation that in this setting, the "mixed" agent spends less time than before in each city, but explores more cities and ends up earning more returns more quickly).}

\textbf{Montezuma's Revenge}~~ \textit{In Figures 8 and 10, for sticky actions, why is the number of rooms visited reduced by around 25\%, whereas the returns are reduced by around 75\%?}

Note that Figure 8 measures the number of different rooms the agent manages to discover over the training run, whereas Figure 10 measures the mean episodic return that the agent obtains. (This is similar to prior work, such as in [12] and [14]). Therefore they are not necessarily one-to-one proportional to each another, as the agent may successfully find more and more later rooms over training, while still spending the majority of its time in earlier rooms.

Broadly speaking, sticky actions have two negative effects on gameplay. First, it obviously adds \textit{randomness} to the environment's dynamics, which throws off curiosity-driven exploration (e.g. BYOL-Explore) since sticky actions are a source of stochastic traps. Second, it also simply makes playing the game \textit{more difficult}, since the agent has less control over what actions are actually executed (e.g. they can die more easily due to unfortunate sticky actions). Now, what BYOL-Hindsight does is mitigate the first problem, such that the agent's policy is optimized using intrinsic rewards that are more or less unaffected by action stickiness. However, it cannot change the fact that the game is in fact more difficult to play, which means that progressing/staying alive is harder. So while the agent may still manage to discover many rooms, it may spend most of its time in earlier rooms, so the mean episodic return is lower.

Consider a "heatmap" of the rooms visited in the game over the training run. Compared to the non-sticky setting, in the sticky setting the heatmap has higher heat in the earlier rooms, and lower heat in the later rooms. However, the number of rooms with non-zero heat do not differ by much.

\textbf{Random Network Distillation}~~ \textit{In Figure 6, why does the performance of RND drop suddenly after $\sim$400K steps?}

This is simply the "vanishing rewards" phenomenon (see e.g. [34] for this terminology). Many methods relying on some notion of "novelty" for intrinsic rewards tend to exhibit this over time: After the novelty of a state has vanished, the agent is not incentivized to visit it again. In a small environment---such as the Pycolab Maze environment in Figure 6---if rewards vanish for all states, then the agent is not incentivized to do anything at all. Note that RND is especially susceptible to this, since (unlike the other algorithms) it does not need to learn any dynamics mapping from histories to future states, but rather it is simply learning a mapping from $x_{t+1}$ to $f_{\text{random}}(x_{t+1})$ for some initially unknown $f_{\text{random}}$. So while its errors \textit{initially} provide good incentive to explore, it much more quickly "loses interest" due to vanished rewards, hence the observed drops.
\section{Further Implementation Detail}\label{app:c}

\begin{figure}[h!]
\vspace{-1.25em}
\subfloat[\textbf{BYOL-Explore}]{
\includegraphics[width=0.46\linewidth, trim=6em 0em 3em 6em, clip]{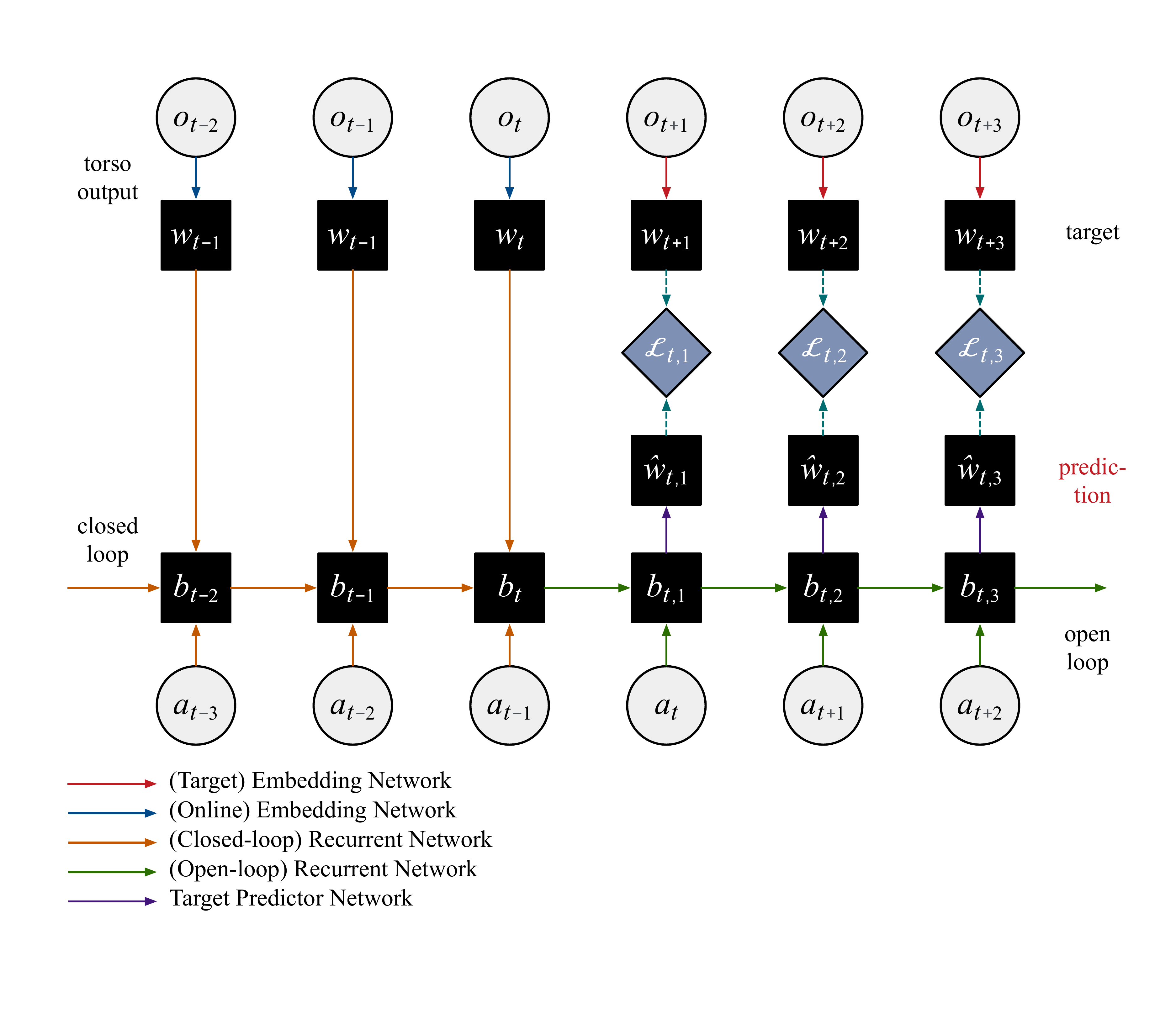}
}
\hfill
\subfloat[\textbf{BYOL-Hindsight}]{
\includegraphics[width=0.46\linewidth, trim=6em 0em 3em 6em, clip]{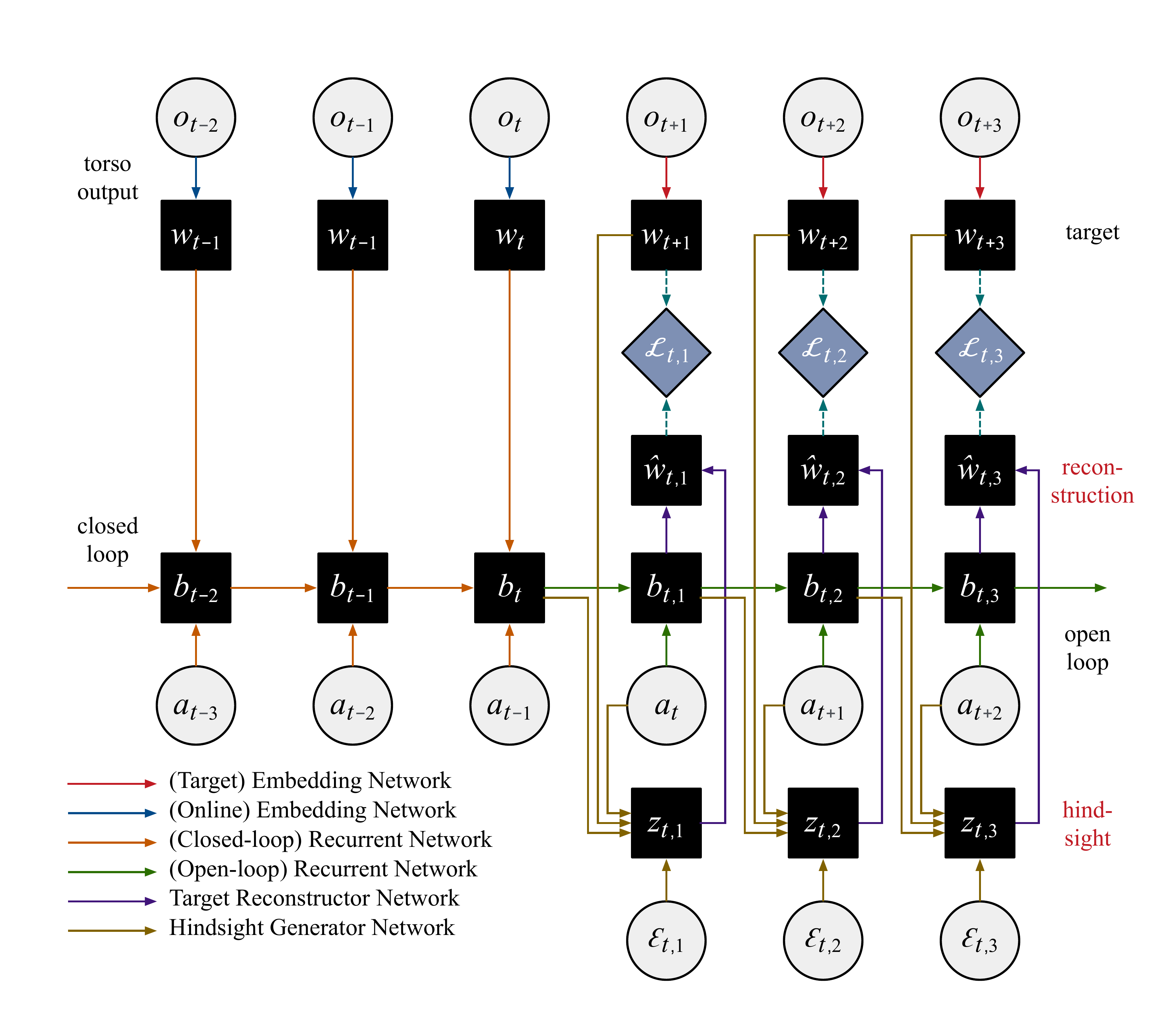}
}
\caption{\small\textit{Neural Architecture of BYOL-Explore and BYOL-Hindsight}.}
\label{fig:byolexplore}
\vspace{-0.5em}
\end{figure}

In all experiments, start from the same architecture/hyperparameters for \tttext{BYOL-Explore} as specified in \cite{guo2022byol}, including target network EMA, open-loop horizon, intrinsic reward normalization/prioritization, representation sharing, and underlying RL algorithm. In the following, we first recall the architecture of \tttext{BYOL-Explore} in detail, then describe \tttext{BYOL-Hindsight}:

\subsection{BYOL-Explore}

\textit{Online Embedding Network}.~
Figure \ref{fig:byolexplore}(a) shows the architecture of \tttext{BYOL-Explore}. First, to compute predictions of target states, an online network is composed of an encoder $\omega$ that transforms observations $o_{t}$ into representations $w_{t}=\omega(o_{t})$.

\dayum{
\textit{Closed-Loop Recurrent Network}.~
Next, a closed-loop RNN computes representations $b_{t}$ on the basis of previous actions \smash{$\{a_{t'}\}_{t'<t}$} and observation encodings \smash{$\{w_{t'}\}_{t'\leq t}$}. Specifically, the closed-loop RNN cell takes in each observation representation $w_{t}$, previous action $a_{t-1}$, and previous belief $b_{t-1}$ as input, and computes a representation $b_{t}$ of the history so far.
}

\textit{Open-Loop Recurrent Network}.~
Then, an open-loop RNN is initialized by this $b_{t}$, and computes forward predictions $b_{t,i}$ for horizon steps indexed as~$i$, on the basis of actions $\{a_{t'}\}_{t'\geq t}$ up to some maximum open-loop horizon. Specifically, the open-loop RNN cell takes in each action $a_{t+i-1}$ and current belief $b_{t,i-1}$ as input, and computes a representation $b_{t,i}$ of the predicted next belief. The purpose of this is to simulate future beliefs using only knowledge of future actions.

\textit{Target Predictor Network}.~
Lastly, a predictor network takes the open-loop belief $b_{t,i}$ as input, and outputs the open-loop (raw) prediction $\hat{w}_{t,i}$. (We say ``raw'' here in order to distinguish from the normalized predictions $\hat{x}_{t,i}$ below).

\textit{Target Embedding Network}.~
Corresponding to the online network is a target network whose parameters are an exponential moving average of the parameters of the online network. The target network encodes observations $o_{t+i}$ as $w_{t+i}=\omega_{\text{target}}(o_{t+i})$, and these targets are used to train the online network. The weights of $\omega_{\text{target}}$ are updated per the averaging rule $\omega_{\text{target}}\leftarrow\alpha\omega_{\text{target}}+(1-\alpha)\omega$ after each training step, with $\alpha$ being the exponential moving average parameter.

\textit{Loss Function}.~
Define target states as the $\ell_{2}$-normalized encodings $x_{t+i}\coloneqq\tttext{sg}(w_{t+i}/\|w_{t+i}\|_{2})$ of future observations, and predictions as the $\ell_{2}$-normalized (raw) predictions $\hat{x}_{t,i}\coloneqq \hat{w}_{t,i}/\|\hat{w}_{t,i}\|_{2}$. Moreover, define input states as the beliefs themselves---that is, $x_{t,i-1}\coloneqq b_{t,i-1}$. Then, the loss function used to train all networks (except the target network) is given by
\smash{$
\mathcal{R}(x_{t,i-1}^{(j)},a_{t+i-1}^{(j)})
\coloneqq
\|
x_{t+i}^{(j)}-\hat{x}_{t,i}^{(j)}
\|_{2}^{2}$}
where $j$ indexes the trajectories within a batch. The overall objective for training the networks is the average over the time, batch, and horizon dimensions.

\textit{Intrinsic Reward}.~
Finally, the intrinsic reward associated to each observed transition (\smash{$o_{s}^{(j)},a_{s}^{(j)},o_{s+1}^{(j)}$}) is the sum of corresponding prediction errors \smash{$\sum_{t+i=s+1}\mathcal{R}(x_{t,i-1}^{(j)},a_{t+i-1}^{(j)})$}, which aggregates all the errors pertaining to the world model relative to the observation \smash{$o_{s+1}^{(j)}$}---the intuition being that the intrinsic reward for a time step is proportional to how difficult it is to predict its observation from partial histories. See Algorithm \ref{alg:byole}.

\subsection{BYOL-Hindsight}

\textit{Target Reconstructor Network}.~
Figure \ref{fig:byolexplore}(b) shows the architecture of \tttext{BYOL-Hindsight}. Starting from the setup for \tttext{BYOL-Explore}, the modification to incorporate hindsight is as follows: First, the predictor network is now replaced by a reconstructor network, which takes the open-loop belief $b_{t,i}$ and hindsight vector $z_{t,i}$ as input, and outputs the open-loop (raw) reconstruction $\hat{w}_{t,i}$. (Notation: At this point, it is helpful to recall that the online embedding network, closed-loop RNN, open-loop RNN, and target predictor/reconstructor network all correspond to what we subsume under parameter $\eta$).

\textit{Generator and Critic Networks}.~
Second, a hindsight generator network $p_{\theta}$ takes in the belief $b_{t,i-1}$, the action $a_{t+i-1}$, and the target $w_{t+i}$, and samples a hindsight vector $Z_{t,i}\sim p_{\theta}(\cdot|b_{t,i-1},a_{t+i-1},w_{t+i})$ by taking an additional noise vector $\varepsilon_{t,i}$ as input.
Finally, a hindsight critic network $g_{\nu}$ takes in any belief $b_{t,i-1}$, any action $a_{t+i-1}$, and any hindsight vector $z_{t,i}$ as input, and outputs the corresponding energy $g_{\nu}(b_{t,i-1},a_{t+i-1},z_{t,i})$.

\textit{Loss Function}.~
There are now two components. Firstly, analogous to before, define target states as the $\ell_{2}$-normalized encodings $x_{t+i}\coloneqq\tttext{sg}(w_{t+i}/\|w_{t+i}\|_{2})$ of future observations, and reconstructions as the $\ell_{2}$-normalized (raw) reconstructions $\hat{x}_{t,i}\coloneqq \hat{w}_{t,i}/\|\hat{w}_{t,i}\|_{2}$.
Moreover, define input states as the beliefs themselves---that is, $x_{t,i-1}\coloneqq b_{t,i-1}$.
Then, the loss function used to train the online embedding network, closed-loop RNN, open-loop RNN, target reconstructor network, and hindsight generator network (i.e. all networks except the target network and critic network) is given by:
\smash{$
\mathcal{R}_{\theta,\eta}^{\text{rec.}}(x_{t,i-1}^{(j)},a_{t+i-1}^{(j)})
\coloneqq
\|
x_{t+i}^{(j)}-\hat{x}_{t,i}^{(j)}
\|_{2}^{2}$}
where $j$ indexes the trajectories within a batch. The overall (reconstructive) objective for training the networks is the average over the time, batch, and horizon dimensions.
Second, the critic needs to ensure that $Z_{t,i}$ be independent of $X_{t,i-1},A_{t+i-1}$. The loss function used to train the hindsight generator and critic networks is given by:
\smash{$
\mathcal{R}_{\theta,\nu}^{K,\text{con.}}(x_{t,i-1}^{(j)},a_{t+i-1}^{(j)})
$\pix$\coloneqq
$}
\smash{\raisebox{-1pt}{$
\log\big[
e^{g_{\nu}(x_{t,i-1}^{(j)},a_{t+i-1}^{(j)},z_{t,i}^{(j)})}
/
\frac{1}{K}(e^{g_{\nu}(x_{t,i-1}^{(j)},a_{t+i-1}^{(j)},z_{t,i}^{(j)})}+\sum_{k=1}^{K-1}e^{g_{\nu}(x_{t,i-1}^{(j)},a_{t+i-1}^{(j)},z_{t,i}^{(k)})})
\big]
$}}, where $k$ is another index into the trajectories within a batch. The overall (contrastive) objective for training the networks is the average over the time, batch, and horizon dimensions.

\textit{Intrinsic Reward}.~
Finally, analogous to before, the intrinsic reward associated to each observed transition (\smash{$o_{s}^{(j)},a_{s}^{(j)},o_{s+1}^{(j)}$}) is the sum of corresponding reconstruction$\pix$+$\pix$contrastive errors \smash{$\sum_{t+i=s+1}\big[\tfrac{1}{\lambda}\mathcal{R}_{\theta,\eta}^{\text{rec.}}(x_{t,i-1}^{(j)},a_{t+i-1}^{(j)})+\mathcal{R}_{\theta,\nu}^{K,\text{con.}}(x_{t,i-1}^{(j)},a_{t+i-1}^{(j)})\big]$}, which aggregates all the errors pertaining to the world model relative to the observation \smash{$o_{s+1}^{(j)}$}---the intuition being that the intrinsic reward for a time step is proportional to how difficult it is to reconstruct its observation from partial histories as well as how difficult it is to generate hindsight representations disentangled from input states and actions. See Algorithm \ref{alg:byolh}.

\begin{algorithm}[t]
\caption{BYOL-Explore}
\label{alg:byole}
\begin{algorithmic}
\STATE \textbf{repeat}\COMMENT{\small batch indices $j$ suppressed unless explicitly required}
\STATE ~~~~execute policy $\pi$ to obtain dataset of actions $a$ and observations $o$
\STATE ~~~~\textbf{for} $j=1,...$ \textbf{do}\COMMENT{\small batch index}
\STATE ~~~~~~~~compute online embedding, $\omega_{t}\leftarrow\omega(o_{t})$ for all time steps $t$
\STATE ~~~~~~~~\textbf{for} $t=1,...$ \textbf{do}\COMMENT{\small time step}
\STATE ~~~~~~~~~~~~compute closed-loop belief, $b_{t}\leftarrow\text{RNN}_{\text{closed-loop}}(\{a_{t'}\}_{t'<t},\{\omega_{t'}\}_{t'\leq t})$
\STATE ~~~~~~~~~~~~\textbf{for} $i=1,...$ \textbf{do}\COMMENT{\small horizon}
\STATE ~~~~~~~~~~~~~~~~compute open-loop belief, $b_{t,i}\leftarrow\text{RNN}_{\text{open-loop}}(b_{t}, \{a_{t'}\}_{i>t'\geq t})$
\STATE ~~~~~~~~~~~~~~~~compute target embedding, $\omega_{t+i}\leftarrow\omega_{\text{target}}(o_{t+i})$ and normalized target, $x_{t+i}\leftarrow\tttext{sg}(w_{t+i}/\|w_{t+i}\|_{2})$
\STATE ~~~~~~~~~~~~~~~~compute predicted embedding, $\hat{\omega}_{t,i} \leftarrow\psi(b_{t,i})$ and normalized prediction, $\hat{x}_{t,i}\coloneqq \hat{w}_{t,i}/\|\hat{w}_{t,i}\|_{2}$
\STATE ~~~~~~~~~~~~\textbf{end for}
\STATE ~~~~~~~~\textbf{end for}
\STATE ~~~~\textbf{end for}
\STATE ~~~~update $\omega,\text{RNN}_{\text{closed-loop}},\text{RNN}_{\text{open-loop}},\psi$ using $\mathcal{R}(x_{t,i-1}^{(j)},a_{t+i-1}^{(j)})\coloneqq\|
x_{t+i}^{(j)}-\hat{x}_{t,i}^{(j)}
\|_{2}^{2}$ averaged over $i,j,t$
\STATE ~~~~update $\omega_{\text{target}}$ using exponential moving averaging, $\omega_{\text{target}}\leftarrow\alpha\omega_{\text{target}}+(1-\alpha)\omega$
\STATE ~~~~update $\pi$ using \smash{$\mathcal{R}_{\text{intrinsic}}(o_{s},a_{s},o_{s+1})\coloneqq\sum_{t+i=s+1}\mathcal{R}(x_{t,i-1},a_{t+i-1})$}
\STATE \textbf{end repeat}
\end{algorithmic}
\end{algorithm}

\begin{algorithm}[t]
\caption{BYOL-Hindsight}
\label{alg:byolh}
\begin{algorithmic}
\STATE \textbf{repeat}\COMMENT{\small batch indices $j$ suppressed unless explicitly required}
\STATE ~~~~execute policy $\pi$ to obtain dataset of actions $a$ and observations $o$
\STATE ~~~~\textbf{for} $j=1,...$ \textbf{do}\COMMENT{\small batch index}
\STATE ~~~~~~~~compute online embedding, $\omega_{t}\leftarrow\omega(o_{t})$ for all time steps $t$
\STATE ~~~~~~~~\textbf{for} $t=1,...$ \textbf{do}\COMMENT{\small time step}
\STATE ~~~~~~~~~~~~compute closed-loop belief, $b_{t}\leftarrow\text{RNN}_{\text{closed-loop}}(\{a_{t'}\}_{t'<t},\{\omega_{t'}\}_{t'\leq t})$
\STATE ~~~~~~~~~~~~\textbf{for} $i=1,...$ \textbf{do}\COMMENT{\small horizon}
\STATE ~~~~~~~~~~~~~~~~compute open-loop belief, $b_{t,i}\leftarrow\text{RNN}_{\text{open-loop}}(b_{t}, \{a_{t'}\}_{i>t'\geq t})$
\STATE ~~~~~~~~~~~~~~~~compute target embedding, $\omega_{t+i}\leftarrow\omega_{\text{target}}(o_{t+i})$ and normalized target, $x_{t+i}\leftarrow\tttext{sg}(w_{t+i}/\|w_{t+i}\|_{2})$
\STATE ~~~~~~~~~~~~~~~~sample hindsight vector, $Z_{t,i}\sim p_{\theta}(\cdot|b_{t,i-1},a_{t+i-1},w_{t+i})$
\STATE ~~~~~~~~~~~~~~~~evaluate hindsight energy, $g_{\nu}(b_{t,i-1},a_{t+i-1},z_{t,i})$
\STATE ~~~~~~~~~~~~~~~~compute reconstructed embedding, $\hat{\omega}_{t,i} \leftarrow\psi(b_{t,i},z_{t,i})$ and normalized prediction, $\hat{x}_{t,i}\coloneqq \hat{w}_{t,i}/\|\hat{w}_{t,i}\|_{2}$
\STATE ~~~~~~~~~~~~\textbf{end for}
\STATE ~~~~~~~~\textbf{end for}
\STATE ~~~~\textbf{end for}
\STATE ~~~~update $\omega,\text{RNN}_{\text{closed-loop}},\text{RNN}_{\text{open-loop}},\psi,p_{\theta}$ using $\mathcal{R}_{\theta,\eta}^{\text{rec.}}(x_{t,i-1}^{(j)},a_{t+i-1}^{(j)})\coloneqq\|
x_{t+i}^{(j)}-\hat{x}_{t,i}^{(j)}
\|_{2}^{2}$ averaged over $i,j,t$
\STATE ~~~~update $p_{\theta},g_{\nu}$ using $
\mathcal{R}_{\theta,\nu}^{K,\text{con.}}(x_{t,i-1}^{(j)},a_{t+i-1}^{(j)})
$\pix$\coloneqq
$
$
\log\big[
e^{g_{\nu}(x_{t,i-1}^{(j)},a_{t+i-1}^{(j)},z_{t,i}^{(j)})}
/
\frac{1}{K}(e^{g_{\nu}(x_{t,i-1}^{(j)},a_{t+i-1}^{(j)},z_{t,i}^{(j)})}$
\STATE ~~~~~~~~$+\sum_{k=1}^{K-1}e^{g_{\nu}(x_{t,i-1}^{(j)},a_{t+i-1}^{(j)},z_{t,i}^{(k)})})\big]$ averaged over $i,j,t$
\STATE ~~~~update $\omega_{\text{target}}$ using exponential moving averaging, $\omega_{\text{target}}\leftarrow\alpha\omega_{\text{target}}+(1-\alpha)\omega$
\STATE ~~~~update $\pi$ using \smash{$\mathcal{R}_{\text{intrinsic}}(o_{s},a_{s},o_{s+1})\coloneqq\sum_{t+i=s+1}\big[\tfrac{1}{\lambda}\mathcal{R}_{\theta,\eta}^{\text{rec.}}(x_{t,i-1},a_{t+i-1})+\mathcal{R}_{\theta,\nu}^{K,\text{con.}}(x_{t,i-1},a_{t+i-1})\big]$}
\STATE \textbf{end repeat}
\end{algorithmic}
\end{algorithm}

\subsection{RL Hyperparameters}

Like with \tttext{BYOL-Explore}, any RL algorithm can be used in conjunction with \tttext{BYOL-Hindsight}. We use VMPO \cite{song2019v} exactly as specified in \cite{guo2022byol}, and reproduce the details as follows: PopArt-style \cite{hessel2019multi} reward normalization is used with step size 0.01, and rewards are subsequently rescaled by $1-\gamma$ with discount factor $\gamma=0.999$. PopArt normalization is also applied to the output of the value network. To train the value function, VTrace is used without off-policy correction to define temporal-difference targets for mean squared error loss with loss weight 0.5, and an entropy loss with loss weight 0.001 is added. The parameters $\eta_{\text{init}}$ and $\alpha_{\text{init}}$ for VMPO are initialized to 0.5, and $\epsilon_{\eta}=0.01$ and $\epsilon_{\alpha}=0.005$.
The top-$k$ parameter for VMPO is set to 0.5. For optimization, the Adam optimizer is used with learning rate $10^{-4}$ and $b_{1}=0.9$. The VMPO target network is updated every 10 learner steps.
In terms of computation, 400 CPU actors generate data through an inference server, using four TPUv2 for evaluating the policy. Curiosity in Hindsight is agnostic to the underlying reinforcement learning algorithm used to optimize intrinsic rewards, so all RL implementation details in \tttext{BYOL-Hindsight} are identical to those in the original \tttext{BYOL-Explore} experiments.

\subsection{BYOL Hyperparameters}

For the components of \tttext{BYOL-Hindsight} that overlap with \tttext{BYOL-Explore}, we use the exact same architecture and hyperparameters. Observation representations of size 512 and history representations of size 256.
The encoder is a Deep ResNet stack \cite{he2016deep}, with grayscale image observations passing through a stack of 3 units, each made up of a 3$\times$3 convolutional layer, a 3$\times$3 max pool layer, and two residual blocks. The convolutional layer and residual blocks have number of channels (16, 32, 32) for each of the 3 units. GroupNorm normalization \cite{wu2018group} is used with one group at the end of each unit, and ReLU activations are used everywhere. At the end of the final residual block, the output is flattened and projected with a single linear layer to embedding dimension 512.
The closed-loop and open-loop RNNs are simple GRUs \cite{cho2014properties}, with actions provided to the RNN cells, embedded to representation size 32.
The policy head and value head are MLPs with one hidden layer of size 256, and the outputs of the policy head are passed through a softmax layer to obtain action probabilities.
Specifically for \tttext{BYOL-Explore} only, the predictor network is an MLP with three hidden layers of size 512 (Note that this detail is the first of two that are different to the original implementation in \cite{guo2022byol}; using three layers of 512 instead of the original single layer of 256 leads to better results).
In the mixed exploration regime, the intrinsic/extrinsic rewards mixing coefficient is 0.2 (This is the second of two details that are different to the original implementation in \cite{guo2022byol}; using 0.2 instead of the original 0.1 leads to better results).
We defer to \cite{guo2022byol} for details on intrinsic reward normalization/prioritization and representation sharing.
We use the classical 30 random no-ops evaluation regime for Atari \cite{mnih2015human,van2016deep}.
The batch size is 32 and sequence length is 128, and four TPUv2 are used in a distributed learning setup.
The open loop horizon is 1 for all Pycolab experiments, and 8 for all Atari experiments.
The target network EMA is 0.99.

\subsection{Hindsight Hyperparameters}

Specifically for \tttext{BYOL-Hindsight}, the reconstructor network is an MLP with three hidden layers of 512, which is the same as the predictor network in \tttext{BYOL-Explore} above. The generator network and critic network are MLPs with three hidden layers of 512. The dimension of the generator noise $\epsilon$ is 256, and the dimension of the hindsight vector is 256. The temperature parameter is $0.5$, except in Montezuma's revenge where we show sensitivity to the temperature. The coefficient $\lambda$\pix$=$\pix$1$ for model learning. For policy optimization, we empirically observe that the value of $\lambda$ has little to no contribution towards the intrinsic reward (and little to no effect on exploration); for simplicity we set $\lambda$ to zero for policy optimization.
For the contrastive loss, negative samples are simply taken from the batch, so the contrastive set is also batch size 32; the time dimension is not used as negatives. For optimization, the Adam optimizer is used with learning rate $10^{-4}$ and $b_{1}=0.9$ for both the reconstruction loss and contrastive loss. Alternating optimization is used to optimize the critic, with single optimization steps for the critic interleaved with single optimization steps for the rest of the networks.
Where necessary to provide multiple input vectors to the reconstructor, generator, and critic networks, inputs are first combined by concatenation before being fed into the first layer of the networks.

\textit{Note}: Relative to the dimensionality of the "true" source of stochasticity in the world (i.e. within any reparameterized model that accurately captures the world's dynamics), the dimensionality of the generator noise $\epsilon$ and hindsight vector $Z$ should not be too small. If it is too small, it may have insufficient capacity to model the stochasticity well, which means there may still be remaining stochastic traps that can negatively affect curiosity-based exploration. On the other hand, making it very large would avoid this problem, but may potentially make learning proceed more slowly. This is because not only does $Z$ have to capture stochasticity, it also has to obey the constraint that it be independent of states and actions, so in very high dimensions this may not be as easy to simultaneously optimize.
Of course, the "true" dimensionality of stochasticity is rarely known, so we may make an educated guess when setting sensible values for the noise and hindsight dimensions. In our experiments, we do not tune these dimensions, and simply set it to 256 for all environments and experiments.
\section{Discussion and Related Work}\label{app:d}

\subsection{Additional Discussion}

It is important to account for stochasticity in reinforcement learning---especially in exploration: Stochasticity may arise naturally due to a variety of factors, such as inherent randomness in the world (e.g. coin flip), or imperfect observations or actions (e.g. faulty sensors or actuators), or un-modeled complexity (e.g. model mismatch or imperfect optimization), or simply due to the existence of other agents (e.g. in a multi-agent game)---all of which would lead to world dynamics that appear stochastic to the agent. In this work, we used a variety of settings to test our hypothesis that hindsight information can mitigate stochastic traps in predictive error-based exploration. Overall, our results verify the fact that learned hindsight representations are able to disentangle the various (unpredictable) stochasticities from the rest of the (predictable) dynamics of the world. Future work may investigate applicability to other scenarios, such as robotics settings, or multi-agent settings.

\dayum{
The invariance objective is reminiscent of that used in counterfactual credit assignment \cite{mesnard2021counterfactual}, where hindsight information in future-conditional value functions are constrained to not contain information about the agent's actions. A key difference is that in the exploration setting, the invariance constraint primarily serves as a way to ensure that intrinsic rewards do not fall to zero prematurely (i.e. when too much information is leaked about the future), so performance is not so sensitive to violations of the independence constraint---unlike in counterfactual credit assignment, where it takes paramount importance for estimators to be unbiased. Another difference is that hindsight variables are deterministic functions of the future in counterfactual credit assignment, whereas in our case they must be stochastic in order to accommodate the general case of any stochasticity.
}

Finally, it is worth emphasizing that the world model that is the focus of curiosity-driven exploration as studied in this work (i.e. either the predictive one in \tttext{BYOL-Explore}, or the reconstructive one in \tttext{BYOL-Hindsight}) is only designed for computing rewards. It need not be related to the underlying RL algorithm, which can be model-free (as is the case for VMPO). While this is a flexibility, it could also be a limitation: Future work may more systematically explore the advantages and disadvantages of sharing learned exploratory world models or representations with the underlying RL algorithm itself.
As another remark for clarification, note that ``contrastive learning'' here refers to a slightly different objective than what is typically referred to in self-supervised representation learning (see e.g. \cite{guo2022byol}), or in noise contrastive estimation (see e.g. \cite{jarrett2021time}). Indeed, a key innovation in BYOL (and inherited by BYOL-Explore) is that contrastive learning using negative samples are not involved at all. In this work, we start from BYOL-Explore but re-introduce contrastive learning for an orthogonal objective---that is, to learn a hindsight vector for addressing stochastic traps.

\textbf{Quality of World Model}~~ From the perspective of the agent, stochasticity can arise due to a variety of reasons. Inherent randomness in the world is one (e.g. a coin flip), imperfect observations is another (e.g. faulty sensors), and imperfectly executed actions is also another (e.g. faulty actuators). Yet another source of "stochasticity" is that the world model is bad, e.g. due to insufficient capacity to model the real world, or due to imperfect optimization procedure. To the agent, this simply appears that there is remaining "stochasticity" even after long periods of training. Importantly, if there is are particular parts of the world that the model is insufficiently expressive to capture relative to other parts of the world, then they effectively become stochastic traps that the agent may become stuck around.
Note that this problem could affect all curiosity-driven exploration methods (as given by Definition 1), and could affect RND, ICM, and BYOL-Explore. Actually, Curiosity in Hindsight should generally potentially \textit{mitigate} this kind of problem, because the hindsight generator will attempt to learn hindsight vectors that capture this otherwise remaining "stochasticity". Now of course, we may then ask the (meta) question: Even with the additional hindsight capacity, what if prediction/reconstruction is still bad, e.g. if the augmented model still has insufficient capacity to model the real world? In general, regardless of whether we are using the augmented model or not, remaining stochasticity is a problem if it is \textit{unevenly} distributed around the world (e.g. high around specific objects, which become traps), but it is not a problem otherwise. So practically, this means the final model needs to be at least "good enough" such that errors are not due to failures to model specific areas of the state space. In our experiments, we empirically observe that reconstruction error becomes close to zero but above zero, but the agents do not get trapped by any remaining unevenly distributed noise.
Finally, going forward, it is beneficial for Crafter \cite{hafner2021benchmarking} to be a testbed for curiosity-driven exploration (and curiosity in hindsight) in future updates to this agenda, because it is also a partially observable stochastic environment in which predictive world models may not easily capture all outcomes.

\textbf{Effects of Two Losses}~~ 
The overall exploration algorithm operates by having the policy parameters maximize the intrinsic rewards, while the model parameters minimize the intrinsic rewards. So, by default the reconstruction loss and contrastive loss affect both the policy and the model. Starting from this base case, we can ask what happens if either loss is omitted, for either policy learning or model learning.
\textit{Model Learning}: This is easiest to reason about. On one hand, if the contrastive loss were omitted from model learning, then there is no reason at all for hindsight vectors $Z$ to be independent of $X,A$, thus $Z$ may very quickly learn to copy all of the information in the outcome itself. This means that reconstruction errors (which yield intrinsic reward for the agent's policy) quickly drops to zero, without the agent having to explore at all. This is disastrous for exploration. On the other hand, if the reconstruction loss were omitted from model learning, then reconstruction errors (which yield intrinsic reward for the agent's policy) will never improve over time no matter how much experience the agent has, which is also pathological for exploration. In sum, to avoid breakdown of exploration, it must be the case that both the reconstruction loss and contrastive loss are applied to model learning.
\textit{Policy Learning}: This is also very easy to reason about. We empirically find that in practice, the invariance constraint is always respected very well (see e.g. Figure 15, where the invariance loss remains close to zero throughout training). Therefore the magnitude of the contrastive component of the intrinsic reward always has little to no contribution to the total intrinsic reward. Empirically, indeed exploration behavior is rather unaffected by whether or not the contrastive loss is included as an intrinsic reward for policy learning. On the other hand, suppose we remove the reconstruction component of the intrinsic reward. Then the agent would practically have no incentive to explore, which is disastrous. In sum, for policy learning the scaling between the two intrinsic rewards is not meaningful, as the reconstruction term dominates.

\textbf{Choice of Baseline Method}~~ Briefly, within the curiosity-driven exploration paradigm (Definition 1), there are two primary choices when it comes to representation: How to represent "input states" $X_{t}$, and how to represent "target states" $X_{t+1}$.
For the former, we simply make the most general choice of using learned RNN "belief" representations for $X_{t}$---that is, rollups of previous actions $\{a_{t'}\}_{t'<t}$ and observation encodings $\{\omega(o_{t'})\}_{t'\leq t}$, where $\omega$ is a learned encoding function. This should be uncontroversial, and the vast majority of methods in Table 1 operate this way, since pretty much any non-trivial environment would require the agent to be aware of histories rather than just immediate-state contexts or features.
For the latter, we choose BYOL-Explore, which has that "target states" are $\ell_{2}$-normalized encodings of future observations, with the target encoding function $\omega_{\text{target}}$ being an exponential moving average of $\omega$. This is due to three reasons: First, pixel-based curiosity has been found too perform worse than using learned representations [9, 15]. So, among popular and successful representation methods, we are looking at autoencoded features, random features, inverse dynamics features, and the recently proposed bootstrapped features. Second, moreover, autoencoded features have been found to be unstable in practice [9], and also bootstrapped features have most recently been found to yield the best performance on benchmarks [14]. Third, bootstrapped representations arguably yield the simplest learning algorithm, as the target embedding network is just the exponential moving average of the online embedding network. In sum, this is why we chose BYOL-Explore as the baseline formulation---because it serves a simple and already state-of-the-art baseline on top of which we can demonstrate how Curiosity in Hindsight further improves its performance in stochastic environments.

\textbf{Generality of Framework}~~ The practical framework for Curiosity in Hindsight (as defined by Equations 21--22) can be used to augment any \textit{curiosity-driven} method (as defined by Equations 1--2). As shown in Table 1, this includes a number of recent methods, i.e. as long as they can be expressed in the form of Definition 1. (Moreover, our Response (A.1) above discusses the main reasons why we particularly select BYOL-Explore as our prime example to augment and conduct empirical experiments for).
On the other hand, there are other exploration paradigms that \textit{cannot} be expressed in the form of Definition 1 (i.e. not "curiosity-driven" as defined there), and hence cannot be readily augmented with our framework for Curiosity in Hindsight. As discussed in Section 2.2 ("Related Work"), this includes for instance methods based on visitation counts, hashes, and density estimates, as well as methods based on estimated uncertainties about the world, such as taking actions that maximize the estimated information gain, etc.
For instance, exploration by the Plan2Explore method operates by first estimating "novelty" through ensemble disagreement in latent predictions made by 1-step transition models, and the agent uses a concurrently trained global recurrent world model to plan to explore on the basis of this novelty measure. So this method operates by directly estimating/maximizing the expected information gain using ensemble disagreement, therefore it does not fall within the curiosity-driven paradigm (i.e. cannot be formulated as an instance of Definition 1), and hence cannot be plugged into Curiosity in Hindsight.

\subsection{Additional Related Work}

\dayum{
Several related works are concurrent to ours.
In the novelty-based family, \cite{henaff2022exploration} extends count-based episodic bonuses to continuous state spaces and encourages exploring states that are diverse under a learned embedding, and \cite{anonymous2022robust} proposes clustering-based density estimation to model a wide range of timescales.
Tackling the problem that stochasticity poses for predictive error-based exploration as we do, \cite{piot2022blade} is an another extension of \tttext{BYOL-Explore} that approaches stochasticity in world dynamics by deliberately and directly leaking a noisy version of future information to the predictor.
Theoretically, \cite{tang2022understanding} studies the learning dynamics of self-predictive learning for reinforcement learning, which is related to \tttext{BYOL-Explore}.
Finally, \cite{yang2022dichotomy} augments future-conditional supervised learning with the ability to remove uncontrollable information from the future-conditioning variable---which is the ``mirror image'' of what we seek from hindsight representations in our framework.
}


\clearpage
\twocolumn

\bibliographystyle{unsrt}
\bibliography{bib}

\begin{thebibliography}{100}

\bibitem{kriegeskorte2018cognitive}
Nikolaus Kriegeskorte and Pamela~K Douglas.
\newblock Cognitive computational neuroscience.
\newblock {\em Nature neuroscience}, 21(9):1148--1160, 2018.

\bibitem{yang2021exploration}
Tianpei Yang, Hongyao Tang, Chenjia Bai, Jinyi Liu, Jianye Hao, Zhaopeng Meng,
  and Peng Liu.
\newblock Exploration in deep reinforcement learning: a comprehensive survey.
\newblock {\em arXiv preprint arXiv:2109.06668}, 2021.

\bibitem{schmidhuber1991possibility}
J{\"u}rgen Schmidhuber.
\newblock A possibility for implementing curiosity and boredom in
  model-building neural controllers.
\newblock In {\em Proc. of the international conference on simulation of
  adaptive behavior: From animals to animats}, pages 222--227, 1991.

\bibitem{thrun1995exploration}
Sebastian Thrun.
\newblock Exploration in active learning.
\newblock {\em Handbook of Brain Science and Neural Networks}, pages 381--384,
  1995.

\bibitem{barto2004intrinsically}
Andrew~G Barto, Satinder Singh, Nuttapong Chentanez, et~al.
\newblock Intrinsically motivated learning of hierarchical collections of
  skills.
\newblock In {\em Proceedings of the 3rd International Conference on
  Development and Learning}, pages 112--19. Piscataway, NJ, 2004.

\bibitem{oh2015action}
Junhyuk Oh, Xiaoxiao Guo, Honglak Lee, Richard~L Lewis, and Satinder Singh.
\newblock Action-conditional video prediction using deep networks in atari
  games.
\newblock {\em Advances in neural information processing systems}, 28, 2015.

\bibitem{finn2016unsupervised}
Chelsea Finn, Ian Goodfellow, and Sergey Levine.
\newblock Unsupervised learning for physical interaction through video
  prediction.
\newblock {\em Advances in neural information processing systems}, 29, 2016.

\bibitem{gregor2019shaping}
Karol Gregor, Danilo Jimenez~Rezende, Frederic Besse, Yan Wu, Hamza Merzic, and
  Aaron van~den Oord.
\newblock Shaping belief states with generative environment models for rl.
\newblock {\em Advances in Neural Information Processing Systems}, 32, 2019.

\bibitem{burda2019large}
Yuri Burda, Harri Edwards, Deepak Pathak, Amos Storkey, Trevor Darrell, and
  Alexei~A Efros.
\newblock Large-scale study of curiosity-driven learning.
\newblock {\em International Conference on Learning Representations}, 2019.

\bibitem{stadie2016incentivizing}
Bradly~C Stadie, Sergey Levine, and Pieter Abbeel.
\newblock Incentivizing exploration in reinforcement learning with deep
  predictive models.
\newblock {\em International Conference on Learning Representations}, 2016.

\bibitem{pathak2017curiosity}
Deepak Pathak, Pulkit Agrawal, Alexei~A Efros, and Trevor Darrell.
\newblock Curiosity-driven exploration by self-supervised prediction.
\newblock In {\em International conference on machine learning}, pages
  2778--2787. PMLR, 2017.

\bibitem{burda2019exploration}
Yuri Burda, Harrison Edwards, Amos Storkey, and Oleg Klimov.
\newblock Exploration by random network distillation.
\newblock {\em International Conference on Learning Representations}, 2019.

\bibitem{kim2019emi}
Hyoungseok Kim, Jaekyeom Kim, Yeonwoo Jeong, Sergey Levine, and Hyun~Oh Song.
\newblock Emi: Exploration with mutual information.
\newblock In {\em International Conference on Machine Learning}, pages
  3360--3369. PMLR, 2019.

\bibitem{guo2022byol}
Zhaohan~Daniel Guo, Shantanu Thakoor, Miruna P{\^\i}slar, Bernardo~Avila Pires,
  Florent Altch{\'e}, Corentin Tallec, Alaa Saade, Daniele Calandriello,
  Jean-Bastien Grill, Yunhao Tang, et~al.
\newblock Byol-explore: Exploration by bootstrapped prediction.
\newblock {\em Advances in neural information processing systems}, 35, 2022.

\bibitem{mavor2022stay}
Augustine Mavor-Parker, Kimberly Young, Caswell Barry, and Lewis Griffin.
\newblock How to stay curious while avoiding noisy tvs using aleatoric
  uncertainty estimation.
\newblock In {\em International Conference on Machine Learning}, pages
  15220--15240. PMLR, 2022.

\bibitem{choshen2018dora}
Leshem Choshen, Lior Fox, and Yonatan Loewenstein.
\newblock Dora the explorer: Directed outreaching reinforcement
  action-selection.
\newblock {\em International Conference on Learning Representations}, 2018.

\bibitem{orseau2013universal}
Laurent Orseau, Tor Lattimore, and Marcus Hutter.
\newblock Universal knowledge-seeking agents for stochastic environments.
\newblock In {\em International conference on algorithmic learning theory},
  pages 158--172. Springer, 2013.

\bibitem{hong2020adversarial}
Zhang-Wei Hong, Tsu-Jui Fu, Tzu-Yun Shann, and Chun-Yi Lee.
\newblock Adversarial active exploration for inverse dynamics model learning.
\newblock In {\em Conference on Robot Learning}, pages 552--565. PMLR, 2020.

\bibitem{pathak2019self}
Deepak Pathak, Dhiraj Gandhi, and Abhinav Gupta.
\newblock Self-supervised exploration via disagreement.
\newblock In {\em International conference on machine learning}, pages
  5062--5071. PMLR, 2019.

\bibitem{kim2020active}
Kuno Kim, Megumi Sano, Julian De~Freitas, Nick Haber, and Daniel Yamins.
\newblock Active world model learning with progress curiosity.
\newblock In {\em International conference on machine learning}, pages
  5306--5315. PMLR, 2020.

\bibitem{henaff2019explicit}
Mikael Henaff.
\newblock Explicit explore-exploit algorithms in continuous state spaces.
\newblock {\em Advances in Neural Information Processing Systems}, 32, 2019.

\bibitem{shyam2019model}
Pranav Shyam, Wojciech Ja{\'s}kowski, and Faustino Gomez.
\newblock Model-based active exploration.
\newblock In {\em International conference on machine learning}, pages
  5779--5788. PMLR, 2019.

\bibitem{osband2016deep}
Ian Osband, Charles Blundell, Alexander Pritzel, and Benjamin Van~Roy.
\newblock Deep exploration via bootstrapped dqn.
\newblock {\em Advances in neural information processing systems}, 29, 2016.

\bibitem{osband2018randomized}
Ian Osband, John Aslanides, and Albin Cassirer.
\newblock Randomized prior functions for deep reinforcement learning.
\newblock {\em Advances in Neural Information Processing Systems}, 31, 2018.

\bibitem{strehl2008analysis}
Alexander~L Strehl and Michael~L Littman.
\newblock An analysis of model-based interval estimation for markov decision
  processes.
\newblock {\em Journal of Computer and System Sciences}, 74, 2008.

\bibitem{tang2017exploration}
Haoran Tang, Rein Houthooft, Davis Foote, Adam Stooke, OpenAI Xi~Chen, Yan
  Duan, John Schulman, Filip DeTurck, and Pieter Abbeel.
\newblock \# exploration: A study of count-based exploration for deep
  reinforcement learning.
\newblock {\em Advances in neural information processing systems}, 30, 2017.

\bibitem{bellemare2016unifying}
Marc Bellemare, Sriram Srinivasan, Georg Ostrovski, Tom Schaul, David Saxton,
  and Remi Munos.
\newblock Unifying count-based exploration and intrinsic motivation.
\newblock {\em Advances in neural information processing systems}, 29, 2016.

\bibitem{ostrovski2017count}
Georg Ostrovski, Marc~G Bellemare, A{\"a}ron Oord, and R{\'e}mi Munos.
\newblock Count-based exploration with neural density models.
\newblock In {\em International conference on machine learning}, pages
  2721--2730. PMLR, 2017.

\bibitem{zhao2019curiosity}
Rui Zhao and Volker Tresp.
\newblock Curiosity-driven experience prioritization via density estimation.
\newblock {\em Advances in neural information processing systems}, 31, 2018.

\bibitem{domingues2021density}
Omar~Darwiche Domingues, Corentin Tallec, Remi Munos, and Michal Valko.
\newblock Density-based bonuses on learned representations for reward-free
  exploration in deep reinforcement learning.
\newblock In {\em ICML 2021 Workshop on Unsupervised Reinforcement Learning},
  2021.

\bibitem{fu2017ex2}
Justin Fu, John Co-Reyes, and Sergey Levine.
\newblock Ex2: Exploration with exemplar models for deep reinforcement
  learning.
\newblock {\em Advances in neural information processing systems}, 30, 2017.

\bibitem{flet2021adversarially}
Yannis Flet-Berliac, Johan Ferret, Olivier Pietquin, Philippe Preux, and
  Matthieu Geist.
\newblock Adversarially guided actor-critic.
\newblock {\em International Conference on Learning Representations}, 2021.

\bibitem{savinov2019episodic}
Nikolay Savinov, Anton Raichuk, Rapha{\"e}l Marinier, Damien Vincent, Marc
  Pollefeys, Timothy Lillicrap, and Sylvain Gelly.
\newblock Episodic curiosity through reachability.
\newblock {\em International Conference on Learning Representations}, 2019.

\bibitem{badia2020never}
Adri{\`a}~Puigdom{\`e}nech Badia, Pablo Sprechmann, Alex Vitvitskyi, Daniel
  Guo, Bilal Piot, Steven Kapturowski, Olivier Tieleman, Mart{\'\i}n Arjovsky,
  Alexander Pritzel, Andew Bolt, et~al.
\newblock Never give up: Learning directed exploration strategies.
\newblock {\em International Conference on Learning Representations}, 2020.

\bibitem{badia2020agent57}
Adri{\`a}~Puigdom{\`e}nech Badia, Bilal Piot, Steven Kapturowski, Pablo
  Sprechmann, Alex Vitvitskyi, Zhaohan~Daniel Guo, and Charles Blundell.
\newblock Agent57: Outperforming the atari human benchmark.
\newblock In {\em International Conference on Machine Learning}, pages
  507--517. PMLR, 2020.

\bibitem{machado2018count}
Marlos~C Machado, Marc~G Bellemare, and Michael Bowling.
\newblock Count-based exploration with the successor representation.
\newblock In {\em ICML Workshop on Exploration in Reinforcement Learning},
  2018.

\bibitem{oh2018directed}
Min-hwan Oh and Garud Iyengar.
\newblock Directed exploration in pac model-free reinforcement learning.
\newblock In {\em ICML Workshop on Exploration in Reinforcement Learning},
  2018.

\bibitem{machado2020count}
Marlos~C Machado, Marc~G Bellemare, and Michael Bowling.
\newblock Count-based exploration with the successor representation.
\newblock In {\em Proceedings of the AAAI Conference on Artificial
  Intelligence}, 2020.

\bibitem{cohn1996active}
David~A Cohn, Zoubin Ghahramani, and Michael~I Jordan.
\newblock Active learning with statistical models.
\newblock {\em Journal of artificial intelligence research}, 4:129--145, 1996.

\bibitem{itti2009bayesian}
Laurent Itti and Pierre Baldi.
\newblock Bayesian surprise attracts human attention.
\newblock {\em Vision research}, 49(10):1295--1306, 2009.

\bibitem{araya2010pomdp}
Mauricio Araya, Olivier Buffet, Vincent Thomas, and Fran{\c{c}}cois Charpillet.
\newblock A pomdp extension with belief-dependent rewards.
\newblock {\em Advances in neural information processing systems}, 23, 2010.

\bibitem{sun2011planning}
Yi~Sun, Faustino Gomez, and J{\"u}rgen Schmidhuber.
\newblock Planning to be surprised: Optimal bayesian exploration in dynamic
  environments.
\newblock In {\em International conference on artificial general intelligence},
  pages 41--51. Springer, 2011.

\bibitem{still2012information}
Susanne Still and Doina Precup.
\newblock An information-theoretic approach to curiosity-driven reinforcement
  learning.
\newblock {\em Theory in Biosciences}, 131(3):139--148, 2012.

\bibitem{houthooft2016vime}
Rein Houthooft, Xi~Chen, Yan Duan, John Schulman, Filip De~Turck, and Pieter
  Abbeel.
\newblock Vime: Variational information maximizing exploration.
\newblock {\em Advances in neural information processing systems}, 29, 2016.

\bibitem{sekar2020planning}
Ramanan Sekar, Oleh Rybkin, Kostas Daniilidis, Pieter Abbeel, Danijar Hafner,
  and Deepak Pathak.
\newblock Planning to explore via self-supervised world models.
\newblock In {\em International Conference on Machine Learning}, pages
  8583--8592. PMLR, 2020.

\bibitem{mendonca2021discovering}
Russell Mendonca, Oleh Rybkin, Kostas Daniilidis, Danijar Hafner, and Deepak
  Pathak.
\newblock Discovering and achieving goals via world models.
\newblock {\em Advances in Neural Information Processing Systems},
  34:24379--24391, 2021.

\bibitem{schmidhuber1991curious}
J{\"u}rgen Schmidhuber.
\newblock Curious model-building control systems.
\newblock In {\em Proc. international joint conference on neural networks},
  pages 1458--1463, 1991.

\bibitem{oudeyer2007intrinsic}
Pierre-Yves Oudeyer, Frdric Kaplan, and Verena~V Hafner.
\newblock Intrinsic motivation systems for autonomous mental development.
\newblock {\em IEEE transactions on evolutionary computation}, 11(2):265--286,
  2007.

\bibitem{azar2019world}
Mohammad~Gheshlaghi Azar, Bilal Piot, Bernardo~Avila Pires, Jean-Bastien Grill,
  Florent Altch{\'e}, and R{\'e}mi Munos.
\newblock World discovery models.
\newblock {\em arXiv preprint arXiv:1902.07685}, 2019.

\bibitem{hazan2019provably}
Elad Hazan, Sham Kakade, Karan Singh, and Abby Van~Soest.
\newblock Provably efficient maximum entropy exploration.
\newblock In {\em International Conference on Machine Learning}, pages
  2681--2691. PMLR, 2019.

\bibitem{liu2021behavior}
Hao Liu and Pieter Abbeel.
\newblock Behavior from the void: Unsupervised active pre-training.
\newblock {\em Advances in Neural Information Processing Systems},
  34:18459--18473, 2021.

\bibitem{guo2021geometric}
Zhaohan~Daniel Guo, Mohammad~Gheshlaghi Azar, Alaa Saade, Shantanu Thakoor,
  Bilal Piot, Bernardo~Avila Pires, Michal Valko, Thomas Mesnard, Tor
  Lattimore, and R{\'e}mi Munos.
\newblock Geometric entropic exploration.
\newblock {\em arXiv preprint arXiv:2101.02055}, 2021.

\bibitem{yarats2021reinforcement}
Denis Yarats, Rob Fergus, Alessandro Lazaric, and Lerrel Pinto.
\newblock Reinforcement learning with prototypical representations.
\newblock In {\em International Conference on Machine Learning}, pages
  11920--11931. PMLR, 2021.

\bibitem{gregor2017variational}
Karol Gregor, Danilo~Jimenez Rezende, and Daan Wierstra.
\newblock Variational intrinsic control.
\newblock {\em International Conference on Learning Representations}, 2017.

\bibitem{achiam2018variational}
Joshua Achiam, Harrison Edwards, Dario Amodei, and Pieter Abbeel.
\newblock Variational option discovery algorithms.
\newblock {\em arXiv preprint arXiv:1807.10299}, 2018.

\bibitem{eysenbach2019diversity}
Benjamin Eysenbach, Abhishek Gupta, Julian Ibarz, and Sergey Levine.
\newblock Diversity is all you need: Learning skills without a reward function.
\newblock {\em International Conference on Learning Representations}, 2019.

\bibitem{lee2019efficient}
Lisa Lee, Benjamin Eysenbach, Emilio Parisotto, Eric Xing, Sergey Levine, and
  Ruslan Salakhutdinov.
\newblock Efficient exploration via state marginal matching.
\newblock {\em arXiv preprint arXiv:1906.05274}, 2019.

\bibitem{campos2020explore}
V{\'\i}ctor Campos, Alexander Trott, Caiming Xiong, Richard Socher, Xavier
  Gir{\'o}-i Nieto, and Jordi Torres.
\newblock Explore, discover and learn: Unsupervised discovery of state-covering
  skills.
\newblock In {\em International Conference on Machine Learning}, pages
  1317--1327. PMLR, 2020.

\bibitem{sharma2020dynamics}
Archit Sharma, Shixiang Gu, Sergey Levine, Vikash Kumar, and Karol Hausman.
\newblock Dynamics-aware unsupervised discovery of skills.
\newblock {\em International Conference on Learning Representations}, 2020.

\bibitem{baumli2021relative}
Kate Baumli, David Warde-Farley, Steven Hansen, and Volodymyr Mnih.
\newblock Relative variational intrinsic control.
\newblock In {\em Proceedings of the AAAI Conference on Artificial
  Intelligence}, pages 6732--6740, 2021.

\bibitem{groth2021curiosity}
Oliver Groth, Markus Wulfmeier, Giulia Vezzani, Vibhavari Dasagi, Tim Hertweck,
  Roland Hafner, Nicolas Heess, and Martin Riedmiller.
\newblock Is curiosity all you need? on the utility of emergent behaviours from
  curious exploration.
\newblock {\em arXiv preprint arXiv:2109.08603}, 2021.

\bibitem{kwon2021variational}
Taehwan Kwon.
\newblock Variational intrinsic control revisited.
\newblock {\em International Conference on Learning Representations}, 2022.

\bibitem{liu2021aps}
Hao Liu and Pieter Abbeel.
\newblock Aps: Active pretraining with successor features.
\newblock In {\em International Conference on Machine Learning}, pages
  6736--6747. PMLR, 2021.

\bibitem{eysenbach2022information}
Benjamin Eysenbach, Ruslan Salakhutdinov, and Sergey Levine.
\newblock The information geometry of unsupervised reinforcement learning.
\newblock {\em International Conference on Learning Representations}, 2022.

\bibitem{laskin2022cic}
Michael Laskin, Hao Liu, Xue~Bin Peng, Denis Yarats, Aravind Rajeswaran, and
  Pieter Abbeel.
\newblock Cic: Contrastive intrinsic control for unsupervised skill discovery.
\newblock {\em arXiv preprint arXiv:2202.00161}, 2022.

\bibitem{andrychowicz2017hindsight}
Marcin Andrychowicz, Filip Wolski, Alex Ray, Jonas Schneider, Rachel Fong,
  Peter Welinder, Bob McGrew, Josh Tobin, OpenAI Pieter~Abbeel, and Wojciech
  Zaremba.
\newblock Hindsight experience replay.
\newblock {\em Advances in neural information processing systems}, 30, 2017.

\bibitem{florensa2018automatic}
Carlos Florensa, David Held, Xinyang Geng, and Pieter Abbeel.
\newblock Automatic goal generation for reinforcement learning agents.
\newblock In {\em International conference on machine learning}, pages
  1515--1528. PMLR, 2018.

\bibitem{nair2018visual}
Ashvin~V Nair, Vitchyr Pong, Murtaza Dalal, Shikhar Bahl, Steven Lin, and
  Sergey Levine.
\newblock Visual reinforcement learning with imagined goals.
\newblock {\em Advances in neural information processing systems}, 31, 2018.

\bibitem{fang2019curriculum}
Meng Fang, Tianyi Zhou, Yali Du, Lei Han, and Zhengyou Zhang.
\newblock Curriculum-guided hindsight experience replay.
\newblock {\em Advances in neural information processing systems}, 32, 2019.

\bibitem{zhang2020automatic}
Yunzhi Zhang, Pieter Abbeel, and Lerrel Pinto.
\newblock Automatic curriculum learning through value disagreement.
\newblock {\em Advances in Neural Information Processing Systems},
  33:7648--7659, 2020.

\bibitem{buesing2018woulda}
Lars Buesing, Theophane Weber, Yori Zwols, Sebastien Racaniere, Arthur Guez,
  Jean-Baptiste Lespiau, and Nicolas Heess.
\newblock Woulda, coulda, shoulda: Counterfactually-guided policy search.
\newblock {\em arXiv preprint arXiv:1811.06272}, 2018.

\bibitem{oberst2019counterfactual}
Michael Oberst and David Sontag.
\newblock Counterfactual off-policy evaluation with gumbel-max structural
  causal models.
\newblock In {\em International Conference on Machine Learning}, pages
  4881--4890. PMLR, 2019.

\bibitem{lorberbom2021learning}
Guy Lorberbom, Daniel~D Johnson, Chris~J Maddison, Daniel Tarlow, and Tamir
  Hazan.
\newblock Learning generalized gumbel-max causal mechanisms.
\newblock {\em Advances in Neural Information Processing Systems},
  34:26792--26803, 2021.

\bibitem{pearl2009causality}
Judea Pearl.
\newblock {\em Causality}.
\newblock Cambridge university press, 2009.

\bibitem{song2019v}
H~Francis Song, Abbas Abdolmaleki, Jost~Tobias Springenberg, Aidan Clark,
  Hubert Soyer, Jack~W Rae, Seb Noury, Arun Ahuja, Siqi Liu, Dhruva Tirumala,
  et~al.
\newblock V-mpo: On-policy maximum a posteriori policy optimization for
  discrete and continuous control.
\newblock {\em arXiv preprint arXiv:1909.12238}, 2019.

\bibitem{stepleton2017pycolab}
Thomas Stepleton.
\newblock The pycolab game engine, 2017.
\newblock {\em URL https://github. com/deepmind/pycolab}, 2017.

\bibitem{bellemare2013arcade}
Marc~G Bellemare, Yavar Naddaf, Joel Veness, and Michael Bowling.
\newblock The arcade learning environment: An evaluation platform for general
  agents.
\newblock {\em Journal of Artificial Intelligence Research}, 47:253--279, 2013.

\bibitem{machado2018revisiting}
Marlos~C Machado, Marc~G Bellemare, Erik Talvitie, Joel Veness, Matthew
  Hausknecht, and Michael Bowling.
\newblock Revisiting the arcade learning environment: Evaluation protocols and
  open problems for general agents.
\newblock {\em Journal of Artificial Intelligence Research}, 61:523--562, 2018.

\bibitem{oord2018representation}
Aaron van~den Oord, Yazhe Li, and Oriol Vinyals.
\newblock Representation learning with contrastive predictive coding.
\newblock {\em arXiv preprint arXiv:1807.03748}, 2018.

\bibitem{tschannen2019mutual}
Michael Tschannen, Josip Djolonga, Paul~K Rubenstein, Sylvain Gelly, and Mario
  Lucic.
\newblock On mutual information maximization for representation learning.
\newblock {\em arXiv preprint arXiv:1907.13625}, 2019.

\bibitem{he2020momentum}
Kaiming He, Haoqi Fan, Yuxin Wu, Saining Xie, and Ross Girshick.
\newblock Momentum contrast for unsupervised visual representation learning.
\newblock In {\em IEEE/CVF conference on computer vision and pattern
  recognition}, 2020.

\bibitem{chen2020simple}
Ting Chen, Simon Kornblith, Mohammad Norouzi, and Geoffrey Hinton.
\newblock A simple framework for contrastive learning of visual
  representations.
\newblock In {\em International conference on machine learning}, pages
  1597--1607. PMLR, 2020.

\bibitem{wang2021understanding}
Feng Wang and Huaping Liu.
\newblock Understanding the behaviour of contrastive loss.
\newblock In {\em Proceedings of the IEEE/CVF conference on computer vision and
  pattern recognition}, pages 2495--2504, 2021.

\bibitem{harutyunyan2019hindsight}
Anna Harutyunyan, Will Dabney, Thomas Mesnard, Mohammad Gheshlaghi~Azar, Bilal
  Piot, Nicolas Heess, Hado~P van Hasselt, Gregory Wayne, Satinder Singh, Doina
  Precup, et~al.
\newblock Hindsight credit assignment.
\newblock {\em Advances in neural information processing systems}, 32, 2019.

\bibitem{nota2021posterior}
Chris Nota, Philip Thomas, and Bruno~C Da~Silva.
\newblock Posterior value functions: Hindsight baselines for policy gradient
  methods.
\newblock In {\em International Conference on Machine Learning}, pages
  8238--8247. PMLR, 2021.

\bibitem{mesnard2021counterfactual}
Thomas Mesnard, Th{\'e}ophane Weber, Fabio Viola, Shantanu Thakoor, Alaa Saade,
  Anna Harutyunyan, Will Dabney, Tom Stepleton, Nicolas Heess, Arthur Guez,
  et~al.
\newblock Counterfactual credit assignment in model-free reinforcement
  learning.
\newblock In {\em International Conference on Machine Learning}, 2021.

\bibitem{zhang2020learning}
Amy Zhang, Rowan McAllister, Roberto Calandra, Yarin Gal, and Sergey Levine.
\newblock Learning invariant representations for reinforcement learning without
  reconstruction.
\newblock {\em arXiv preprint arXiv:2006.10742}, 2020.

\bibitem{bica2021invariant}
Ioana Bica, Daniel Jarrett, and Mihaela van~der Schaar.
\newblock Invariant causal imitation learning for generalizable policies.
\newblock {\em Advances in Neural Information Processing Systems},
  34:3952--3964, 2021.

\bibitem{lu2022invariant}
Chaochao Lu, Jos{\'e}~Miguel Hern{\'a}ndez-Lobato, and Bernhard Sch{\"o}lkopf.
\newblock Invariant causal representation learning for generalization in
  imitation and reinforcement learning.
\newblock In {\em ICLR 2022}, 2022.

\bibitem{louizos2016variational}
Christos Louizos, Kevin Swersky, Yujia Li, Max Welling, and Richard Zemel.
\newblock The variational fair autoencoder.
\newblock {\em International Conference on Learning Representations}, 2016.

\bibitem{moyer2018invariant}
Daniel Moyer, Shuyang Gao, Rob Brekelmans, Aram Galstyan, and Greg Ver~Steeg.
\newblock Invariant representations without adversarial training.
\newblock {\em Advances in Neural Information Processing Systems}, 31, 2018.

\bibitem{lotfollahi2019conditional}
Mohammad Lotfollahi, Mohsen Naghipourfar, Fabian~J Theis, and F~Alexander Wolf.
\newblock Conditional out-of-sample generation for unpaired data using trvae.
\newblock {\em arXiv preprint arXiv:1910.01791}, 2019.

\bibitem{foster2022contrastive}
Adam Foster, {\'A}rpi Vez{\'e}r, Craig~A Glastonbury, P{\'a}id{\'\i} Creed,
  Samer Abujudeh, and Aaron Sim.
\newblock Contrastive mixture of posteriors.
\newblock In {\em International Conference on Machine Learning}, pages
  6578--6621. PMLR, 2022.

\bibitem{agakov2004algorithm}
David Barber~Felix Agakov.
\newblock The im algorithm: a variational approach to information maximization.
\newblock {\em Advances in neural information processing systems}, 16(320):201,
  2004.

\bibitem{poole2019variational}
Ben Poole, Sherjil Ozair, Aaron Van Den~Oord, Alex Alemi, and George Tucker.
\newblock On variational bounds of mutual information.
\newblock In {\em International Conference on Machine Learning}, pages
  5171--5180. PMLR, 2019.

\bibitem{hessel2019multi}
Matteo Hessel, Hubert Soyer, Lasse Espeholt, Wojciech Czarnecki, Simon Schmitt,
  and Hado van Hasselt.
\newblock Multi-task deep reinforcement learning with popart.
\newblock In {\em Proceedings of the AAAI Conference on Artificial
  Intelligence}, 2019.

\bibitem{he2016deep}
Kaiming He, Xiangyu Zhang, Shaoqing Ren, and Jian Sun.
\newblock Deep residual learning for image recognition.
\newblock In {\em Proceedings of the IEEE conference on computer vision and
  pattern recognition}, pages 770--778, 2016.

\bibitem{wu2018group}
Yuxin Wu and Kaiming He.
\newblock Group normalization.
\newblock In {\em Proceedings of the European conference on computer vision
  (ECCV)}, pages 3--19, 2018.

\bibitem{cho2014properties}
Kyunghyun Cho, Bart Van~Merri{\"e}nboer, Dzmitry Bahdanau, and Yoshua Bengio.
\newblock On the properties of neural machine translation: Encoder-decoder
  approaches.
\newblock {\em arXiv preprint arXiv:1409.1259}, 2014.

\bibitem{mnih2015human}
Volodymyr Mnih, Koray Kavukcuoglu, David Silver, Andrei~A Rusu, Joel Veness,
  Marc~G Bellemare, Alex Graves, Martin Riedmiller, Andreas~K Fidjeland, Georg
  Ostrovski, et~al.
\newblock Human-level control through deep reinforcement learning.
\newblock {\em nature}, 518(7540):529--533, 2015.

\bibitem{van2016deep}
Hado Van~Hasselt, Arthur Guez, and David Silver.
\newblock Deep reinforcement learning with double q-learning.
\newblock In {\em AAAI conference on artificial intelligence}, 2016.

\bibitem{jarrett2021time}
Daniel Jarrett, Ioana Bica, and Mihaela van~der Schaar.
\newblock Time-series generation by contrastive imitation.
\newblock {\em Advances in Neural Information Processing Systems},
  34:28968--28982, 2021.

\bibitem{hafner2021benchmarking}
Danijar Hafner.
\newblock Benchmarking the spectrum of agent capabilities.
\newblock {\em arXiv preprint arXiv:2109.06780}, 2021.

\bibitem{henaff2022exploration}
Mikael Henaff, Roberta Raileanu, Minqi Jiang, and Tim Rockt{\"a}schel.
\newblock Exploration via elliptical episodic bonuses.
\newblock {\em arXiv preprint arXiv:2210.05805}, 2022.

\bibitem{anonymous2022robust}
Alaa Saade, Steven Kapturowski, Daniele Calandriello, Charles Blundell, Michal
  Valko, Pablo Sprechmann, and Bilal Piot.
\newblock Robust exploration via clustering-based online density estimation.
\newblock 2023.

\bibitem{piot2022blade}
Bilal Piot, Zhaohan~Daniel Guo, Shantanu Thakoor, and Mohammad~Gheshlaghi Azar.
\newblock Blade: Robust exploration via diffusion models.
\newblock In {\em Deep Reinforcement Learning Workshop NeurIPS 2022}, 2022.

\bibitem{tang2022understanding}
Yunhao Tang, Zhaohan~Daniel Guo, Pierre~Harvey Richemond, Bernardo~{\'A}vila
  Pires, Yash Chandak, R{\'e}mi Munos, Mark Rowland, Mohammad~Gheshlaghi Azar,
  Charline~Le Lan, Clare Lyle, et~al.
\newblock Understanding self-predictive learning for reinforcement learning.
\newblock {\em arXiv preprint arXiv:2212.03319}, 2022.

\bibitem{yang2022dichotomy}
Mengjiao Yang, Dale Schuurmans, Pieter Abbeel, and Ofir Nachum.
\newblock Dichotomy of control.
\newblock {\em arXiv preprint arXiv:2210.13435}, 2022.

\end{thebibliography}

\end{document}